\documentclass[11pt]{article}

\usepackage[margin=1in]{geometry}
\usepackage{amsmath,amssymb,amsthm,mathtools}

\usepackage{tikz}
\usetikzlibrary{arrows.meta,positioning}
\usepackage[colorlinks=true,linkcolor=blue,urlcolor=blue]{hyperref}
\usepackage{xcolor}
\usepackage{authblk}
\usepackage[nameinlink,capitalise]{cleveref}

\newtheorem{theorem}{Theorem}[section] 
\newtheorem{proposition}[theorem]{Proposition}
\newtheorem{definition}{Definition}[section]
\newtheorem{lemma}[theorem]{Lemma}
\newtheorem{assumption}{Assumption}
\newtheorem{corollary}[theorem]{Corollary}

\newtheorem*{theorem*}{Theorem}
\newtheorem*{proposition*}{Proposition}
\newtheorem*{corollary*}{Corollary}

\newcommand{\Comments}{1}
\newcommand{\kibitz}[2]{\ifnum\Comments=1\textcolor{#1}{#2}\fi}

\newcommand{\Xcal}{\mathcal X}
\newcommand{\Hcal}{\mathcal H}
\newcommand{\N}{\mathbb N}
\newcommand{\one}{\mathbf 1}
\newcommand{\MF}{\operatorname{MF}}
\newcommand{\MFUnif}{\MF_{\mathrm{unif}}}
\newcommand{\SMF}{\operatorname{SMF}}
\newcommand{\Reg}{\operatorname{Reg}}

\newcommand{\supp}{\operatorname{supp}}
\newcommand{\Ldim}{\operatorname{Ldim}}

\newcommand{\Vcal}{\mathcal V}
\newcommand{\Acal}{\mathcal A}
\newcommand{\Bcal}{\mathcal B}
\newcommand{\LeafCap}{\operatorname{Cap}}

\newcommand{\naturals}{\mathbb{N}}

\title{No-Regret Language Generation}
\author{Jiaxun Li, Vinod Raman, Ambuj Tewari}
\affil{Department of Statistics, University of Michigan}
\affil{\texttt{\{jasonli, vkraman, tewaria\}@umich.edu}}
\date{\today}
\date{August 2026}

\begin{document}
\maketitle

\begin{abstract}
We introduce no-regret generation, which requires a generator to make a sublinear number of mistakes uniformly over all demonstration sequences and compatible target languages. We also formulate an agnostic variant that allows arbitrary demonstrations that may not be compatible with any languages. For both settings, we introduce tree-based mistake-forcing dimensions that exactly characterize finite-horizon minimax complexity and establish necessary and sufficient conditions for no-regret generation. Our characterizations reveal a rich spectrum of generation rates: up to constant rescaling, every admissible nondecreasing profile is attainable in both the realizable and agnostic settings. Finally, we connect generation complexity to online prediction: for language classes of finite Littlestone dimension \(d\), we obtain realizable mistake bounds of \(O(d\log(1+T/d))\) and agnostic regret bounds of \(O(\sqrt{dT\log(1+T/d)})\). The realizable bound is optimal up to universal constants.
\end{abstract}

\tableofcontents

\newpage
\section{Introduction}
\label{sec:introduction}

The formal study of language generation has emerged as an important theoretical
enterprise with the rapid development of modern generative models. The success of these models raises a fundamental unsupervised learning question:
given only positive examples from an unknown
language, when can an algorithm produce new examples that also belong to that
language? This question has its origins in Gold's classical framework of
identification in the limit \cite{gold1967language} and Angluin's work on
learning formal languages from positive data
\cite{angluin1980inductive}. Whereas identification requires recovering the
unknown language, generation asks only for new valid examples. Kleinberg and
Mullainathan formalized this distinction and proved the striking result that
every countable collection of infinite languages is generatable in the limit
\cite{kleinberg2024language}. Their result initiated a rapidly developing
literature studying uniform and non-uniform generation
\cite{li2024generationlenslearningtheory,charikar2025exploring,
charikar2026pareto}, breadth and density
\cite{kalavasis2025limits,kalavasis2026characterizations,
kleinberg2025density}, robustness to noise and incomplete demonstrations
\cite{ramanRaman2025noisy,bai2026noise,mehrotra2026contamination}, and the
mistake, memory, and computational resources required for generation
\cite{kleinberg2026mistakebounded,arenas2025complexity,
kleinberg2026memory}.

In an usual generation game studied in these works, an unknown target language
\(h^\star\in\Hcal\) is fixed before the interaction. At every round, the
generator observes a new example from \(h^\star\) and must produce
a new example that has not been demonstrated. The generator succeeds
if, for every target and every enumeration of its positive examples, all of
its outputs are valid after some finite time. 

Requiring a finite last mistake may be too rigid for evaluating generation over
a long interaction. A generator may continue to make
occasional mistakes at arbitrarily late times while their frequency converges
to zero. This behavior is fundamentally different from making mistakes at a
constant rate and therefore motivates a distinct criterion for successful
generation.

\paragraph{Motivating example 1.}
Requiring a finite last mistake can exclude classes with strong
finite-horizon behavior. To see this, consider an example space

$$
    \Xcal
    =
    \bigcup_{n=1}^{\infty}B_n,
$$
where each block \(B_n\) contains $n$ different elements and are pairwise disjoint. For every set \(R\subseteq\Xcal\) satisfying

$$
    |R\cap B_n|\leq1
    \qquad
    \text{for every }n,
$$
let \(h_R\) be the language with

$$
    \supp(h_R):=\Xcal\setminus R,
$$
and let \(\Hcal_{\mathrm{tr}}\) be the collection of all such languages.
Intuitively, each block is a pool of candidate outputs containing at most one
invalid point. If the generator knows that the interaction will last \(T\)
rounds, it can choose one sufficiently large block and remain there throughout
the interaction, and will eventually make at most one mistake. 

The infinite interaction behaves differently. Because every block is finite,
an anytime generator cannot rely on one block forever. As it moves through
new blocks, a target can exclude one of its outputs in each block, causing
mistakes at arbitrarily late times. As we will prove later, \(\Hcal_{\mathrm{tr}}\) is
not generatable in the limit. Nevertheless, it admits one generator
whose worst-case number of mistakes through time \(T\) is \(o(T)\). Thus,
\(\Hcal_{\mathrm{tr}}\) has strong finite-horizon generation performance even
though no generator can guarantee a finite last mistake.

This example suggests replacing eventual perfect validity with a rate-sensitive
criterion. Panigrahi, Wei, and Zhang pursue this direction in their study of
\emph{hallucination rates}
\cite{panigrahi2026hallucinationrates}. Their framework permits infinitely
many invalid outputs and instead controls their asymptotic frequency. It
therefore distinguishes sparse hallucinations from persistent ones and
captures the desirable behavior exhibited by \(\Hcal_{\mathrm{tr}}\).

Their guarantee, however, is evaluated separately for each target and each
enumeration, so the resulting convergence rate may depend on both. This
evaluation is from the perspective of the adversary.
From the generator’s perspective, the target remains unknown: after any
finite number of rounds, many different languages may still be compatible
with the demonstrations. This motivates requiring a
finite-horizon guarantee that holds uniformly over all compatible targets.

\paragraph{Motivating example 2.}
This lack of finite-horizon control can occur even for a class that is
generatable in the limit. Consider the cofinite class

$$
    \Hcal_{\mathrm{cof}}
    :=
    \left\{
        h_F:
        \supp(h_F)=\Xcal\setminus F,\quad
        F\subseteq\Xcal\text{ finite}
    \right\}.
$$

Each language in this class contains every example except those in a finite
exceptional set. The class is countable and is therefore generatable in the
limit. Consequently, for every fixed target and enumeration, the generator
can eventually produce only valid examples and hence achieve a vanishing
hallucination rate.

From the generator's viewpoint, however, the exceptional set \(F\) is
unknown. After any finite number of demonstrations, many different choices of
\(F\) remain possible. Indeed, for any horizon \(T\), the language \(h_F\) can be chosen such that it  contains none of the generator's outputs. Thus, 
among the targets compatible with the generator's observations, there is
always one under which it makes \(T\) mistakes in \(T\) rounds.

Motivated by these examples, we study a finite-horizon minimax formulation of
language generation. 
For a deterministic generator \(G\), demonstrations \(x_{1:T}\), and a
compatible target
\[
    h^\star\in\Hcal(x_{1:T})
    :=
    \left\{
        h\in\Hcal:
        h(x_t)=1\text{ for every }t\leq T
    \right\},
\]
let
\[
    M_T(G,h^\star;x_{1:T})
    :=
    \sum_{t=1}^T
    \one\{h^\star(\widehat x_t)=0\}
\]
denote the number of invalid outputs through time \(T\). We call a class \emph{realizably no-regret generatable} if one deterministic
anytime generator \(G\) satisfies
\[
    \sup_{\substack{
        x_{1:T}\\
        h^\star\in\Hcal(x_{1:T})
    }}
    M_T(G,h^\star;x_{1:T})
    =
    o(T).
\]

The no-regret criterion is well established in online learning. In online prediction, sublinear
regret requires the learner's cumulative loss to approach that of the best
comparator \cite{cesa2006prediction,rakhlin2015online}. This
viewpoint also extends to partial-feedback problems such as apple tasting,
where the learner observes feedback only after taking certain actions
\cite{helmbold2000apple,raman2024apple}. Language generation presents a
related feedback limitation: the generator never observes whether its own
output is valid. 

The realizable framework still assumes that the demonstrations observed
through time \(T\) share a common explanation in \(\Hcal\). In many
applications, however, the demonstrations may contain noise or may not be
jointly consistent with any candidate language. In that case, there is no
distinguished target relative to which mistakes can be counted.

We therefore also introduce an agnostic framework. For every comparator
\(h\in\Hcal\), we compare the number of generated examples that are invalid
under \(h\) with the number of demonstrations that are invalid under the same
hypothesis. Writing
\[
    \ell_h(x):=\one\{h(x)=0\},
\]
the resulting regret is
\[
\begin{aligned}
    \Reg_T(G,h;x_{1:T})
    &:=
    \sum_{t=1}^T
    \left(
        \ell_h(\widehat x_t)-\ell_h(x_t)
    \right)\\
    &=
    \sum_{t=1}^T
    \bigl(h(x_t)-h(\widehat x_t)\bigr).
\end{aligned}
\]
A class is \emph{agnostically no-regret generatable} if one deterministic
anytime generator satisfies
\[
    \sup_{\substack{x_{1:T}\\h\in\Hcal}}
    \Reg_T(G,h;x_{1:T})
    =
    o(T).
\]
Thus the generator must asymptotically perform at least as well as every
language in \(\Hcal\), measured on the same sequence of demonstrations and
generated examples. When \(h\) contains every demonstration, its
demonstration loss is zero and the regret reduces exactly to the realizable
mistake count.

\textit{Remark:} Our definition is \emph{one} possible formulation of agnostic language generation. Other possibilities include competing only with hypotheses that best explain the demonstrations, or having a budget on the corrupted demonstrations. Studying these alternatives remains open. The framework developed here provides a general starting point that can be adapted to such settings by modifying the comparator class or the loss function.

\subsection{Main contributions}

\label{subsec:introduction-contributions}

Our main contributions are as follows.

\begin{itemize}
    \item
    \textbf{No-regret and agnostic language generation.}
    We introduce no-regret generation as a new success criterion that
    measures how the cumulative number of mistakes grows with the horizon. We introduce the mistake-forcing dimensions
\(\MF_T(\Hcal)\) and \(\MFUnif(\Hcal)\) for realizable generation and the
signed mistake-forcing dimension \(\SMF_T(\Hcal)\) for agnostic generation.
These dimensions exactly capture the optimal realizable mistake count, the
optimal uniform total-mistake bound, and the optimal agnostic regret,
respectively. In particular, \(\MFUnif\) gives an answer
to the first open question posed by Kleinberg, Peale, and Reingold
\cite{kleinberg2026mistakebounded}.

\begin{theorem*}[Informal]
A UUS class \(\Hcal\) is realizably no-regret generatable if and only if
\(\MF_T(\Hcal)=o(T)\), and it is agnostically no-regret generatable if and
only if \(\SMF_T(\Hcal)=o(T)\). Moreover, \(\Hcal\) is uniformly
mistake-bounded generatable if and only if
\(\MFUnif(\Hcal)<\infty\).
\end{theorem*}

\item
\textbf{The spectrum of generation rates.}
Our characterizations reveal that no-regret generation admits a much richer
range of behavior than a bounded-versus-unbounded classification suggests.
Essentially every admissible sublinear rate can arise in both the realizable
and agnostic settings.

\begin{theorem*}[Informal]
Up to constant factors, essentially every admissible nondecreasing profile
can occur as the optimal realizable mistake rate, and also as the optimal
agnostic regret rate, for suitable UUS classes. In particular, logarithmic,
polynomial, and arbitrarily slow sublinear rates are all attainable.
\end{theorem*}

    \item
    \textbf{From prediction to generation.}
    Previous work showed that online predictability and generatability are
    qualitatively incomparable
    \cite{li2024generationlenslearningtheory}. We show that a quantitative
    comparison emerges when introducing this no-regret generation.
    \begin{theorem*}[Informal]
If \(d:=\Ldim(\Hcal)<\infty\), then
\[
    \MF_T(\Hcal)
    =
    O\left(
        d\log\left(1+\frac{T}{d}\right)
    \right)
\]
and
\[
    \SMF_T(\Hcal)
    =
    O\left(
        \sqrt{
            dT\log\left(1+\frac{T}{d}\right)
        }
    \right).
\]
Hence every UUS class of finite Littlestone dimension is both realizably and agnostically no-regret generatable.
\end{theorem*}
     We also construct
    classes for which the realizable bound is
    \(\Theta(d\log(1+T/d))\), showing that its dependence on \(d\) and
    \(T\) is optimal up to universal constants.
\end{itemize}

\subsection{Related work}
\label{sec:related}

\paragraph{Language identification and generation.}
Gold introduced identification in the limit \cite{gold1967language}, and Angluin characterized identification from positive data \cite{angluin1980inductive}. Kleinberg and Mullainathan showed that every countable collection of infinite languages is generatable in the limit and established uniform guarantees for finite collections \cite{kleinberg2024language}. Li, Raman, and Tewari formalized uniform and non-uniform generation, characterized them using closure-based dimensions, and compared generation with prediction \cite{li2024generationlenslearningtheory}. Charikar and Pabbaraju studied non-uniform generation and its variants \cite{charikar2025exploring,charikar2026pareto}. Other work examines generation under unions \cite{hanneke2025union,bai2026noise}.

\paragraph{Breadth, density, and representation.}
Generation in the limit guarantees validity but may cover only a small part of the target language. Prior work studies validity and breadth \cite{kalavasis2025limits,kalavasis2026characterizations}, output density \cite{kleinberg2025density,kleinberg2026partial}, and representation of prescribed groups \cite{peale2025representative}. These works study which valid examples are generated; we study how mistakes accumulate over time.

\paragraph{Noise, contamination, and feedback.}
Prior work studies generation with finitely many noisy demonstrations \cite{ramanRaman2025noisy}, missing examples and feedback \cite{bai2026noise}, infinitely many corruptions \cite{mehrotra2026contamination}, and mistake and query feedback \cite{hanneke2026feedback}. These models retain a target language and restrict deviations from it. Our agnostic model assumes neither a target nor a noise budget. Instead, it compares output loss with demonstration loss under each \(h\in\Hcal\). Section~\ref{subsec:noisy-versus-agnostic} makes this connection quantitative.

Høgsgaard and Pabbaraju also study \emph{agnostic generation}, but in a statistical setting with i.i.d.\ samples \cite{hogsgaard2026agnostic}. They evaluate whether an output belongs to the unseen support of the sampling distribution. We study adversarial sequences and measure regret against candidate languages.

\paragraph{Mistake bounds and resource limits.}
Kleinberg, Peale, and Reingold studied mistake-bounded generation, including logarithmic mistake bounds for finite classes and extensions to noisy demonstrations \cite{kleinberg2026mistakebounded}. We instead study how the worst-case mistake count grows with the horizon and give an exact combinatorial characterization. Other work studies sample and computational complexity \cite{arenas2025complexity}, space efficiency \cite{flammarion2026space}, and memory constraints \cite{kleinberg2026memory}.

\paragraph{Online prediction.}
Littlestone dimension characterizes realizable mistake bounds in online binary prediction \cite{littlestone1988learning}. We show that it also controls realizable and agnostic generation rates, even though a generator chooses its outputs and never observes their labels. This provides a quantitative connection between prediction and generation.

\section{Preliminaries}
\label{sec:preliminaries}

\subsection{Language classes}
\label{subsec:language-classes}

Let \(\Xcal\) be a countable example space, and let $\Hcal\subseteq\{0,1\}^{\Xcal}$
be a nonempty binary hypothesis class. In the language-generation
interpretation, each \(h\in\Hcal\) represents a language, with
\[
    \supp(h):=\{x\in\Xcal:h(x)=1\}
\]
denoting its set of valid examples.

Write \(\mathbb N_0:=\mathbb N\cup\{0\}\). For \(n\in\mathbb N_0\),
write \([n]:=\{1,\ldots,n\}\), with \([0]:=\varnothing\), and abbreviate
a finite sequence \((x_1,\ldots,x_n)\) by \(x_{1:n}\). For an infinite
sequence, write $ x_{1:\infty}:=(x_1,x_2,\ldots).$
Throughout, \(\log\) denotes the base-two logarithm.
For a finite sequence \(x_{1:n}\), define its positive version space by
\[
    \Hcal(x_{1:n})
    :=
    \left\{
        h\in\Hcal:
        h(x_i)=1\ \text{for every }i\in[n]
    \right\}.
\]
For an infinite sequence \(x_{1:\infty}\), define
\[
    \Hcal(x_{1:\infty})
    :=
    \left\{
        h\in\Hcal:
        h(x_t)=1\ \text{for every }t\geq1
    \right\}.
\]
Throughout the paper, we impose the following nondegeneracy condition.

\begin{assumption}[Uniformly unbounded support]
\label{ass:uus}
A hypothesis class \(\Hcal\subseteq\{0,1\}^{\Xcal}\) satisfies the
\emph{uniformly unbounded support}\/ (UUS) condition if
\[
    |\supp(h)|=\infty
    \qquad
    \text{for every }h\in\Hcal.
\]
\end{assumption}

The UUS condition ensures that after any finite sequence of distinct
positive demonstrations from a language \(h\), another unseen positive
example from \(h\) remains available.

\subsection{Generation game and trajectories}
\label{subsec:generation-game}
The generation game proceeds in rounds. Before round \(t\), the observed
history consists of distinct demonstrations
\[
    x_{1:t-1}=(x_1,\ldots,x_{t-1}).
\]
Throughout, we adopt the convention that every demonstration sequence in the
generation game, whether finite \(x_{1:t}\) or infinite
\(x_1,x_2,\ldots\), consists of pairwise distinct examples unless stated
otherwise.

At round \(t\):

\begin{enumerate}
    \item the generator selects an example
    \[
        \widehat x_t
        \in
        \Xcal\setminus\{x_1,\ldots,x_{t-1}\};
    \]
    \item after observing \(\widehat x_t\), the adversary reveals a fresh
    demonstration
    \[
        x_t
        \in
        \Xcal\setminus\{x_1,\ldots,x_{t-1}\}.
    \]
\end{enumerate}

Following \cite{kleinberg2026mistakebounded}, we adopt the convention that
the generator moves before the adversary in each round. Generated examples
need not be distinct: the generator may reuse an example until that example
is revealed as a demonstration. Thus only previously demonstrated examples
are unavailable. The adversary may also reveal
\(x_t=\widehat x_t\).

\begin{definition}[Trajectory]
\label{def:trajectory}
A \emph{trajectory of length \(T\)} is a sequence
\[
    \tau
    =
    \bigl(
        (\widehat x_1,x_1),\ldots,
        (\widehat x_T,x_T)
    \bigr).
\]
Its set of revealed demonstrations is
\[
    D(\tau):=\{x_1,\ldots,x_T\}.
\]
We denote the empty trajectory by \(\tau_\varnothing:=()\).
The trajectory is \emph{legal} if, for every \(t\leq T\),
\[
    \widehat x_t,x_t
    \notin
    \{x_1,\ldots,x_{t-1}\}.
\]
Unless stated otherwise, every trajectory considered in this paper is
understood to be legal.

If \(\widehat x,x\notin D(\tau)\), we write
\[
    \tau\circ(\widehat x,x)
\]
for the trajectory obtained by extending \(\tau\) by one round, with
generated example \(\widehat x\) and revealed example \(x\).
\end{definition}

\begin{definition}[Generator]
\label{def:generator}
A \emph{deterministic anytime generator} \(G\) assigns to every finite
sequence of distinct demonstrations \(x_{1:n}\) an output
\[
    G(x_{1:n})
    \in
    \Xcal\setminus\{x_1,\ldots,x_n\}.
\]
At round \(t\), the generator outputs
\[
    \widehat x_t:=G(x_{1:t-1}).
\]

A \emph{horizon-\(T\) generator} is defined analogously, but only for
histories of length less than \(T\). It may depend on the horizon \(T\).
\end{definition}

Our formulation builds on the standard language-generation frameworks of
\cite{kleinberg2024language,li2024generationlenslearningtheory,
kleinberg2026mistakebounded}. In this setting, a target language
\(h^\star\in\Hcal\) is fixed, and the adversary reveals distinct
demonstrations
\(x_1,x_2,\ldots\in\supp(h^\star)\). At round \(t\), the generator incurs
the mistake loss
\(\one\{\widehat{x}_t\notin\supp(h^\star)\}\), which it does not observe
because the target language is unknown.

\subsection{Realizable and agnostic generation}
\label{subsec:realizable-agnostic-generation}

We study two extensions of this framework.  First, we introduce a
finite-horizon analysis of realizable generation.  A
length-\(T\) play remains realizable only when some language contains all
\(T\) demonstrations.  Second, we introduce an agnostic
game in which the demonstrations need not be jointly realized by any
language in \(\Hcal\).  The finite-horizon viewpoint
leads naturally to no-regret generation, which requires the corresponding
cumulative loss to grow sublinearly in \(T\).

We define the two settings formally. Fix a horizon \(T\), a deterministic generator \(G\), and distinct
demonstrations \(x_{1:T}\). The generated examples are determined
recursively by
\[
    \widehat{x}_t=G(x_{1:t-1}).
\]

For \(h\in\Hcal\) and \(x\in\Xcal\), define the \emph{invalidity loss}
by
\[
    \ell_h(x)
    :=
    \one\{h(x)=0\}
    =
    1-h(x).
\]

\begin{definition}[Realizable generation]
\label{def:realizable-generation}
The resulting length-\(T\) trajectory is \emph{realizable by}
\(h\in\Hcal\) if
\[
    h(x_t)=1
    \qquad
    \text{for every }t\leq T.
\]
The generator's realizable mistake count with respect to \(h\) is
\[
    M_T(G,h;x_{1:T})
    :=
    \sum_{t=1}^T
    \ell_h(\widehat x_t).
\]
For an infinite realizable trajectory \(x_{1:\infty}\) and
\(h\in\Hcal(x_{1:\infty})\), the total mistake count of an anytime generator
\(G\) is
\[
    M_\infty(G,h;x_{1:\infty})
    :=
    \sup_{T\geq1}M_T(G,h;x_{1:T})
    =
    \sum_{t=1}^{\infty}\ell_h(\widehat{x}_t).
\]
\end{definition}

\begin{definition}[Agnostic generation]
\label{def:agnostic-generation}
In the agnostic game, the demonstrations need not be realized by any
language in \(\Hcal\). For \(h\in\Hcal\), the generator's regret with
respect to \(h\) is
\[
\begin{aligned}
    \Reg_T(G,h;x_{1:T})
    &:=
    \sum_{t=1}^T
    \left(
        \ell_h(\widehat x_t)
        -
        \ell_h(x_t)
    \right)\\
    &=
    \sum_{t=1}^T
    \bigl(h(x_t)-h(\widehat x_t)\bigr).
\end{aligned}
\]
\end{definition}

Thus agnostic regret compares the number of generated examples outside
\(h\) with the number of demonstrations outside \(h\). Positive regret
means that the generator has produced more invalid examples than the
adversary has demonstrated, as measured with respect to \(h\).

If \(h\) realizes the demonstrations, then \(h(x_t)=1\) for every
\(t\leq T\), and hence
\[
    \Reg_T(G,h;x_{1:T})
    =
    M_T(G,h;x_{1:T}).
\]
Realizable generation is therefore the special case of agnostic generation.

\subsection{Success criteria}
\label{subsec:no-regret-success}

We first formalize the uniform total-mistake criterion, which is introduced by \cite{kleinberg2026mistakebounded}. 

\begin{definition}[Uniform mistake-bounded generation]
\label{def:uniform-mistake-bounded-generation}
A class \(\Hcal\) is \emph{uniformly mistake-bounded generatable} if there
exist a deterministic anytime generator \(G\) and a finite constant
\(B<\infty\) such that
\[
    \sup_{\substack{x_{1:\infty}\\h\in\Hcal(x_{1:\infty})}}
    M_\infty(G,h;x_{1:\infty})
    \leq B.
\]
\end{definition}

We next define the weaker no-regret criteria. All asymptotic bounds below
are uniform over the indicated targets and demonstration sequences.

\begin{definition}[Realizable no-regret generation]
\label{def:no-regret-realizable-generation}
A class \(\Hcal\) is \emph{realizably no-regret generatable} if there exists
one deterministic anytime generator \(G\) such that
\[
    \sup_{\substack{ x_{1:T}\\ h\in \Hcal(x_{1:T})
    }}
    M_T(G,h;x_{1:T})
    =
    o(T).
\]
\end{definition}

\begin{definition}[Agnostic no-regret generation]
\label{def:no-regret-agnostic-generation}
A class \(\Hcal\) is \emph{agnostically no-regret generatable} if there
exists one deterministic anytime generator \(G\) such that
\[
    \sup_{\substack{ x_{1:T}\\ h\in\Hcal}}
    \Reg_T(G,h;x_{1:T})
    =
    o(T).
\]
\end{definition}

A uniform finite mistake bound immediately implies realizable no-regret
generation. Moreover, because agnostic regret reduces to the realizable
mistake count whenever \(h\) contains every demonstration, agnostic
no-regret generation implies realizable no-regret generation. The converse
implications need not hold.

\section{Realizable No-Regret  Generation}
\label{sec:realizable-generation}

\subsection{Mistake-forcing dimension}

We introduce a tree-based dimension that measures how many mistakes an adversary can force, regardless of the generator's strategy. As with Littlestone dimension \cite{littlestone1988learning}, the tree represents adaptive interaction. The branching structure here reflects the move order of the generation game: the generator moves first, and the adversary chooses a response after observing the generated example.

\begin{definition}[Trajectory tree]
\label{def:trajectory-tree}
A \emph{trajectory tree rooted at a legal trajectory \(\tau_0\)} is a
rooted directed tree with the following labels:
\begin{enumerate}
    \item The root \(r\) carries \(\tau(r)=\tau_0\), and every node \(u\)
    carries a legal trajectory \(\tau(u)\) extending \(\tau_0\).
    \item Every edge \((u,v)\) is labeled by a legal generator output $\widehat{x}_{u,v}\notin D(\tau(u)).$
    
    \item The child \(v\) is labeled by a legal adversary response $x_v\notin D(\tau(u))$
    and carries the trajectory
    \[
        \tau(v)
        =
        \tau(u)\circ(\widehat{x}_{u,v},x_v).
    \]
\end{enumerate}
The tree is \emph{generator-complete} if, for every nonleaf \(u\) and every
legal output \(\widehat{x}\notin D(\tau(u))\), at least one outgoing edge
from \(u\) is labeled by \(\widehat{x}\).

A trajectory tree has \emph{horizon \(T\)} if it has depth \(T\) and every leaf has depth \(T\).
It is \emph{well-founded} if it has no infinite
branch. Thus, a well-founded tree may have unbounded finite depth, but every
maximal path is finite and ends at a leaf.
\end{definition}

A trajectory tree can be viewed as an adversarial strategy. At a node \(u\), the generator may choose any legal output \(\widehat{x}\). Generator-completeness guarantees that the tree contains at least one outgoing edge with this label, and the adversary may select one of the corresponding responses. Consequently, the game can be determined by a path through the tree. 
\begin{definition}[Mistake pattern]
\label{def:mistake-pattern}
Let \(\Hcal\subseteq\{0,1\}^{\Xcal}\) be nonempty, and let
\[
    \tau
    =
    \bigl((\widehat{x}_1,x_1),\ldots,
          (\widehat{x}_n,x_n)\bigr)
\]
be a legal trajectory. A \emph{mistake pattern} for \(\tau\) is a set
\(I\subseteq[n]\), where \(i\in I\) indicates that
\(\widehat{x}_i\) is a mistake.

The subclass compatible with the mistake pattern \(I\) is
\[
    \Hcal(\tau;I)
    :=
    \left\{
        h\in\Hcal:
        h(x_i)=1\ \text{for every }i\in[n],\quad
        h(\widehat{x}_i)=0\ \Longleftrightarrow i\in I
    \right\}.
\]
We call \(I\) \emph{feasible} if
\(\Hcal(\tau;I)\neq\varnothing\), meaning that some language in
\(\Hcal\) realizes the indicated pattern.
The set of feasible mistake patterns for \(\tau\) is
\[
    S_{\Hcal}(\tau)
    :=
    \left\{
        I\subseteq[n]:
        \Hcal(\tau;I)\neq\varnothing
    \right\}.
\]
In particular, \(S_{\Hcal}(\tau_\varnothing)=\{\varnothing\}\).
\end{definition}

\begin{definition}[\(m\)-bad trajectory]
\label{def:m-bad-trajectory}
Let \(m\in\mathbb{N}_0\). A legal trajectory \(\tau\) is
\emph{\(m\)-bad for \(\Hcal\)} if it admits a feasible mistake pattern
containing at least \(m\) mistakes; equivalently,
\[
    \tau \text{ is \(m\)-bad for \(\Hcal\)}
    \quad\Longleftrightarrow\quad
    \exists I\in S_{\Hcal}(\tau)
    \text{ such that } |I|\geq m.
\]
\end{definition}

Thus, a feasible mistake pattern certifies that the entire trajectory can be explained by a single language in $\Hcal$, and an $m$-bad trajectory further certifies that the generator may have committed at least $m$ mistakes.

\begin{definition}[Mistake-forcing tree]
\label{def:mistake-forcing-tree}
Let \(m\in\mathbb{N}_0\), and let \(\tau_0\) be a legal trajectory.

An \emph{\(m\)-mistake-forcing tree of horizon \(T\) for \(\Hcal\) rooted
at \(\tau_0\)} is a generator-complete trajectory tree of horizon \(T\)
rooted at \(\tau_0\) such that every leaf \(v\) satisfies
\[
    \tau(v)\text{ is \(m\)-bad for \(\Hcal\)}.
\]

A \emph{well-founded \(m\)-mistake-forcing tree for \(\Hcal\) rooted at
\(\tau_0\)} is defined analogously, with the horizon-\(T\) requirement
replaced by well-foundedness.
\end{definition}
The two forms of forcing trees capture different guarantees. A horizon-\(T\) tree forces \(m\) mistakes by the common terminal time \(T\). A well-founded tree allows the terminal time to depend on the path, but guarantees that every interaction reaches a leaf after finitely many rounds. Its depth may therefore be unbounded across different paths even though no individual path is infinite.

\begin{definition}[Mistake-forcing dimensions]
\label{def:mistake-forcing-dimensions}
For \(T\in\mathbb{N}_0\), the \emph{horizon-$T$ mistake-forcing
dimension} of \(\Hcal\) is
\[
    \MF_T(\Hcal)
    :=
    \max
    \left\{
        m\in\{0,\ldots,T\}:
        \begin{array}{c}
        \text{there exists an \(m\)-mistake-forcing tree of horizon \(T\)}\\
        \text{for \(\Hcal\) rooted at \(\tau_\varnothing\)}
        \end{array}
    \right\}.
\]
The \emph{uniform mistake-forcing dimension} of \(\Hcal\) is
\[
    \MFUnif(\Hcal)
    :=
    \sup
    \left\{
        m\in\mathbb{N}_0:
        \begin{array}{c}
        \text{there exists a well-founded \(m\)-mistake-forcing tree}\\
        \text{for \(\Hcal\) rooted at \(\tau_\varnothing\)}
        \end{array}
    \right\}.
\]
We set \(\MFUnif(\Hcal)=\infty\) when the forceable integers are unbounded.
\end{definition}
In summary, \(\MF_T(\Hcal)\) is the largest number of mistakes that the
adversary can guarantee by horizon \(T\), whereas \(\MFUnif(\Hcal)\) is the
largest uniform number of mistakes it can guarantee at some finite,
path-dependent stopping time. The next subsection shows that these
combinatorial quantities exactly characterize finite-horizon and uniform
realizable mistake complexity, respectively, and that the growth of
\(\MF_T(\Hcal)\) characterizes realizable no-regret generation.

In the next chapter, we will show that the mistake-forcing dimension plays a central role in characterizing no-regret generation.
\subsection{Characterizations of realizable no-regret generation}
\label{subsec:realizable-characterizations}

We now show that the mistake-forcing dimensions characterize three notions
of realizable generation. The finite-horizon dimension \(\MF_T(\Hcal)\)
gives the optimal worst-case mistake count at horizon \(T\), while the
    uniform dimension \(\MFUnif(\Hcal)\) characterizes the existence of a
uniform bound on the total number of mistakes. We then use the entire
profile \((\MF_T(\Hcal))_{T\geq1}\) to characterize realizable no-regret
generation.

\begin{theorem}
\label{thm:mf-finite-characterization}
For every nonempty UUS class \(\Hcal\) and every \(T\geq1\),
\[
    \MF_T(\Hcal)
    =
    \min_{G_T}
    \sup_{\substack{x_{1:T}\\ h\in\Hcal(x_{1:T})}}
    M_T(G_T,h;x_{1:T}).
\]
For each \(T\), the minimum is attained by a generator \(G_T\), which may
depend on \(T\).
\end{theorem}
The intuition behind the proof is that a generator-complete tree describes
an adversarial response to every possible output of the generator. Thus,
after fixing any generator, the adversary can follow the corresponding path
to a leaf and then choose a compatible target language witnessing at least
\(\MF_T(\Hcal)\) mistakes. Conversely, if the adversary cannot force one
more mistake from the current trajectory, the generator can choose an output
that keeps this impossible after every legal response. The complete
proof appears in Appendix~\ref{app:realizable-proofs}.

Similarly, we have a characterization for uniform total-mistake bounds, whose proof is also in Appendix~\ref{app:realizable-proofs}.

\begin{theorem}
\label{thm:mf-infinite-characterization}
For every nonempty UUS class \(\Hcal\),
\[
    \MFUnif(\Hcal)
    =
    \min_G
    \sup_{\substack{x_{1:\infty}\\h\in\Hcal(x_{1:\infty})}}
    M_\infty(G,h;x_{1:\infty}).
\]
The minimum is taken over deterministic anytime generators.
\end{theorem}

Consequently, \(\MFUnif(\Hcal)<\infty\) if and only if there exists a generator
with a finite uniform bound on the total number of mistakes over all
complete realizable plays. This characterization resolves Open Problem~1
posed by \cite{kleinberg2026mistakebounded}.

A finite uniform mistake bound immediately implies no-regret generation,
but no-regret generation only requires the cumulative number of mistakes to grow sublinearly. The natural candidate
characterization is therefore \(\MF_T(\Hcal)=o(T)\). The challenge is that
\(\MF_T(\Hcal)\) optimizes separately at each horizon and may be attained by
a different generator \(G_T\), whereas no-regret generation requires a
single anytime generator. The following theorem shows that this distinction
does not create a gap.

\begin{theorem}[Characterization of realizable no-regret generation]
\label{thm:horizon-anytime-main}
For every nonempty UUS class \(\Hcal\), the following are equivalent:
\begin{enumerate}
    \item \(\Hcal\) is realizably no-regret generatable;
    \item \(\MF_T(\Hcal)=o(T)\) as \(T\to\infty\).
\end{enumerate}
Moreover, let \(b:\mathbb N\to[1,\infty)\) be nondecreasing and
unbounded. If
\[
    \MF_T(\Hcal)\leq b(T)
    \qquad
    \text{for every }T\geq1,
\]
then there exists one deterministic anytime generator \(G\) such that,
simultaneously for every \(T\geq1\),
\[
    \sup_{\substack{x_{1:T}\\h\in\Hcal(x_{1:T})}}
    M_T(G,h;x_{1:T})
    \leq 4b(T).
\]
\end{theorem}

The proof of this theorem combines a sequence of horizon-dependent generators to obtain one anytime
generator. The proof can be found in Appendix~\ref{app:realizable-proofs}.

\subsection{Finite-horizon versus uniform mistake bounds}
\label{subsec:finite-versus-uniform}
Every finite-horizon mistake-forcing tree is well-founded, and therefore
\[
    \sup_{T\geq1}\MF_T(\Hcal)\leq\MFUnif(\Hcal).
\]
It is natural to ask whether equality always holds. In classical online
prediction, Littlestone dimension characterizes both finite-horizon
and uniform mistake complexity: if at most \(d\) mistakes can be forced at
every finite horizon, the standard optimal algorithm makes at most \(d\)
mistakes over the entire interaction. Thus, horizon-dependent guarantees
combine into a single uniform guarantee.

Language generation behaves differently. The following theorem shows that
the gap between finite-horizon and uniform mistake forcing can be
arbitrarily large.

\begin{theorem}
\label{thm:compactness-gap}
For every \(k\in\mathbb{N}\cup\{\infty\}\), there exists a countable UUS
class \(\Hcal_k\) such that
\[
    \MF_T(\Hcal_k)=1
    \qquad
    \text{for every }T\geq1.
\]
For every finite \(k\),
\[
    \MFUnif(\Hcal_k)=k,
\]
whereas
\[
    \MFUnif(\Hcal_\infty)=\infty.
\]
\end{theorem}

The construction partitions the example space into disjoint finite blocks whose sizes grow without bound. For any fixed horizon \(T\), the generator can remain inside
a sufficiently large block and make at most one mistake. An anytime
generator, however, must eventually exhaust each finite block and move to
new ones, allowing the adversary to reserve one mistaken output from each
visited block. Thus, mistakes may accumulate without bound even though the
finite-horizon mistake count is always at most one. The whole construction is provided in Appendix~\ref{app:realizable-proofs}.

\subsection{The spectrum of realizable no-regret rates}
\label{subsec:realizable-spectrum}

Theorem~\ref{thm:horizon-anytime-main} characterizes realizable no-regret
generation through the condition \(\MF_T(\Hcal)=o(T)\), but it does not
restrict how slowly or quickly this sublinear profile may grow. We now show
that the range of possible behaviors is remarkably broad. We first examine
the bounded endpoint and then construct language classes with essentially
arbitrary finite-horizon profiles.

Suppose first that \(\MF_T(\Hcal)\) remains bounded. Although the generator
attaining \(\MF_T(\Hcal)\) may depend on \(T\), the corresponding
horizon-dependent strategies can be combined into one anytime generator
whose mistakes grow as slowly as any prescribed unbounded function.

\begin{theorem}
\label{thm:slow-growth}
Suppose
$
    d:=\sup_{T\geq1}\MF_T(\Hcal)<\infty.
$
For every nondecreasing unbounded function
\(r:\mathbb{N}\to\mathbb{N}\) satisfying \(r(T)\geq1\), there exists one
anytime generator \(G\) such that
\[
    \sup_{\substack{h\in\Hcal(x_{1:T})}}
    M_T(G,h;x_{1:T})
    \leq d\,r(T)
    \qquad
    \text{for every }T\geq1.
\]
\end{theorem}

The intuition is to run fixed-horizon generators over increasingly long
phases. Each phase contributes at most \(d\) mistakes, and the phase
lengths can be chosen so that the number of phases completed by time \(T\)
grows no faster than \(r(T)\). The complete proof appears
in Appendix~\ref{app:realizable-proofs}.

The bounded case is only one endpoint of a much richer spectrum. Our next
result shows that, up to constant rescaling of the horizon, essentially
every admissible nondecreasing function can occur as a finite-horizon
mistake-forcing profile.

\begin{theorem}[Arbitrary finite-horizon profiles]
\label{thm:arbitrary-profile}
Let \(f:\mathbb{N}\to\mathbb{N}\) be nondecreasing and satisfy
\[
    1\leq f(T)\leq T
    \qquad
    \text{for every }T\geq1.
\]
There exists a countable UUS class \(\Hcal_f\) such that, for every
\(T\geq4\),
\[
    f(\lfloor T/4\rfloor)
    \leq
    \MF_T(\Hcal_f)
    \leq
    f(8T).
\]
Consequently, if \(f\) is stable under constant rescaling, in the sense
that, for every fixed \(c>0\),
\[
    f(cT)=\Theta(f(T)),
\]
where the implicit constants
may depend on \(c\), then
\[
    \MF_T(\Hcal_f)=\Theta(f(T)).
\]
\end{theorem}

The complete proof appears in Appendix~\ref{app:realizable-proofs}. The following
corollary records several familiar rates obtained from the theorem.

\begin{corollary}
\label{cor:common-profiles}
Each of the following asymptotic profiles is attained by
\(\MF_T(\Hcal)\) for some realizably no-regret generatable class \(\Hcal\):
\[
    (\log(1+T))^\beta,
    \qquad
    T^\alpha,
    \qquad
    T^\alpha(\log(1+T))^\beta,
    \qquad
    \frac{T}{(\log(1+T))^\beta},
\]
where \(\alpha\in(0,1)\) and \(\beta>0\). 
\end{corollary}

This spectrum is richer than those of several classical online-learning
problems. For example, realizable online prediction has minimax mistake rates
\(\Theta(1)\) or \(\Theta(T)\)
\cite{littlestone1988learning}; randomized realizable apple tasting has
minimax expected mistake rates
\(\Theta(1)\), \(\Theta(\sqrt{T})\), or \(\Theta(T)\)
\cite{raman2024apple}; and adversarial multi-armed bandits have minimax
regret \(\Theta(\sqrt{T})\) for a fixed number of arms
\cite{auer2002nonstochastic}. In contrast, realizable language generation
admits essentially every admissible sublinear profile, reflecting its
unusual combination of generator-chosen outputs and indirect adversarial
feedback.
\section{Agnostic No-Regret Generation}
\label{sec:agnostic-generation}

In the realizable setting, every revealed example belongs to one target
language, so the generator is evaluated by counting its invalid outputs.
In the agnostic setting, no language need contain all demonstrations.
Instead, for each language \(h\in\Hcal\), we compare the invalid outputs
of the generator with the invalid demonstrations revealed by the adversary.
This comparison is signed: invalid generated examples increase regret,
whereas invalid demonstrations decrease it.

\subsection{Signed mistake-forcing dimension}
\label{subsec:signed-mistake-forcing}

We use the same trajectory trees as in the realizable setting, but replace
the mistake pattern at each leaf by a signed pattern that records both
positive and negative contributions to regret.
\begin{definition}[Signed mistake pattern]
\label{def:signed-mistake-pattern}
Let
\[
    \tau
    =
    \bigl((\widehat{x}_1,x_1),\ldots,
          (\widehat{x}_n,x_n)\bigr)
\]
be a legal trajectory. A \emph{signed mistake pattern} for \(\tau\) is a
pair \((I,J)\) with \(I,J\subseteq[n]\), where \(I\) records the rounds on
which the generated example is invalid and \(J\) records the rounds on
which the revealed demonstration is invalid.

The subclass compatible with \((I,J)\) is
\[
    \Hcal^{\pm}(\tau;I,J)
    :=
    \left\{
        h\in\Hcal:
        \begin{array}{ll}
        h(\widehat{x}_t)=0
        &\Longleftrightarrow t\in I,\\
        h(x_t)=0
        &\Longleftrightarrow t\in J
        \end{array}
        \quad
        \text{for every }t\in[n]
    \right\}.
\]
We call \((I,J)\) \emph{feasible} if
\(\Hcal^{\pm}(\tau;I,J)\neq\varnothing\), and write
\[
    S_{\Hcal}^{\pm}(\tau)
    :=
    \left\{
        (I,J):
        I,J\subseteq[n]
        \text{ and }
        \Hcal^{\pm}(\tau;I,J)\neq\varnothing
    \right\}
\]
for the set of feasible signed mistake patterns. In particular,
\[
    S_{\Hcal}^{\pm}(\tau_\varnothing)
    =
    \{(\varnothing,\varnothing)\}.
\]
The signed score of a pattern \((I,J)\) is
\[
    |I|-|J|.
\]
\end{definition}

\begin{definition}[Signed-bad trajectory]
\label{def:signed-bad-trajectory}
For a legal trajectory \(\tau\), define
\[
    \rho_{\Hcal}(\tau)
    :=
    \max_{(I,J)\in S_{\Hcal}^{\pm}(\tau)}
    \bigl(|I|-|J|\bigr).
\]
For \(m\in\mathbb{N}_0\), the trajectory \(\tau\) is
\emph{\(m\)-signed-bad for \(\Hcal\)} if
\[
    \rho_{\Hcal}(\tau)\geq m.
\]
\end{definition}

The realizable mistake patterns from
Definition~\ref{def:mistake-pattern} are recovered by setting
\(J=\varnothing\). More precisely,
\[
    S_{\Hcal}(\tau)
    =
    \left\{
        I\subseteq[n]:
        (I,\varnothing)\in S_{\Hcal}^{\pm}(\tau)
    \right\}.
\]
For such patterns, the signed score \(|I|-|J|\) reduces to the ordinary
mistake count \(|I|\).

\begin{definition}[Signed mistake-forcing dimension]
\label{def:signed-mistake-forcing-dimension}
Let \(m,T\in\mathbb{N}_0\). An
\emph{\(m\)-signed-mistake-forcing tree of horizon \(T\)} for \(\Hcal\)
is a generator-complete trajectory tree of horizon \(T\), rooted at
\(\tau_\varnothing\), whose leaves are all \(m\)-signed-bad for \(\Hcal\).

The \emph{horizon-\(T\) signed mistake-forcing dimension} of \(\Hcal\) is
\[
    \SMF_T(\Hcal)
    :=
    \max
    \left\{
        m\in\{0,\ldots,T\}:
        \begin{array}{c}
        \text{there exists an \(m\)-signed-mistake-forcing tree}\\
        \text{of horizon \(T\) for \(\Hcal\)}
        \end{array}
    \right\}.
\]
\end{definition}

Unlike in the realizable setting, we do not introduce a well-founded analogue of the signed mistake-forcing dimension. Under realizability, every revealed example is valid, so the generator's cumulative number of mistakes is nondecreasing: once a leaf certifies \(m\) mistakes, later rounds cannot invalidate that certificate. Agnostic regret, however, is the difference between the generator's mistakes and the comparator's mistakes on the revealed examples. Its value may increase or decrease from one round to the next, so it may not converge in the end. A common finite horizon is therefore essential to the signed-tree characterization.

\subsection{Characterizations of agnostic no-regret generation}
\label{subsec:agnostic-characterizations}
Similarly as in the realizable setting, we next show that \(\SMF_T(\Hcal)\) is exactly the optimal worst-case
agnostic regret at horizon \(T\), and the entire profile
\((\SMF_T(\Hcal))_{T\geq1}\) can be used to characterize agnostic no-regret
generation.

\begin{theorem}
\label{thm:smf-characterization}
For every nonempty UUS class \(\Hcal\) and every \(T\geq1\),
\[
    \SMF_T(\Hcal)
    =
    \min_{G_T}
    \sup_{\substack{x_{1:T}\\ h\in\Hcal}}
    \Reg_T(G,h;x_{1:T}).
\]

For each \(T\), the minimum is attained by a generator \(G_T\), which may
depend on \(T\).
\end{theorem}

The intuition is the same as in the realizable characterization. The
complete proof appears in Appendix~\ref{app:agnostic-proofs}.

The preceding theorem allows the optimal generator to depend on \(T\), but
agnostic no-regret generation requires one anytime generator. As in the
realizable setting, this distinction does not create a gap.

\begin{theorem}[Characterization of agnostic no-regret generation]
\label{thm:agnostic-horizon-anytime-main}
For every nonempty UUS class \(\Hcal\), the following are equivalent:
\begin{enumerate}
    \item \(\Hcal\) is agnostically no-regret generatable;
    \item \(\SMF_T(\Hcal)=o(T)\) as \(T\to\infty\).
\end{enumerate}
Moreover, let \(b:\mathbb N\to[1,\infty)\) be nondecreasing and
unbounded. If
\[
    \SMF_T(\Hcal)\leq b(T)
    \qquad
    \text{for every }T\geq1,
\]
then there exists one deterministic anytime generator \(G\) such that,
simultaneously for every \(T\geq1\),
\[
    \sup_{\substack{x_{1:T}\\h\in\Hcal}}
    \Reg_T(G,h;x_{1:T})
    \leq 4b(T).
\]
\end{theorem}

The proof combines horizon-dependent generators over a sequence of
doubling epochs. A new issue arises because agnostic regret increments may
be negative, so the regret incurred during one step cannot be bounded by
simply viewing that step as part of a longer run. The proof resolves this
using a virtual simulation in which skipped rounds are assigned
\(x_t=\widehat{x}_t\), producing zero signed increment for every language.
The complete construction appears in Appendix~\ref{app:agnostic-proofs}.

\subsection{Realizable versus agnostic generation}
\label{subsec:realizable-versus-agnostic}

Every realizable play is a special case of an agnostic play. This relation
is also reflected by the two forcing dimensions.

\begin{proposition}
\label{prop:mf-le-smf}
\label{prop:necessary}
For every nonempty UUS class \(\Hcal\) and every \(T\geq1\),
\[
    \MF_T(\Hcal)\leq\SMF_T(\Hcal).
\]
\end{proposition}

Indeed, if \(h\) realizes the demonstrations, then \(h(x_t)=1\) on every
round. Its negative-regret set is empty, and its signed score equals the
number of generated examples outside \(h\). Consequently, agnostic
no-regret generation implies realizable no-regret generation.

Finite classes admit both realizable and agnostic guarantees, as shown later
in Theorem~\ref{thm:finite-class-bounds}. For infinite classes, the converse
implication from realizable to agnostic
no-regret generation can fail completely. We now give a class with a
constant realizable mistake bound but linear agnostic regret.

\begin{theorem}
\label{thm:pair-separation}
There exists a countable UUS class \(\Hcal_{\mathrm{pair}}\) satisfies
\[
    \MFUnif(\Hcal_{\mathrm{pair}})\leq2,
\]
whereas
\[
    \SMF_T(\Hcal_{\mathrm{pair}})
    \geq
    \frac{T}{3}
    \qquad
    \text{for every }T\geq1.
\]
\end{theorem}

The witness class is designed so that, in the realizable game, two positive
demonstrations identify infinitely many safe outputs.  In the agnostic game,
however, a comparator chosen after the trajectory can reject every generated
example while accepting enough demonstrations to yield regret at least
\(T/3\).  Thus, a class may be realizably mistake-bounded but not
agnostically no-regret generatable.  The construction and proof appear in
Appendix~\ref{app:agnostic-proofs}.

\subsection{The spectrum of agnostic no-regret rates}
\label{subsec:agnostic-spectrum}

Theorem~\ref{thm:agnostic-horizon-anytime-main} reduces agnostic no-regret
generation to the growth of \(\SMF_T(\Hcal)\). As in the realizable
setting, the possible sublinear profiles are remarkably rich. Indeed, the
same block construction used for realizable generation produces essentially
arbitrary agnostic profiles.

\begin{theorem}
\label{thm:arbitrary-agnostic-profile}
Let \(f:\mathbb{N}\to\mathbb{N}\) be nondecreasing and satisfy
\[
    1\leq f(T)\leq T
    \qquad
    \text{for every }T\geq1.
\]
There exists a countable UUS class \(\Hcal_f\) such that, for every
\(T\geq4\),
\[
    f(\lfloor T/4\rfloor)
    \leq
    \MF_T(\Hcal_f)
    \leq
    \SMF_T(\Hcal_f)
    \leq
    f(8T).
\]
Consequently, if \(f\) is stable under constant rescaling, then
\[
    \SMF_T(\Hcal_f)=\Theta(f(T)).
\]
\end{theorem}

The lower bound follows from
Proposition~\ref{prop:mf-le-smf} and the realizable profile of
\(\Hcal_f\). The upper bound uses the same block structure to control the
generator's loss and the comparator's demonstration loss simultaneously.
The complete construction appears in Appendix~\ref{app:agnostic-proofs}.

Similarly as in the realizable setting, there is a wide range of rates for a
single anytime generator.

\begin{corollary}
\label{cor:arbitrary-anytime-agnostic-rates}
Each of the following asymptotic profiles is attained by
\(\SMF_T(\Hcal)\) for some agnostically no-regret generatable class \(\Hcal\):
\[
    (\log(1+T))^\beta,
    \qquad
    T^\alpha,
    \qquad
    T^\alpha(\log(1+T))^\beta,
    \qquad
    \frac{T}{(\log(1+T))^\beta},
\]
where \(\alpha\in(0,1)\) and \(\beta>0\). 
\end{corollary}

The results in Section~\ref{subsec:realizable-versus-agnostic} and \ref{subsec:agnostic-spectrum} reveal two sharply different relationships between realizable
and agnostic generation. For the block classes \(\Hcal_f\), the realizable
and agnostic profiles have the same asymptotic order, even across essentially
arbitrary rates. For the class \(\Hcal_{\mathrm{pair}}\), by contrast, the
realizable mistake bound is constant while the agnostic regret is linear.
Thus, the cost of moving from realizable to agnostic generation can range
from no asymptotic loss to the largest possible separation.

\section{Comparison with Other Generation Criteria}
\label{sec:comparison-generation-criteria}

\subsection{Comparison of realizable generation criteria}
\label{subsec:comparison-realizable-criteria}


Existing realizable generation criteria differ along two axes. First, they control different aspects of mistakes: the time of the last mistake, the total number of mistakes, or their asymptotic frequency. Second, they differ in whether the guarantee holds uniformly or separately for different targets and demonstration sequences. We present the different definitions in Figure~\ref{fig:generation-criteria} and leave the formal definition in Appendix~\ref{app:comparison-criteria-definitions}.

\begin{figure}[t]
\centering
\begin{tikzpicture}[
    criterion/.style={
        align=center,
        font=\small,
        inner sep=2pt
    },
    implication/.style={
        -{Stealth[length=2mm]},
        semithick
    }
]
\node[criterion] (uniform) at (0,4.5)
    {Uniform generation\\
     \cite{kleinberg2024language, li2024generationlenslearningtheory}};

\node[criterion] (mistake) at (-4.6,2.5)
    {Uniform mistake-bounded generation\\
     \cite{kleinberg2026mistakebounded}};

\node[criterion] (nonuniform) at (4.6,2.5)
    {Non-uniform generation\\
     \cite{li2024generationlenslearningtheory}};

\node[criterion] (noregret) at (-4.6,0.5)
    {Realizable no-regret generation\\
     (Our Work)};

\node[criterion] (limit) at (4.6,0.5)
    {Generation in the limit\\
     \cite{kleinberg2024language}};

\node[criterion] (zerorate) at (0,-1.7)
    {Zero-rate generation in the limit\\
     \cite{panigrahi2026hallucinationrates}};

\draw[implication] (uniform) -- (mistake);
\draw[implication] (uniform) -- (nonuniform);
\draw[implication] (mistake) -- (noregret);
\draw[implication] (nonuniform) -- (limit);
\draw[implication] (mistake) -- (limit);
\draw[implication] (noregret) -- (zerorate);
\draw[implication] (limit) -- (zerorate);
\end{tikzpicture}

\caption{Implications among generation criteria. An arrow denotes implication.
Every displayed implication is strict, and no additional implication holds
except by transitivity.}
\label{fig:generation-criteria}
\end{figure}
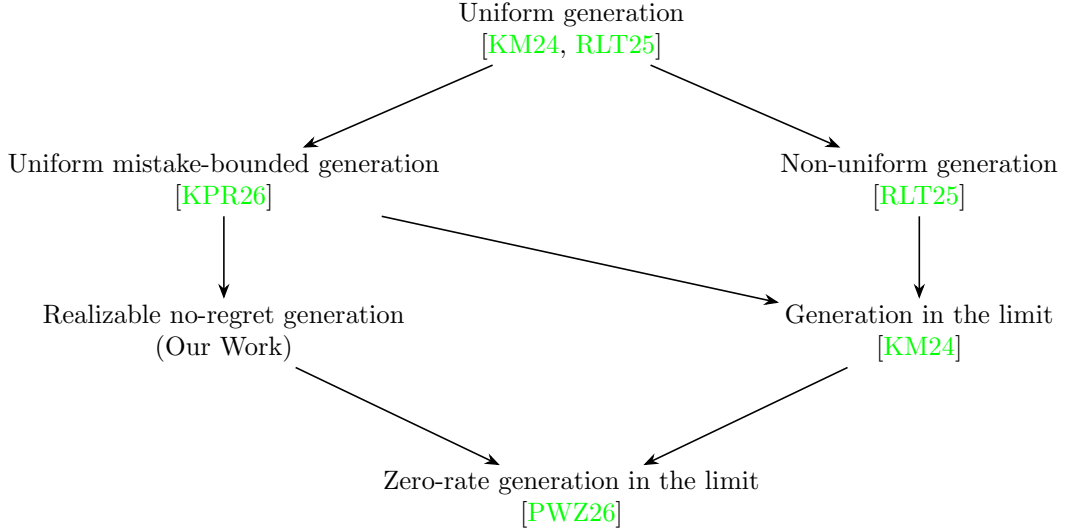

\begin{theorem}[Informal]
\label{thm:comparison-realizable-criteria}
For every nonempty UUS class, the implication diagram in
Figure~\ref{fig:generation-criteria} is complete: every arrow represents a
strict implication, and two criteria are incomparable whenever neither can
be reached from the other by a directed path.
\end{theorem}

The strictness statements are witnessed by a small collection of examples
developed earlier in the paper and in the appendix. In Appendix~\ref{app:comparison-proofs}, we formally present the theorem, and justify all strictness and incomparability statements.

\subsection{Noisy generation versus agnostic generation}
\label{subsec:noisy-versus-agnostic}

We clarify that noisy generation and agnostic generation relax realizability in different
ways. A noisy model retains a target language and constrains the
demonstrations that fall outside it. The agnostic model imposes no such
target or constraint and instead compares the generator with every
\(h\in\Hcal\). Nevertheless, the regret identity gives a direct transfer from an agnostic
guarantee to any contamination bound in the noisy setting.

\begin{proposition}
\label{prop:agnostic-contamination-transfer}
Suppose a deterministic generator \(G\) and a function
\(r:\mathbb N\to[0,\infty)\) satisfy
\[
    \sup_{\substack{x_{1:T}\\h\in\Hcal}}
    \Reg_T(G,h;x_{1:T})
    \leq r(T)
    \qquad
    \text{for every }T\geq1.
\]
Then, for every \(h\in\Hcal\) and every demonstration sequence,
\[
    M_T(G,h;x_{1:T})
    \leq
    r(T)+\sum_{t=1}^T\ell_h(x_t).
\]
Consequently, if \(r(T)=o(T)\) and the demonstrations contain \(o(T)\)
invalid examples relative to a target \(h^\star\), then the generator
produces \(o(T)\) invalid examples relative to the same target.
\end{proposition}

The proof of the transfer statement appears in
Appendix~\ref{app:comparison-proofs}.

\subsection{Finite and countable classes}
\label{subsec:finite-countable}

Previous work \cite{kleinberg2024language,li2024generationlenslearningtheory}
established that every finite UUS hypothesis class is uniformly generatable
and every countable UUS hypothesis class is non-uniformly generatable; see
Appendix~\ref{app:comparison-criteria-definitions} for the formal definitions.
We now compare their quantitative guarantees. Finite classes admit a
uniform bound on total realizable mistakes and a sublinear agnostic regret
bound. Countability alone, however, does not guarantee sublinear worst-case
mistakes or regret: a countable class can have both profiles equal to \(T\)
at every horizon \(T\).

We first record the quantitative guarantees available for finite classes.
Both bounds follow from results of
\cite{kleinberg2026mistakebounded}: the realizable bound from their
Theorem~5.1 and the agnostic bound from their Lemma~7.1.

\begin{theorem}[Finite-class bounds]
\label{thm:finite-class-bounds}
If \(1\leq|\Hcal|=N<\infty\), then
\[
    \MFUnif(\Hcal)
    \leq
    \lfloor\log N\rfloor
\]
and
\[
    \SMF_T(\Hcal)
    \leq
    \sqrt{8T\log N}
    \qquad
    \text{for every }T\geq1.
\]
\end{theorem}

Theorem~\ref{thm:finite-class-bounds} shows that every finite nonempty UUS
class is uniformly mistake-bounded in the realizable setting and agnostically
no-regret generatable. 
The situation changes sharply for countably infinite classes. Every
countable UUS class is non-uniformly generatable
\cite{li2024generationlenslearningtheory}: there exists one generator such
that, for each fixed target \(h^\star\), all sufficiently late outputs are
valid for \(h^\star\). This is a target-by-target guarantee, however. The
last-mistake bound may depend arbitrarily on \(h^\star\), and countability
alone gives no uniform control of the finite-horizon mistake profile. 
The following proposition makes this limitation explicit. 
\begin{proposition}
\label{prop:cofinite-linear-profile}
There exists a class \(\Hcal_{\mathrm{cof}}\) which is countable, UUS, and non-uniformly
generatable. Nevertheless,
\[
    \MF_T(\Hcal_{\mathrm{cof}})
    =
    \SMF_T(\Hcal_{\mathrm{cof}})
    =T
    \qquad
    \text{for every }T\geq1.
\]
Consequently, \(\Hcal_{\mathrm{cof}}\) is neither realizably nor
agnostically no-regret generatable.
\end{proposition}

The complete proof appears in
Appendix~\ref{app:comparison-proofs}.

\section{From Prediction to Generation}
\label{sec:prediction-to-generation}

Previous work \cite{li2024generationlenslearningtheory} has showed that online learnability and generatability are
incomparable as qualitative properties. We show that a quantitative
comparison nevertheless emerges when generation is measured through our
setting. Using Littlestone dimension as the prediction
complexity and our mistake-forcing dimensions as the corresponding
generation complexities, we establish explicit prediction-to-generation
bounds in both the realizable and agnostic settings. We further show that the
realizable bound we derived is optimal.
\subsection{Online prediction and Littlestone dimension}
\label{subsec:prediction-preliminaries}

We briefly recall online prediction and Littlestone
dimension. In binary online prediction, at each round \(t\), the learner
observes an example \(x_t\), predicts a label
\(\widehat y_t\in\{0,1\}\), and then observes the true label \(y_t\).
A mistake occurs when \(\widehat y_t\neq y_t\). In the realizable setting,
the labels satisfy
\[
    y_t=h^\star(x_t)
\]
for some fixed target \(h^\star\in\Hcal\). In the agnostic setting, the
labels are arbitrary, and the learner is evaluated by its expected excess
number of mistakes relative to the best hypothesis in \(\Hcal\).

\begin{definition}[Littlestone dimension]
A \emph{binary instance tree of depth \(d\)} is a complete binary tree whose
internal nodes are labeled by examples from \(\Xcal\) and whose two outgoing
edges are labeled by \(0\) and \(1\). The tree is \emph{shattered} by
\(\Hcal\) if every root-to-leaf sequence of edge labels is realized by some
\(h\in\Hcal\) on the corresponding node labels.

The \emph{Littlestone dimension} \(\Ldim(\Hcal)\) is the largest depth of a
binary instance tree shattered by \(\Hcal\), or infinity if such trees exist
at every finite depth.
\end{definition}

Classical results show that Littlestone dimension characterizes both
realizable and agnostic online prediction
\cite{littlestone1988learning,ben2009agnostic}. The optimal uniform number of
realizable prediction mistakes is exactly \(\Ldim(\Hcal)\). Moreover,
agnostic online prediction admits sublinear regret if and only if
\(\Ldim(\Hcal)<\infty\). In particular, if
\(d:=\Ldim(\Hcal)<\infty\), then its minimax regret is at most
\[
    O\left(\sqrt{dT\log(1+T)}\right).
\]

Surprisingly, although generation provides different feedback,
we show that Littlestone dimension also controls the generation
complexities \(\MF_T(\Hcal)\) and \(\SMF_T(\Hcal)\).

\subsection{Upper bounds from finite Littlestone dimension}
We begin with realizable generation. Finite Littlestone dimension limits
the number of distinct membership patterns that the adversary can maintain
along a trajectory. This yields the following finite-horizon bound.
\begin{theorem}
\label{thm:finite-ldim-realizable}
Let \(\Hcal\) be nonempty and UUS, and suppose
\[
0<d:=\Ldim(\Hcal)<\infty.
\]
Then for every \(T\),
\[
\MF_T(\Hcal)
  \leq
  \left\lfloor
  \log\left(
    \sum_{i=0}^{\min\{d,2T\}}\binom{2T}{i}
  \right)
  \right\rfloor.
\]
Consequently, for fixed \(d\),
\[
\MF_T(\Hcal)=O\left(d\log \left(1+\frac{T}{d}\right)\right)=o(T).
\]
Moreover, there exists one deterministic anytime generator \(G\) such that,
simultaneously for every \(T\geq1\),
\[
    \sup_{\substack{x_{1:T}\\h\in\Hcal(x_{1:T})}}
    M_T(G,h;x_{1:T})
    =
    O\left(
        d\log\left(1+\frac{T}{d}\right)
    \right).
\]
\end{theorem}
The intuition is that forcing \(m\) mistakes requires the class to support
at least \(2^m\) distinguishable membership patterns along an associated
binary tree. Finite Littlestone dimension bounds the number of such patterns,
and taking the logarithm gives the stated mistake bound. The complete proof appears in
Appendix~\ref{app:finite-ldim-upper}.
A similar conclusion holds in the agnostic setting. Here the language may
also make mistakes on the adversary's demonstrations, so the argument must
control the difference between the two mistake counts. A weighted version of
the realizable potential gives the following result.
\begin{theorem}
\label{thm:finite-ldim-agnostic}
Let \(\Hcal\) be nonempty and UUS, and suppose
\[
0<d:=\Ldim(\Hcal)<\infty.
\]
For every \(T\) and every \(\eta\in(0,1/2]\), there is a deterministic generator
such that
\[
\SMF_T(\Hcal)
  \leq
  \frac{1}{\eta}
  \log\left(
    \sum_{i=0}^{\min\{d,2T\}}\binom{2T}{i}
  \right)
  +\eta T.
\]
Consequently, for fixed \(d\),
\[
\SMF_T(\Hcal)
  =O\left(\sqrt{dT\log\left(1+\frac{T}{d}\right)}
  \right)
  =o(T).
\]
Moreover, there exists one deterministic anytime generator \(G\) such that,
simultaneously for every \(T\geq1\),
\[
    \sup_{\substack{x_{1:T}\\h\in\Hcal}}
    \Reg_T(G,h;x_{1:T})
    =
    O\left(
        \sqrt{
            dT\log\left(1+\frac{T}{d}\right)
        }
    \right).
\]
\end{theorem}
The two terms in the bound reflect a familiar tradeoff as in Lemma~7.1 in \cite{kleinberg2026mistakebounded}. A smaller value of
\(\eta\) reduces the cost accumulated over the \(T\) rounds but increases the
cost of maintaining the compatible membership patterns. Balancing the two
terms gives the displayed sublinear rate. The complete proof appears in
Appendix~\ref{app:finite-ldim-upper}.
Together with the characterizations in
Theorems~\ref{thm:horizon-anytime-main}
and~\ref{thm:agnostic-horizon-anytime-main}, these results show that every
UUS class of finite Littlestone dimension is both realizably and agnostically
no-regret generatable.

We emphasize that the finite-Littlestone assumption in Theorems~\ref{thm:finite-ldim-realizable} and~\ref{thm:finite-ldim-agnostic} cannot be replaced by finite VC dimension. Indeed, there exists a countable UUS class \(\mathcal H\) such that
$$
\operatorname{VCdim}(\mathcal H)=1,
\qquad
\operatorname{MF}_T(\mathcal H)
=
\operatorname{SMF}_T(\mathcal H)
=T
\quad\text{for every }T\ge1.
$$
Thus even VC dimension one permits linear generation regret. Appendix~\ref{app:finite-vc} gives the construction and proof.

\subsection{Sharpness in the realizable setting}
The next theorem
shows that the preceding realizable upper bound captures the optimal
dependence on both parameters.
\begin{theorem}[Sharpness of the realizable finite-Littlestone bound]
\label{thm:sharp-ldim-generation}
For every integer \(d\geq2\), there exists a countable UUS class
\(\Hcal_d^{\mathrm{sharp}}\) satisfying
\[
\Ldim(\Hcal_d^{\mathrm{sharp}})=d
\]
such that, for every \(T\geq3\),
\[
\MF_T(\Hcal_d^{\mathrm{sharp}})
    =
    \Theta\left(
        d\log\left(1+\frac{T}{d}\right)
    \right),
\]
where the implicit constants are universal.
\end{theorem}
The construction combines several independent finite blocks. Within each
block, the adversary can repeatedly force the generator to distinguish
among many candidate languages, producing logarithmically many mistakes.
Combining \(d\) such components makes these costs accumulate while keeping
the Littlestone dimension equal to \(d\). The complete construction and a
more precise quantitative statement appear in
Appendix~\ref{app:finite-ldim-sharpness}.
Thus the realizable upper bound is optimal, up to universal constants, as a
function of \(d\) and \(T\). 
For agnostic generation, we do not presently have a corresponding sharp
lower bound for fixed finite Littlestone dimension. Determining the optimal
dependence of \(\SMF_T(\Hcal)\) on \(d\) and \(T\) remains an open problem.

\section{Conclusion and Future Work}
\label{sec:discussion}

Our results separate the number of mistakes a generator makes from the time at which it stops making them. A class may admit at most one mistake at every fixed horizon while admitting no finite bound on total mistakes for any single anytime generator. Nevertheless, sublinear finite-horizon mistake bounds suffice for anytime no-regret generation. The agnostic setting introduces a further distinction: even a constant realizable mistake bound need not imply sublinear agnostic regret. These separations show why qualitative generation guarantees alone do not determine quantitative performance.

Several questions remain open. First, the trajectory-tree characterizations give exact minimax values, but evaluating them for concrete hypothesis classes may be difficult. Second, when having a class $\Hcal$ with Littlestone dimension $d$, the optimal bound of $\SMF_T(\Hcal)$ on $d$ and $T$ remains open.  Finally, the present results impose no computational constraints; understanding when the minimax guarantees can be achieved efficiently is a separate challenge.

\section*{AI Disclosure}
The main technical ideas underlying this work emerged through an extensive, iterative process involving the authors and ChatGPT 5.6 Sol. ChatGPT 5.6 Sol also assisted with writing the paper. The authors take full responsibility for all content in the paper.

\bibliographystyle{alpha}
\bibliography{yourbibfile}
\appendix

\section{Proofs for Section~\ref{sec:realizable-generation}}
\label{app:realizable-proofs}
\subsection{Proofs for Section~\ref{subsec:realizable-characterizations}}

\begin{theorem*}[Restatement of Theorem~\ref{thm:mf-finite-characterization}]
For every nonempty UUS class \(\Hcal\) and every \(T\geq1\),
\[
  \MF_T(\Hcal)
  =
  \min_{G_T}
  \sup_{\substack{x_{1:T}\\h\in\Hcal(x_{1:T})}}
  M_T(G_T,h;x_{1:T}).
\]
For each \(T\), the minimum is attained by a generator \(G_T\), which may
depend on \(T\).
\end{theorem*}

\begin{proof}
We first prove the lower bound.  Let \(m\le \MF_T(\Hcal)\), fix a corresponding \(m\)-mistake-forcing tree rooted at
\(\tau_\varnothing\), and fix an arbitrary horizon-\(T\) generator \(G_T\).
Suppose that the interaction has reached a node \(u\).  Generator-completeness supplies at least
one outgoing edge with precisely this output label.  The adversary selects
one such edge and reveals its response label.  Repeating this procedure
reaches a leaf after exactly \(T\) rounds.

Let \(\tau\) be the resulting trajectory at that leaf.  Since the leaf is \(m\)-bad,
there is an \(I\in S_{\Hcal}(\tau)\) with \(|I|\geq m\).  Feasibility of
\(I\) means that \(\Hcal(\tau;I)\) is nonempty, so we can choose
\(h\in\Hcal(\tau;I)\).  Consequently,
\[
  M_T(G_T,h;x_{1:T})
  =
  |I|
  \geq m.
\]
This proves the desiring lower bound.

For the reverse inequality, put \(d:=\MF_T(\Hcal)\).  Say that a legal
trajectory \(\tau\) of length at most \(T\) is \emph{forcing} if it roots a
generator-complete trajectory tree whose leaves have total trajectory
length \(T\) and are all \((d+1)\)-bad.  If \(\tau\) has
length \(s\), the attached tree has \(T-s\) further rounds.  The empty
trajectory \(\tau_\varnothing\) is not forcing, since otherwise it would
root a horizon-\(T\) tree forcing \(d+1\) mistakes, contrary to the
definition of \(d\).

At every non-forcing trajectory $\tau$ of length less than \(T\), we claim there is a legal
output \(\widehat x\) such that every legal response produces a non-forcing
successor.  Indeed, suppose that no such output existed.  Then, for each
legal output \(\widehat x\), one could choose a legal response \(x\) for
which \(\tau\circ(\widehat x,x)\) is forcing.  Attach one forcing
subtree below the edge labeled by \(\widehat x\) and response \(x\).  The
result is generator-complete, all its leaves have total length \(T\), and
all those leaves are \((d+1)\)-bad.  It would therefore make \(\tau\)
forcing, a contradiction.

We now define \(G_T\) recursively.  On each demonstration history reached
before time \(T\),  choose one of the safe outputs that is non-forcing.  This output
is legal, and every legal adversarial response preserves the non-forcing
invariant.  

Now fix a length-\(T\) realizable play with target
\(h^\star\in\Hcal(x_{1:T})\).  Every trajectory reached by \(G_T\) is
non-forcing, so the terminal trajectory \(\tau\) is not
\((d+1)\)-bad.  Moreover, \(h^\star\) witnesses the actual mistake pattern
\[
  I_{h^\star}
  :=
  \{t\in[T]:h^\star(\widehat{x}_t)=0\}
  \in
  S_{\Hcal}(\tau).
\]
Since \(\tau\) is not \((d+1)\)-bad, we have
\[
  M_T(G_T,h^\star;x_{1:T})
  =
  |I_{h^\star}|
  \leq
  d.
\]
Thus \(G_T\) has worst-case realizable loss at most \(d\), which proves the
reverse inequality.  Together with the lower bound, this also shows that
the displayed minimum is attained by \(G_T\).
\end{proof}

\begin{theorem*}[Restatement of Theorem~\ref{thm:mf-infinite-characterization}]
For every nonempty UUS class \(\Hcal\),
\[
  \MFUnif(\Hcal)
  =
  \min_G
  \sup_{\substack{x_{1:\infty}\\h\in\Hcal(x_{1:\infty})}}
  M_\infty(G,h;x_{1:\infty}).
\]
The minimum is taken over deterministic anytime generators.
\end{theorem*}

\paragraph{Proof idea.}
We use the same safe-move strategy as in the finite-horizon proof. The only difference is that there is no common deadline. A well-founded forcing tree may stop at a different finite time on each interaction path. At every history, the generator chooses an output such that, regardless of the next demonstration, the adversary cannot still force \(d+1\) mistakes before the interaction eventually stops. If no such output existed, the corresponding forcing subtrees could be joined to form a well-founded tree forcing \(d+1\) mistakes from the current history.

\begin{proof}
We begin with the lower bound.  Suppose that a well-founded tree forces
\(m\) mistakes, and fix an anytime generator \(G\).  At each node, the
demonstration history determines a legal output of \(G\), and
generator-completeness provides an edge carrying that output.  Let the
adversary follow one such edge.  This process cannot produce an infinite
branch, so well-foundedness ensures that it reaches a leaf after some
finite number of rounds.

At the leaf, choose a feasible mistake pattern \(I\) with \(|I|\geq m\) and
a witness \(h\in\Hcal(\tau;I)\).  The forcing tree gives only a finite adversarial path, but we can extend it to an infinite realizable sequence: Because \(\supp(h)\) is infinite while the current prefix is finite, there is always another positive example that has not appeared before. Repeatedly choosing such examples produces an infinite sequence of distinct positive demonstrations.

The generator has already made at least \(|I|\geq m\) mistakes along the original finite path. Therefore,

$$
M_\infty(G,h;x_{1:\infty})\geq |I|\geq m.
$$

For the reverse inequality, assume \(d:=\MFUnif(\Hcal)<\infty\).  We use the same trick as in the proceeding proof here. Call a
legal trajectory forcing if it roots a well-founded generator-complete tree
whose leaves are all \((d+1)\)-bad.  The empty trajectory is non-forcing by
the definition of \(d\).

We claim that every non-forcing trajectory admits a legal output for which
every legal response-successor remains non-forcing.  Otherwise, for each
legal output, choose one response leading to a forcing successor, and attach
one witnessing well-founded subtree below that edge.  This union is
generator-complete.  It is also well-founded. Indeed, after leaving the new root, every branch remains inside one attached subtree. Since each attached subtree has no infinite branch, the resulting tree has no infinite branch either.  Since
all leaves are \((d+1)\)-bad, the union would make the current trajectory
forcing, a contradiction.

Define \(G\) recursively to choose such a safe output on every finite
demonstration history.  As in the finite-horizon proof, every adversarial
response preserves the non-forcing invariant.  A \((d+1)\)-bad trajectory
would itself be a one-node well-founded forcing tree, so every finite
trajectory reached by \(G\) is not \((d+1)\)-bad.

Now fix an infinite realizable play with target
\(h^\star\in\Hcal(x_{1:\infty})\).  At every finite time \(T\),
\(h^\star\) witnesses the actual mistake pattern, so the fact that the
length-\(T\) trajectory is not \((d+1)\)-bad gives
\(M_T(G,h^\star;x_{1:T})\leq d\).  Hence
\[
  M_\infty(G,h^\star;x_{1:\infty})
  =
  \sup_{T\geq1}M_T(G,h^\star;x_{1:T})
  \leq d.
\]
This proves the upper bound and its attainment by an anytime generator.  If
\(\MFUnif(\Hcal)=\infty\), the lower-bound argument already shows that every
generator has infinite worst-case total mistake count, so the equality also
holds in that case.
\end{proof}

We establish the regularity of $\MF_T$ as a useful tool in proving the next theorem.

\begin{lemma}[Regularity of the realizable profile]
\label{lem:app-realizable-profile-regularity}
For every nonempty UUS class and every \(T\geq1\),
\[
  \MF_T(\Hcal)
  \leq
  \MF_{T+1}(\Hcal)
  \leq
  \MF_T(\Hcal)+1.
\]
\end{lemma}

\begin{proof}
Fix an arbitrary horizon-\((T+1)\) generator and restrict its decision rules
to histories of length less than \(T\).  By
Theorem~\ref{thm:mf-finite-characterization}, there are demonstrations
\(x_{1:T}\) and \(h\in\Hcal(x_{1:T})\) on which this restriction makes at
least \(\MF_T(\Hcal)\) mistakes.  UUS supplies a fresh
\(x_{T+1}\in\supp(h)\), so the same \(h\) realizes the extended play.  The
nonnegative loss on round \(T+1\) cannot erase an earlier mistake, thus
\(\MF_{T+1}(\Hcal)\geq\MF_T(\Hcal)\).

For the second inequality, take a horizon-\(T\) generator attaining
\(\MF_T(\Hcal)\).
Use its decision rules for the first \(T\) rounds and choose any legal output
on round \(T+1\).  
On every length-\((T+1)\) realizable play, the first \(T\) rounds contribute
at most \(\MF_T(\Hcal)\) mistakes and the last round contributes at most one.
Applying the characterization at horizon \(T+1\) yields
\(\MF_{T+1}(\Hcal)\leq\MF_T(\Hcal)+1\).
\end{proof}

\begin{theorem*}[Restatement of Theorem~\ref{thm:horizon-anytime-main}]
For every nonempty UUS class \(\Hcal\), the following are equivalent:
\begin{enumerate}
  \item \(\Hcal\) is realizably no-regret generatable;
  \item \(\MF_T(\Hcal)=o(T)\) as \(T\to\infty\).
\end{enumerate}
Moreover, let \(b:\mathbb N\to[1,\infty)\) be nondecreasing and unbounded.
If
\[
  \MF_T(\Hcal)\leq b(T)
  \qquad
  \text{for every }T\geq1,
\]
then there exists one deterministic anytime generator \(G\) such that,
simultaneously for every \(T\geq1\),
\[
  \sup_{\substack{x_{1:T}\\h\in\Hcal(x_{1:T})}}
  M_T(G,h;x_{1:T})
  \leq4b(T).
\]
\end{theorem*}

\paragraph{Proof idea.}
Think of \(b(T)\) as the mistake allowance available by time \(T\). We divide the whole generation process into epochs. During epoch \(j\), we follow a finite-horizon generator whose total mistake bound is \(2^j\), choosing the epoch so that \(b(T)\) is already at this scale. Since the budgets double across epochs, the total cost of all completed epochs is dominated by the most recent one. Thus, combining the horizon-dependent generators into one anytime generator increases the mistake bound by only a constant factor.

\begin{proof}
  Suppose first that one anytime generator \(G\) has worst-case realizable
  mistake count \(o(T)\).  Restricting \(G\) to its first \(T\) decision
  rules gives a horizon-\(T\) generator, so
  Theorem~\ref{thm:mf-finite-characterization} implies that
  \(\MF_T(\Hcal)\) is at most this worst-case count.  Hence
  \(\MF_T(\Hcal)=o(T)\).  

We next prove the quantitative assertion.  Let
\(b:\mathbb N\to[1,\infty)\) be nondecreasing and unbounded, with
\(\MF_T(\Hcal)\leq b(T)\) for every \(T\).  Put
\[
  j_0:=\left\lceil\log b(1)\right\rceil.
\]
and, for every \(j\geq j_0\), define
\[
  N_j:=\max\{T\in\mathbb N:b(T)\leq2^j\}.
\]

Each \(N_j\) is
well-defined because \(b\) is unbounded and nondecreasing.  Moreover, from our definition, the sequence \((N_j)_{j\geq j_0}\) is also unbounded and nondecreasing.

For each \(j\), Theorem~\ref{thm:mf-finite-characterization} supplies a
horizon-\(N_j\) generator \(G_j\) whose worst-case realizable mistake count
is
\[
  \MF_{N_j}(\Hcal)
  \leq
  b(N_j)
  \leq
  2^j.
\]

Define the anytime generator \(G\) as follows. On round \(t\), let
\[
j(t):=\min\{j\geq j_0:t\leq N_j\},
\]and set
\[
G(x_{1:t-1}):=G_{j(t)}(x_{1:t-1}).
\]The index \(j(t)\) exists because \(N_j\to\infty\). We show $G$ has the desired upper bound for its mistakes.

For any \(T\), demonstration sequence \(x_{1:T}\), and target
\(h^\star\in\Hcal(x_{1:T})\), let \(J\) be the least index with
\(T\leq N_J\). We can divide the rounds into $J$ epochs, with epoch \(j\) consists of the rounds
\[
N_{j-1}<t\leq N_j.
\]
Here we assume $N_{j_0-1}:=0$.  
For a completed epoch \(j<J\), the loss incurred during
\(N_{j-1}<t\leq N_j\) is at most the total loss of \(G_j\) on the realizable
prefix \(x_{1:N_j}\), and is therefore at most \(2^j\).  For the possibly
partial final epoch \(J\), UUS property lets us extend \(x_{1:T}\) to \(N_J\)
distinct demonstrations in \(\supp(h^\star)\), so its loss through time \(T\) is
again at most \(2^J\).  Summing the
epoch bounds gives
\[
  M_T(G,h^\star;x_{1:T})
  \leq
  \sum_{j=j_0}^{J}2^j
  <2^{J+1}.
\]
If \(J>j_0\), minimality of \(J\) gives \(T>N_{J-1}\), which by the
definition of \(N_{J-1}\) implies \(b(T)>2^{J-1}\).  If \(J=j_0\), then
the definition \(j_0=\lceil\log b(1)\rceil\) give
\(b(T)\geq b(1)>2^{j_0-1}\).  Consequently,
\[
  M_T(G,h^\star;x_{1:T})
  <
  4b(T).
\]
The pair \((x_{1:T},h^\star)\) was arbitrary, so this proves the asserted uniform
bound.

Finally, if \(\MF_T(\Hcal)=o(T)\), Lemma~\ref{lem:app-realizable-profile-regularity}
shows that the profile \(T\mapsto\MF_T(\Hcal)\) is nondecreasing.  Hence
\[
  b(T):=\max\{1,\sqrt{T},\MF_T(\Hcal)\}
\]
is nondecreasing.  The term \(\sqrt T\) makes \(b\) unbounded even when the
mistake-forcing profile is bounded, and all three terms are \(o(T)\), so
\(b(T)=o(T)\).  Applying the quantitative assertion produces one anytime
generator with worst-case realizable mistake count at most \(4b(T)=o(T)\).
Thus \(\Hcal\) is realizably no-regret generatable.
\end{proof}

\begin{theorem*}[Restatement of Theorem~\ref{thm:slow-growth}]
Suppose
\(
  d:=\sup_{T\geq1}\MF_T(\Hcal)<\infty.
\)
For every nondecreasing unbounded function
\(r:\mathbb N\to\mathbb N\) satisfying \(r(T)\geq1\), there exists one
anytime generator \(G\) such that
\[
  \sup_{\substack{h\in\Hcal(x_{1:T})}}
  M_T(G,h;x_{1:T})
  \leq d\,r(T)
  \qquad
  \text{for every }T\geq1.
\]
\end{theorem*}

\begin{proof}
Let \(d:=\sup_{T\geq1}\MF_T(\Hcal)<\infty\).  Choose epoch endpoints
\[
  0=N_0<N_1<N_2<\cdots
\]
as follows.  Having chosen \(N_{j-1}\), choose \(N_j>N_{j-1}\) sufficiently
large that
\[
  r(N_j+1)\geq j+1.
\]
Such a choice is possible because \(r\) is unbounded.  

For each \(j\), Theorem~\ref{thm:mf-finite-characterization} supplies a
horizon-\(N_j\) generator \(G_j\) with worst-case realizable mistake count
at most \(d\).  During the epoch \(N_{j-1}<t\leq N_j\), let the anytime
generator use the decision rule of \(G_j\) on the complete demonstration
history \(x_{1:t-1}\).

For any \(T\), demonstration sequence \(x_{1:T}\), and target
\(h^\star\in\Hcal(x_{1:T})\), let \(J\) be the least index such that
\(T\leq N_J\).  The rounds are divided into epochs, with epoch \(j\)
consisting of the rounds
\[
  N_{j-1}<t\leq N_j.
\]

For a completed epoch \(j<J\), the loss incurred during
\(N_{j-1}<t\leq N_j\) is at most the total loss of \(G_j\) on the realizable
prefix \(x_{1:N_j}\), and is therefore at most \(d\).  For the possibly
partial final epoch \(J\), the UUS property lets us extend \(x_{1:T}\) to
\(N_J\) distinct demonstrations in \(\supp(h^\star)\).  The loss is also at most \(d\).  Summing the epoch bounds gives
\[
  M_T(G,h^\star;x_{1:T})
  \leq
  dJ.
\]

By the minimality of \(J\), we have
\(T\geq N_{J-1}+1\).  Therefore, monotonicity of \(r\) and the choice of the
epoch endpoints give
\[
  r(T)
  \geq
  r(N_{J-1}+1)
  \geq
  J.
\]
Consequently,
\[
  \sup_{\substack{x_{1:T}\\h\in\Hcal(x_{1:T})}}
  M_T(G,h;x_{1:T})
  \leq
  d\,r(T).
\]
\end{proof}

\subsection{Proofs for Section~\ref{subsec:finite-versus-uniform}}

Partition the example space into finite blocks of unbounded size:
\[
  \Xcal=\bigsqcup_{n\geq1}B_n,
  \qquad |B_n|=n.
\]

We now construct our hypothesis  class such that each language have at most one invalid point in each block. Formally, for $k\in\naturals$, define hypotheses $h_E$ by 
\[
  \supp(h_E):=\Xcal\setminus E
\]
and set
\[
\begin{aligned}
\Hcal_k
&:=
\left\{
h_E:
|E|\leq k
\ \text{and}\
|E\cap B_n|\leq1\ \text{for every }n
\right\},\\
\Hcal_\infty
&:=
\left\{
h_E:
E\text{ is finite and }
|E\cap B_n|\leq1\ \text{for every }n
\right\}.
\end{aligned}
\]
The classes are countable and UUS because every support is cofinite.

\begin{theorem*}[Restatement of Theorem~\ref{thm:compactness-gap}]
For every \(k\in\mathbb N\cup\{\infty\}\), there exists a countable UUS
class \(\Hcal_k\) such that
\[
  \MF_T(\Hcal_k)=1
  \qquad
  \text{for every }T\geq1.
\]
For every finite \(k\),
\[
  \MFUnif(\Hcal_k)=k,
\]
whereas
\[
  \MFUnif(\Hcal_\infty)=\infty.
\]
\end{theorem*}

\paragraph{Proof idea.}
Intuitively, at a fixed horizon, the generator can choose distinct outputs
from one sufficiently large block.  Since each target has at most one invalid
point in that block, the generator makes at most one mistake.  An anytime
generator, however, cannot rely on a single finite block forever and must
eventually move to new blocks.  The adversary can then reserve an invalid
output in each newly visited block, allowing the uniform mistake-forcing
dimension to be much larger than one.

\begin{proof}
Fix \(T\geq1\) and choose a block \(B_n\) with \(n\geq2T\).  Define a
horizon-\(T\) generator that outputs points from \(B_n\),
always avoiding the demonstrations already revealed.  Before round \(t\),
at most \(2(t-1)\) points in the
block are excluded.  Since \(2(t-1)<2T\leq|B_n|\), an admissible point
remains.  Every \(E\) contains at most one point of
\(B_n\), so on every realizable play this generator makes at most one
mistake relative to \(h_E\).  Therefore
\[
  \MF_T(\Hcal_k)\leq1.
  \]
  This construction uses only the fact that each admissible set contains at
  most one point of \(B_n\), which also holds for every member of
  \(\Hcal_\infty\).  Hence the same upper bound holds for
  \(\Hcal_\infty\).

For the reverse inequality, fix an arbitrary horizon-\(T\) generator and
write \(y\) for its first output.  Reveal a point
\(x_1\in\Xcal\setminus\{y\}\), and then reveal distinct points outside
\(\{y\}\) for all remaining rounds.  These choices are possible because
\(\Xcal\setminus\{y\}\) is infinite.  The hypothesis \(h_{\{y\}}\), which
belongs to every class under consideration, realizes all the
demonstrations and makes the first output a mistake.  Thus every generator
has worst-case realizable loss at least one.  Applying
Theorem~\ref{thm:mf-finite-characterization} gives
\[
  \MF_T(\Hcal_k)=1
\]
for all \(T\geq1\) and all \(k\in\mathbb N\cup\{\infty\}\).

We now prove the property for $\MFUnif$.
For any finite \(k\) since every
\(h_E\in\Hcal_k\) has at most \(k\) invalid points, every
generator that does not repeat its outputs can make at most \(k\) mistakes. Thus
\[
  \MFUnif(\Hcal_k)\leq k.
\]

For the matching lower bounds, fix an arbitrary anytime generator and a
desired finite mistake count \(m\).  We construct a finite realizable prefix
on which it makes \(m\) mistakes.  Maintain a reserve \(R\), initially
empty, containing at most one point from each block, and ensure that every
demonstration lies outside the current reserve.  When the generator outputs
\(y\), respond as follows:
\begin{enumerate}
  \item If the block of \(y\) has no reserved point, add \(y\) to \(R\),
        count this output as a mistake, and reveal a fresh point outside both \(R\) and
        the previous demonstrations.
  \item If \(y\) is already the reserved point of its block, count this mistake and again reveal a fresh point outside \(R\) and the previous
        demonstrations.
  \item Otherwise, reveal \(y\) itself, making that nonreserved point
        unavailable as a future output.
\end{enumerate}
The fresh responses in the first two cases exist because \(R\) and the set
of previous demonstrations are finite.  A point added to \(R\) cannot have
been demonstrated earlier, since it is the generator's current legal
output.  It follows inductively that every demonstration remains outside
the eventual reserve.

Stop after \(m\) counted mistakes.  We claim that the stopping time is finite.  Indeed, before \(m\) outputs have been counted,
the generator can visit at most \(m-1\) blocks, because its first output in
each new block is reserved and counted.  These blocks contain only finitely
many points.  Every further output is either counted or immediately revealed
as a demonstration and hence becomes unavailable.  Thus the interaction
cannot continue indefinitely without reaching \(m\) counted outputs.

Now choose the target \(h^\star:=h_R\), whose invalid points are exactly the
reserved points in \(R\).  By construction, the adversary never reveals a
point in \(R\), so every demonstration is positive for \(h^\star\) and the
play is realizable.   For finite \(k\), take \(m=k\); then \(h_R\in\Hcal_k\).  For
\(\Hcal_\infty\), the finite reserve \(R\) is allowed for every finite
\(m\).  Combing the lower bound, we therefore have
\[
  \MFUnif(\Hcal_k)= k
  \quad\text{and}\quad
  \MFUnif(\Hcal_\infty)=\infty.
\]

\end{proof}

\subsection{Proofs for Section~\ref{subsec:realizable-spectrum}}

\begin{theorem*}[Restatement of Theorem~\ref{thm:arbitrary-profile}]
Let \(f:\mathbb N\to\mathbb N\) be nondecreasing and satisfy
\[
  1\leq f(T)\leq T
  \qquad
  \text{for every }T\geq1.
\]
There exists a countable UUS class \(\Hcal_f\) such that, for every
\(T\geq4\),
\[
  f(\lfloor T/4\rfloor)
  \leq
  \MF_T(\Hcal_f)
  \leq
  f(8T).
\]
Consequently, if \(f\) is stable under constant rescaling, in the sense
that, for every fixed \(c>0\),
\[
    f(cT)=\Theta(f(T)),
\]
where the implicit constants
may depend on \(c\), then
\[
    \MF_T(\Hcal_f)=\Theta(f(T)).
\]
\end{theorem*}

\paragraph{Proof idea.}
This construction reuses the block-and-reserve proof idea from
Theorem~\ref{thm:compactness-gap}. The earlier construction allowed one reserved invalid point per
block.  Here block \(B_n\) has size \(4^n\) and allows
\(k_n=f(4^n)\) reserved points.  A generator confined to one suitable block
then pays at most \(k_n\), while the adversary fills block reserves to force
the corresponding number of mistakes.

\begin{proof}
Partition a countably infinite example space into blocks
\[
  \Xcal=\bigsqcup_{n\geq1}B_n,
  \qquad
  |B_n|=s_n:=4^n,
\]
and put \(k_n:=f(s_n)\).  For every finite \(E\subseteq\Xcal\) satisfying
\(|E\cap B_n|\leq k_n\) for every \(n\), define \(h_E\) by
\(\supp(h_E)=\Xcal\setminus E\), and let
\[
  \Hcal_f
  :=
  \{h_E:E\subseteq\Xcal\text{ is finite and }
               |E\cap B_n|\leq k_n\text{ for every }n\}.
\]
There are only countably many finite subsets of the countable space
\(\Xcal\), so \(\Hcal_f\) is countable.  Moreover, each \(E\) is finite and
therefore \(\supp(h_E)=\Xcal\setminus E\) is infinite.  Thus \(\Hcal_f\)
satisfies UUS.

We first prove the upper bound.  Fix \(T\geq4\), and let \(n\) be the least
index with \(s_n\geq2T\).  Use the horizon-\(T\) generator that outputs
pairwise distinct points of \(B_n\) and avoids all earlier demonstrations.
Before round \(t\), at most \(t-1\) previous outputs and \(t-1\) previous
demonstrations in \(B_n\) are excluded.  Since this is fewer than \(2T\), a
legal choice always remains.  Fix a target \(h^\star:=h_E\).  A generated
point is a mistake only if it lies in \(E\cap B_n\).  The outputs are
distinct and \(|E\cap B_n|\leq k_n\), so the generator makes at most
\(k_n\) mistakes.
By Theorem~\ref{thm:mf-finite-characterization},
\[
  \MF_T(\Hcal_f)\leq k_n.
\]
Minimality of \(n\) gives \(s_{n-1}<2T\), and hence
\(s_n=4s_{n-1}<8T\).  Monotonicity of \(f\) now yields
\[
  \MF_T(\Hcal_f)
  \leq
  k_n
  =
  f(s_n)
  \leq
  f(8T).
\]

For the lower bound, retain this fixed \(T\), and set
\[
  u:=\lfloor T/4\rfloor,
  \qquad
  m:=f(u).
\]
Fix an arbitrary horizon-\(T\) generator.  We use the reserve idea from
the previous proof, with capacity \(k_n\) in block \(B_n\).  Maintain
a finite reserve \(R\) and a counter \(c\), initially \(R=\varnothing\) and
\(c=0\).  Given an output \(y\in B_n\), respond according to the following
rules until \(c=m\):
\begin{enumerate}
  \item If \(y\in R\), increase \(c\) by one and reveal a fresh point outside
        \(R\) and all previous demonstrations.
  \item If \(y\notin R\) and \(|R\cap B_n|<k_n\), add \(y\) to \(R\),
        increase \(c\) by one, and reveal a fresh point outside \(R\) and
        all previous demonstrations.
  \item If \(y\notin R\) and \(|R\cap B_n|=k_n\), reveal \(y\) itself.
\end{enumerate}

It remains to show that the counter reaches \(m\) before horizon \(T\).

Before the \(m\)-th counted output, we have \(c<m\) and \(|R|\leq c\).
Therefore, rule 3 can be used in block \(B_n\) only if
\[
  k_n<m,
\]
because that rule requires \(B_n\) to contain \(k_n\) reserved points.
Moreover, every use of rule 3 reveals the current output as a demonstration,
so that point cannot be output again.  Since \(B_n\) contains \(s_n\) points,
rule 3 can be used at most \(s_n\) times in that block.

Thus, before the \(m\)-th counted output, the total number of rounds is at
most
\[
  m+\sum_{n:k_n<m}s_n.
\]
Recall that \(m=f(u)\) and \(k_n=f(s_n)\).  By monotonicity of \(f\),
\[
  k_n<m
  \quad\Longrightarrow\quad
  f(s_n)<f(u)
  \quad\Longrightarrow\quad
  s_n<u.
\]
Consequently,
\[
  \sum_{n:k_n<m}s_n
  \leq
  \sum_{n:4^n<u}4^n
  <
  \frac{4}{3}u.
\]
Since \(m=f(u)\leq u\) and \(u=\lfloor T/4\rfloor\), the \(m\)-th counted
output occurs by round
\[
  m+\sum_{n:k_n<m}s_n
  <
  u+\frac{4}{3}u
  =
  \frac{7}{3}u
  \leq
  \frac{7}{12}T
  <
  T.
\]

At that time, \(R\) is finite and satisfies
\[
  |R\cap B_n|\leq k_n
  \qquad
  \text{for every }n.
\]
Hence \(h^\star:=h_R\) belongs to \(\Hcal_f\).  By construction, no point
of \(R\) has been demonstrated, so all demonstrations are positive for
\(h^\star\).  Every counted output, however, belongs to \(R\) and is
therefore a mistake for \(h^\star\).

Extend the demonstration sequence through round \(T\) using fresh points
from \(\supp(h^\star)\).  The \(m\) mistakes already made remain in the
cumulative loss.  Thus every horizon-\(T\) generator admits a realizable
play on which it makes at least \(m\) mistakes.  By
Theorem~\ref{thm:mf-finite-characterization},
\[
  \MF_T(\Hcal_f)
  \geq
  m
  =
  f(\lfloor T/4\rfloor).
\]

Finally, if \(f\) is stable under constant rescaling, then both
\(f(\lfloor T/4\rfloor)\) and \(f(8T)\) are within constant factors of
\(f(T)\), with constants depending only on the fixed rescaling factors.
The two bounds just proved therefore imply
\(\MF_T(\Hcal_f)=\Theta(f(T))\).
\end{proof}

\begin{corollary*}[Restatement of Corollary~\ref{cor:common-profiles}]
Each of the following asymptotic profiles is attained by
\(\MF_T(\Hcal)\) for some realizably no-regret generatable class \(\Hcal\):
\[
  (\log(1+T))^\beta,
  \qquad
  T^\alpha,
  \qquad
  T^\alpha(\log(1+T))^\beta,
  \qquad
  \frac{T}{(\log(1+T))^\beta},
\]
where \(\alpha\in(0,1)\) and \(\beta>0\).
\end{corollary*}

\begin{proof}
Apply Theorem~\ref{thm:arbitrary-profile} to the displayed functions.
Each resulting function is nondecreasing, lies between one and \(T\), sublinear in $T$ and
is stable under constant rescaling.  Therefore the theorem constructs a
countable UUS class whose mistake-forcing profile has the asserted
asymptotic order.
\end{proof}

\section{Proofs for Section~\ref{sec:agnostic-generation}}
\label{app:agnostic-proofs}
\subsection{Proofs for Section~\ref{subsec:agnostic-characterizations}}

\begin{theorem*}[Restatement of Theorem~\ref{thm:smf-characterization}]
For every nonempty UUS class \(\Hcal\) and every \(T\geq1\),
\[
  \SMF_T(\Hcal)
  =
  \min_{G_T}
  \sup_{\substack{x_{1:T}\\h\in\Hcal}}
  \Reg_T(G,h;x_{1:T}).
\]
For each \(T\), the minimum is attained by a generator \(G_T\), which may
depend on \(T\).
\end{theorem*}

\paragraph{Proof idea.}
The proof adapts the safe-move idea from the proof of
Theorem~\ref{thm:mf-finite-characterization}. We just replace the score recorded by the leaf: A
realizable leaf records only
the rounds on which the generator makes mistakes, here a leaf records a signed pair
\((I,J)\),  with score \(|I|-|J|\) being exactly the regret.  

\begin{proof}
We first prove the lower bound.  Suppose that an
\(m\)-signed-mistake-forcing tree of horizon \(T\) is given, and fix an
arbitrary horizon-\(T\) generator \(G_T\).  Starting at the root, query
\(G_T\) on the demonstrations already seen.  Generator-completeness says
that the current node has an outgoing edge whose generated-example label is
the output of \(G_T\).  Select one such edge and use its response label as
the next demonstration.  Repeating this procedure produces a root-to-leaf
trajectory
\[
  \tau=((\widehat x_1,x_1),\ldots,(\widehat x_T,x_T))
\]
that is consistent with \(G_T\).

The leaf is \(m\)-signed-bad, so there is a feasible pattern
\((I,J)\in S_{\Hcal}^{\pm}(\tau)\) such that \(|I|-|J|\geq m\).  Choose
\(h\in\Hcal^{\pm}(\tau;I,J)\).  By the definition of compatibility,
\(I\) is exactly the set of rounds on which \(h(\widehat x_t)=0\), and
\(J\) is exactly the set of rounds on which \(h(x_t)=0\).  Consequently,
\[
\begin{aligned}
  \Reg_T(G_T,h;x_{1:T})
  &=\sum_{t=1}^T
    \bigl(\ell_h(\widehat x_t)-\ell_h(x_t)\bigr)\\
  &=|I|-|J|
  \geq m.
\end{aligned}
\]
Since \(G_T\) was arbitrary, taking the largest forceable \(m\) gives
\[
  \min_{G_T}
  \sup_{\substack{x_{1:T}\\h\in\Hcal}}
  \Reg_T(G_T,h;x_{1:T})
  \geq \SMF_T(\Hcal).
\]

For the reverse inequality, put \(d:=\SMF_T(\Hcal)\).  For a legal
trajectory \(\tau\) of length at most \(T\), call \(\tau\) forcing if it
roots a generator-complete continuation tree all of whose leaves have total
length \(T\) and are \((d+1)\)-signed-bad.  The empty trajectory is not
forcing, since otherwise that continuation tree would be a
\((d+1)\)-signed-mistake-forcing tree of horizon \(T\), contradicting the
definition of \(d\).

We claim that every non-forcing trajectory \(\tau\) of length less than
\(T\) admits a safe legal output: an output \(\widehat x\) for which every
legal response \(x\) makes
\(\tau\circ(\widehat x,x)\) non-forcing.  If no such output existed, then
for every legal \(\widehat x\) we could choose a legal response
\(x\) whose successor is forcing.  Attach below the edge
\((\widehat x,x)\) a forcing continuation tree for that
successor.  Doing this for every legal output creates a generator-complete
forcing tree rooted at \(\tau\), a contradiction.  This proves the claim.

Starting from \(\tau_\varnothing\), let \(G_T\) recursively choose a safe
output.  Induction on the number of rounds shows that every trajectory
reached by this generator is non-forcing.  At length \(T\), a trajectory
consisting of a single leaf would itself be a forcing continuation tree if
that leaf were \((d+1)\)-signed-bad.  Hence every terminal trajectory
\(\tau\) reached by \(G_T\) satisfies
\(\rho_{\Hcal}(\tau)\leq d\).

For any \(h\in\Hcal\), its behavior on \(\tau\) determines a feasible
signed pattern \((I_h,J_h)\).  As in the lower bound, its regret equals the
signed score of that pattern.  Therefore
\[
  \Reg_T(G_T,h;x_{1:T})\leq d
  \qquad
  \text{for every }h\in\Hcal
\]
on every legal demonstration sequence.  Thus the minimax value is at most
\(d\).  Together with the lower bound, this proves equality, and the safe
generator constructed above attains the minimum.
\end{proof}

Similar as in the realizable setting, we prove the regularity of $\SMF_T$.
\begin{lemma}[Regularity of the agnostic profile]
\label{lem:app-agnostic-profile-regularity}
For every nonempty UUS class and every \(T\geq1\),
\[
  0
  \leq
  \SMF_T(\Hcal)
  \leq
  \SMF_{T+1}(\Hcal)
  \leq
  \SMF_T(\Hcal)+1.
\]
\end{lemma}

\begin{proof}
First note that, in each round, since generator always outputs an $\widehat x_t$ first, the adversary can always choose to output $x_t=\widehat x_t$. Using this adversarial strategy, every language class then has
increment
\[
  \ell_h(\widehat x_t)-\ell_h(x_t)=0
\]
on every round.  Hence the worst-case regret of every generator is
nonnegative, and Theorem~\ref{thm:smf-characterization} gives
\(\SMF_T(\Hcal)\geq0\).

To prove monotonicity, fix a horizon-\((T+1)\) generator and restrict it to
its first \(T\) rounds.  Any legal length-\(T\) play can be extended by
responding on round \(T+1\) with the generator's output.  This final round
has zero increment for every comparator, so the worst-case regret of the
restriction is no larger than that of the original generator.  Minimizing
the two quantities and applying
Theorem~\ref{thm:smf-characterization} yields
\[
  \SMF_T(\Hcal)\leq\SMF_{T+1}(\Hcal).
\]

For the last inequality, use a minimizing horizon-\(T\) generator for the
first \(T\) rounds, whose existence follows from
Theorem~\ref{thm:smf-characterization}, and choose any legal output on the
last round.  The last signed increment belongs to \(\{-1,0,1\}\), and in
particular is at most one.  The resulting horizon-\((T+1)\) generator has
worst-case regret at most \(\SMF_T(\Hcal)+1\).  Another application of the
characterization proves
\(\SMF_{T+1}(\Hcal)\leq\SMF_T(\Hcal)+1\).
\end{proof}

\begin{theorem*}[Restatement of Theorem~\ref{thm:agnostic-horizon-anytime-main}]
For every nonempty UUS class \(\Hcal\), the following are equivalent:
\begin{enumerate}
  \item \(\Hcal\) is agnostically no-regret generatable;
  \item \(\SMF_T(\Hcal)=o(T)\) as \(T\to\infty\).
\end{enumerate}
Moreover, let \(b:\mathbb N\to[1,\infty)\) be nondecreasing and unbounded.
If
\[
  \SMF_T(\Hcal)\leq b(T)
  \qquad
  \text{for every }T\geq1,
\]
then there exists one deterministic anytime generator \(G\) such that,
simultaneously for every \(T\geq1\),
\[
  \sup_{\substack{x_{1:T}\\h\in\Hcal}}
  \Reg_T(G,h;x_{1:T})
  \leq4b(T).
\]
\end{theorem*}

\paragraph{Proof idea.}
We adapt the value-doubling construction from the proof of
Theorem~\ref{thm:horizon-anytime-main}.  The realizable argument does not
extend directly because cumulative mistakes never decrease, whereas agnostic
regret can.  We overcome this difficulty using a virtual-generator
construction, described in the proof below.

\begin{proof}
Suppose first that one anytime generator \(G\) has worst-case agnostic regret
\(o(T)\).  Restricting \(G\) to its first \(T\) decision rules gives a
horizon-\(T\) generator.  By
Theorem~\ref{thm:smf-characterization},
\(\SMF_T(\Hcal)\) is at most the horizon-\(T\) worst-case regret of this
generator, and hence
\[
  \SMF_T(\Hcal)=o(T).
\]

We now prove the quantitative assertion.  Let
\(b:\mathbb N\to[1,\infty)\) be nondecreasing and unbounded, with
\(\SMF_T(\Hcal)\leq b(T)\) for every \(T\).  Put
\[
  j_0:=\left\lceil\log b(1)\right\rceil,
  \qquad
  N_{j_0-1}:=0,
\]
and, for every \(j\geq j_0\), define
\[
  N_j:=\max\{T\in\mathbb N:b(T)\leq2^j\}.
\]
These maxima are finite, the sequence \((N_j)_{j\geq j_0}\) is
nondecreasing, and \(N_j\to\infty\). 

For each \(j\), Theorem~\ref{thm:smf-characterization} supplies a minimizing
horizon-\(N_j\) generator \(G_j\), whose worst-case regret is at most
\[
  \SMF_{N_j}(\Hcal)\leq2^j.
\]

We now combine the horizon-dependent generators \(G_j\) into one anytime
generator \(G\).  We divide the rounds into epochs.  For each
\(j\geq j_0\), epoch \(j\) consists of the rounds
\[
  N_{j-1}<t\leq N_j.
\]

We cannot directly use the argument from the realizable proof.
There, all mistake increments are nonnegative, so the horizon guarantee for
\(G_j\) controls both its contribution during an epoch and every prefix of
that epoch.  Agnostic regret has signed increments: negative regret before
an epoch may hide a large positive contribution during it.

We therefore start a fresh copy of \(G_j\) in each epoch, so that its regret
records only the contribution of that epoch.  Two details remain.  First,
this fresh copy has not seen the demonstrations from earlier epochs and may
propose one of them, which would be illegal in the real interaction.
Second, its guarantee of regrets applies at the end of horizon \(N_j\), whereas the
current epoch may stop earlier.  We handle both issues using zero-regret
virtual rounds.

We address the first point by using virtual rounds.  At the beginning of
epoch \(j\), start a fresh copy of \(G_j\) and let
\[
  S_j:=\{x_1,\ldots,x_{N_{j-1}}\}
\]
be the set of earlier real demonstrations.  Query \(G_j\) until it proposes
a point outside \(S_j\).  If it proposes \(z\in S_j\), return \(z\) as its
virtual response without advancing the real interaction.  This round is
legal in the virtual interaction because \(z\) has not previously appeared
in its virtual history.  It also contributes zero regret for every
\(h\in\Hcal\), since
\[
  \ell_h(z)-\ell_h(z)=0.
\]
After receiving \(z\), the virtual generator cannot propose \(z\) again.

Once \(G_j\) proposes \(z\notin S_j\), use \(z\) as the next real output.
After the adversary reveals the real response, feed that response to the
virtual generator.  The output \(z\) is legal in the real interaction: it
does not belong to \(S_j\), and the legality of \(G_j\) ensures that it is
different from all demonstrations made during the current epoch.

This simulation fits within the horizon of \(G_j\).  There are
\(N_j-N_{j-1}\) real rounds in epoch \(j\).  In addition, each point of
\(S_j\) can produce at most one skipped virtual round, so there are at most
\(N_{j-1}\) such rounds.  Therefore the total number of virtual rounds is at
most
\[
  (N_j-N_{j-1})+N_{j-1}
  =
  N_j.
\]
We next extend its virtual to $N_j$ rounds, in which the regret of $G_j$ is guaranteed to be bounded above by $2^j$.
Let \(s\leq N_j\) be the number of virtual
rounds used in its simulation. Extend this virtual play to \(N_j\)
rounds by setting \(x_t=\widehat{x}_t\) on each remaining round.
Each response is legal because \(G_j\)'s output avoids all previous
virtual demonstrations, and contributes zero regret for every
\(h\in\Hcal\). The padded play therefore has the same regret as  the real
prefix because all inserted simulation rounds also contribute zero
regret. Applying the horizon-\(N_j\) guarantee for \(G_j\) to the
padded play bounds the regret on the chosen prefix by \(2^j\)
for every \(h\in\Hcal\).

Now fix a horizon \(T\), and let \(J\) be the least index such that
\(T\leq N_J\).  The regret through time \(T\) is the sum of the regret from
the completed epochs \(j_0,\ldots,J-1\) and the regret from the current
prefix of epoch \(J\).  Thus
\[
  \sup_{\substack{x_{1:T}\\h\in\Hcal}}
  \Reg_T(G,h;x_{1:T})
  \leq
  \sum_{j=j_0}^{J}2^j
  <
  2^{J+1}.
\]

If \(J>j_0\), then the minimality of \(J\) gives \(T>N_{J-1}\).  
Definition of \(N_{J-1}\) implies $ b(T)>2^{J-1}.$
If \(J=j_0\), then $b(T)\geq b(1)>2^{j_0-1}.$
Therefore, in both cases,
\[
  \sup_{\substack{x_{1:T}\\h\in\Hcal}}
  \Reg_T(G,h;x_{1:T})
  <
  4b(T).
\]
This proves the quantitative assertion.

Finally, suppose that \(\SMF_T(\Hcal)=o(T)\).  Define
\[
  b(T):=\max\{1,\sqrt{T},\SMF_T(\Hcal)\}.
\]
Lemma~\ref{lem:app-agnostic-profile-regularity} implies that \(b\) is
nondecreasing.  The term \(\sqrt{T}\) makes it unbounded, and
\(b(T)=o(T)\).  Applying the quantitative assertion gives one anytime
generator with worst-case regret at most
\[
  4b(T)=o(T).
\]
Hence \(\Hcal\) is agnostically no-regret generatable.
\end{proof}

\subsection{Proofs for Section~\ref{subsec:realizable-versus-agnostic}}

\begin{proposition*}[Restatement of Proposition~\ref{prop:mf-le-smf}]
For every nonempty UUS class \(\Hcal\) and every \(T\geq1\),
\[
  \MF_T(\Hcal)\leq\SMF_T(\Hcal).
\]
\end{proposition*}

\begin{proof}
Let a leaf of an ordinary \(m\)-mistake-forcing tree be witnessed by
\(I\in S_{\Hcal}(\tau)\) and \(h\in\Hcal(\tau;I)\), with \(|I|\geq m\).
Because \(h(x_t)=1\) on every round, the pair \((I,\varnothing)\) is a
feasible signed pattern and has signed score \(|I|\geq m\).  Thus every
ordinary \(m\)-mistake-forcing tree is also an
\(m\)-signed-mistake-forcing tree.
\end{proof}

\begin{theorem*}[Restatement of Theorem~\ref{thm:pair-separation}]
There exists a countable UUS class \(\Hcal_{\mathrm{pair}}\) satisfies
\[
  \MFUnif(\Hcal_{\mathrm{pair}})\leq2,
\]
whereas
\[
  \SMF_T(\Hcal_{\mathrm{pair}})
  \geq
  \frac{T}{3}
  \qquad
  \text{for every }T\geq1.
\]
\end{theorem*}

\begin{proof}
Let
\[
    \Xcal
    :=
    \{a_i:i\in\mathbb{N}\}
    \mathbin{\dot\cup}
    \{c_{\{i,j\},k}:i<j,\ k\in\mathbb{N}\}.
\]
We call \(a_i\) the \emph{base point} associated with index \(i\), and
\(c_{\{i,j\},k}\) the \(k\)-th \emph{pair point} associated with the pair
\(\{i,j\}\).

For every cofinite \(P\subseteq\mathbb{N}\), define
\[
    \supp(h_P)
    :=
    \{a_i:i\in P\}
    \cup
    \{c_{\{i,j\},k}:i,j\in P,\ i<j,\ k\in\mathbb{N}\},
\]
where \(h_P\in\{0,1\}^{\Xcal}\), and let
\[
    \Hcal_{\mathrm{pair}}
    :=
    \left\{
        h_P:
        P\subseteq\mathbb{N}\text{ is cofinite}
    \right\}.
\]
We first construct a realizable anytime generator.  Fix a realizable play
with target \(h^\star:=h_P\).  Every index appearing in a demonstration
belongs to \(P\).  A  pair point $c_{\{i,j\},k}$ reveals two distinct indices $i,j$
in \(P\).  If no pair point appears among the first two demonstrations, then
those demonstrations are distinct base points \(a_i\) and \(a_j\), and hence
reveal two distinct indices \(i,j\in P\).  Thus, after at most two rounds,
the generator knows two distinct indices \(i,j\in P\).

From then on, the generator outputs points \(c_{\{i,j\},k}\) with fresh
indices \(k\).  Such a point is always available and belongs to
\(\supp(h^\star)\).  Therefore, only the first two outputs can be mistakes.
By Theorem~\ref{thm:mf-infinite-characterization},
\[
  \MFUnif(\Hcal_{\mathrm{pair}})\leq2.
\]

For the signed lower bound, fix an arbitrary deterministic horizon-\(T\)
generator \(G\).  Our goal is to construct a legal demonstration sequence
and then choose a comparator \(h_P\) such that every generated point is
invalid for \(h_P\), while at most \(2T/3\) demonstrations are invalid.
This will give regret at least \(T/3\).

After observing the output on round \(t\), choose an index \(i_t\) that has
not appeared among \(i_1,\ldots,i_{t-1}\) and has not been mentioned by any
generator output up to that round, and reveal
\[
  x_t:=a_{i_t}.
\]
Only finitely many indices are
excluded at each round, so such an \(i_t\) always exists.  The resulting
demonstrations are pairwise distinct and form a legal stream.

Let
\[
  D:=\{i_1,\ldots,i_T\}.
\]
By our construction of $i_t$, no generated base point has its index in \(D\). 
Now form a simple graph \(\Gamma\) with vertex set \(D\).  Add the edge
\(\{i,j\}\) whenever the generator outputs a pair point associated with
\(\{i,j\}\subseteq D\).  Since at most one edge is added per round,
\(\Gamma\) has \(T\) vertices and at most \(T\) edges. By Lemma~\ref{lem:sparse-graph-cover}, \(\Gamma\) has a vertex cover
\(C\subseteq D\), meaning that every edge has at least one endpoint in
\(C\), such that
\[
  |C|\leq\frac{2T}{3}.
\]

We now choose the comparator.  Construct a finite set \(F\) by including:
\begin{enumerate}
  \item every index in \(C\);
  \item the index of every generated base point;
  \item for every generated pair point \(c_{\{i,j\},k}\) with
        \(\{i,j\}\not\subseteq D\), one index from
        \(\{i,j\}\setminus D\).
\end{enumerate}
Every index added in the second item lies outside \(D\), because no
generated base point has its index in \(D\).  Every index added in the third
item also lies outside \(D\) by construction.  Therefore, the only elements
of \(D\) included in \(F\) are those in \(C\), and hence
\[
  F\cap D=C.
\]

Set \(P:=\mathbb N\setminus F\).  Then \(P\) is cofinite, so
\(h_P\in\Hcal_{\mathrm{pair}}\).

Every generated point is invalid for \(h_P\).  A generated base point has
its index in \(F\).  A generated pair point contained in \(D\) meets the vertex
cover \(C\subseteq F\), while a generated pair not contained in \(D\) has
one index explicitly included in \(F\).  On the other hand, a demonstration
\(a_{i_t}\) is invalid exactly when \(i_t\in F\cap D=C\).  Therefore,
\[
  \Reg_T(G,h_P;x_{1:T})
  =
  T-|C|
  \geq
  \frac{T}{3}.
\]
Since \(G\) was arbitrary, Theorem~\ref{thm:smf-characterization} gives
\[
  \SMF_T(\Hcal_{\mathrm{pair}})
  \geq
  \frac{T}{3}.
\]
\end{proof}

The next lemma is the classical Caro--Wei bound
\cite{caro1979new,wei1981lower}.

\begin{lemma}[Caro--Wei bound]
\label{lem:sparse-graph-cover}
Let \(\Gamma=(V,E)\) be a finite simple graph.  A set \(I\subseteq V\) is
\emph{independent} if no edge has both endpoints in \(I\), and a set
\(C\subseteq V\) is a \emph{vertex cover} if every edge has at least one
endpoint in \(C\).

The graph \(\Gamma\) contains an independent set \(I\) satisfying
\[
  |I|
  \geq
  \frac{|V|^2}{2|E|+|V|}.
\]
Consequently, if \(|E|\leq|V|\), then \(\Gamma\) has a vertex cover \(C\)
satisfying
\[
  |C|\leq\frac{2|V|}{3}.
\]
\end{lemma}

\begin{proof}
Choose a uniformly random ordering of \(V\), and include a vertex \(v\) in
\(I\) if it precedes all of its neighbors.  No two adjacent vertices can
both be selected, so \(I\) is independent.  Moreover, \(v\) is selected
with probability \(1/(d(v)+1)\).  Therefore,
\[
  \mathbb E|I|
  =
  \sum_{v\in V}\frac{1}{d(v)+1}
  \geq
  \frac{|V|^2}{\sum_{v\in V}(d(v)+1)}
  =
  \frac{|V|^2}{2|E|+|V|},
\]
where the inequality follows from Cauchy--Schwarz.  Hence some ordering
produces an independent set of the stated size.

If \(|E|\leq|V|\), then \(|I|\geq|V|/3\).  Its complement
\(C:=V\setminus I\) is a vertex cover, because no edge can have both
endpoints in \(I\).  Thus
\[
  |C|
  =
  |V|-|I|
  \leq
  \frac{2|V|}{3}.
\]
\end{proof}

\subsection{Proofs for Section~\ref{subsec:agnostic-spectrum}}

\begin{theorem*}[Restatement of Theorem~\ref{thm:arbitrary-agnostic-profile}]
Let \(f:\mathbb N\to\mathbb N\) be nondecreasing and satisfy
\[
  1\leq f(T)\leq T
  \qquad
  \text{for every }T\geq1.
\]
There exists a countable UUS class \(\Hcal_f\) such that, for every
\(T\geq4\),
\[
  f(\lfloor T/4\rfloor)
  \leq
  \MF_T(\Hcal_f)
  \leq
  \SMF_T(\Hcal_f)
  \leq
  f(8T).
\]
Consequently, if \(f\) is stable under constant rescaling, then
\[
  \SMF_T(\Hcal_f)=\Theta(f(T)).
\]
\end{theorem*}

\paragraph{Proof idea.}
We use the class \(\Hcal_f\) already constructed in the proof of
Theorem~\ref{thm:arbitrary-profile}.  For the lower bound, we invoke the result
\(\MF_T\leq\SMF_T\).  For the agnostic upper bound, we  adapt the
finite-block generator from that theorem's proof and use a new observation: if the generator stays inside
one sufficiently large block, the regret is
at most the generator's number of mistakes. 

\begin{proof}
Let \(\Hcal_f\) be the block class constructed in the proof of
Theorem~\ref{thm:arbitrary-profile}: the example space is partitioned into
blocks satisfying
\[
  |B_n|=s_n:=4^n,
  \qquad
  k_n:=f(s_n),
\]
and \(h_E\in\Hcal_f\) exactly when \(E\) is finite,
\(\supp(h_E)=\Xcal\setminus E\), and
\(|E\cap B_n|\leq k_n\) for every \(n\).  That construction also proves
that \(\Hcal_f\) is countable and UUS.

We first invoke two earlier results.
The lower bound from Theorem~\ref{thm:arbitrary-profile}, followed by
Proposition~\ref{prop:mf-le-smf}, gives
\[
  f(\lfloor T/4\rfloor)
  \leq
  \MF_T(\Hcal_f)
  \leq
  \SMF_T(\Hcal_f).
\]
It remains to prove the agnostic upper bound.  We adapt the
finite-block generator from the proof of
Theorem~\ref{thm:arbitrary-profile}; the agnostic regret comparison below is
the only new ingredient.  Fix \(T\geq4\), and let \(n\) be the least index
with \(s_n\geq2T\).  On every round, output
a point of \(B_n\) that has appeared neither as an earlier output nor as a
demonstration.  Before round \(t\), at most \(t-1\) earlier outputs and at
most \(t-1\) demonstrations are unavailable.  Their total is less than
\(2T\leq s_n\), so such an output always exists and is legal.  Minimality
of \(n\) gives
\[
  s_n=4s_{n-1}<8T.
\]

Now fix an arbitrary legal demonstration stream and comparator
\(h_E\in\Hcal_f\).  The comparator's loss on the demonstrations is
nonnegative, so discarding it can only increase regret.  All generated
points are distinct and lie in \(B_n\); hence precisely those belonging to
\(E\cap B_n\) can be mistakes.  Therefore
\[
\begin{aligned}
  \Reg_T(G,h_E;x_{1:T})
  &=
  \sum_{t=1}^{T}
  \bigl(\ell_{h_E}(\widehat x_t)-\ell_{h_E}(x_t)\bigr)\\
  &\leq
  \sum_{t=1}^{T}\ell_{h_E}(\widehat x_t)\\
  &\leq
  |E\cap B_n|\\
  &\leq
  k_n.
\end{aligned}
\]
The displayed bound holds simultaneously for every stream and comparator.
Theorem~\ref{thm:smf-characterization} and monotonicity of \(f\) therefore
give
\[
  \SMF_T(\Hcal_f)\leq k_n=f(s_n)\leq f(8T).
\]

If \(f\) is stable under constant rescaling, then both
\(f(\lfloor T/4\rfloor)\) and \(f(8T)\) have order \(f(T)\).  The lower
and upper bounds just proved consequently imply
\(\SMF_T(\Hcal_f)=\Theta(f(T))\).
\end{proof}

\begin{corollary*}[Restatement of Corollary~\ref{cor:arbitrary-anytime-agnostic-rates}]
Each of the following asymptotic profiles is attained by
\(\SMF_T(\Hcal)\) for some agnostically no-regret generatable class \(\Hcal\):
\[
    (\log(1+T))^\beta,
    \qquad
    T^\alpha,
    \qquad
    T^\alpha(\log(1+T))^\beta,
    \qquad
    \frac{T}{(\log(1+T))^\beta},
\]
where \(\alpha\in(0,1)\) and \(\beta>0\). 
\end{corollary*}
The proof is essentially the same as Corollary~\ref{cor:common-profiles}.

\section{Proofs for Section~\ref{sec:comparison-generation-criteria}}
\label{app:comparison-proofs}
\subsection{Definitions of the generation criteria}
\label{app:comparison-criteria-definitions}

Throughout this subsection, \(\Hcal\subseteq\{0,1\}^{\Xcal}\) is nonempty
and UUS. All generators are deterministic and anytime, all demonstration
sequences consist of pairwise distinct points, and
\(\widehat x_t=G(x_{1:t-1})\). An infinite play is \emph{realizable by}
\(h\in\Hcal\) if \(x_t\in\supp(h)\) for every \(t\geq1\).
Such a sequence need not exhaust \(\supp(h)\). We call it an
\emph{enumeration} of \(\supp(h)\) if it additionally satisfies
$$
    \{x_t:t\geq1\}=\supp(h).
$$

\begin{definition}[Uniform generation
\cite{kleinberg2024language,li2024generationlenslearningtheory}]
\label{def:uniform-generation}
A class \(\Hcal\) is \emph{uniformly generatable} if there exist a generator
\(G\) and \(B\in\mathbb N_0\) such that, for every \(h\in\Hcal\), every
infinite play realizable by \(h\), and every \(t>B\),
$$
    \ell_h(\widehat x_t)=0.
$$
\end{definition}

\begin{definition}[Non-uniform generation
\cite{li2024generationlenslearningtheory}]
\label{def:nonuniform-generation}
A class \(\Hcal\) is \emph{non-uniformly generatable} if there exists a
generator \(G\) such that, for every \(h\in\Hcal\), there is a
\(B_h\in\mathbb N_0\) for which every infinite play realizable by \(h\)
satisfies
$$
    \ell_h(\widehat x_t)=0\qquad\text{for every }t>B_h.
$$
\end{definition}

\begin{definition}[Generation in the limit
\cite{kleinberg2024language}]
\label{def:generation-in-limit}
A class \(\Hcal\) is \emph{generatable in the limit} if there exists a
generator \(G\) such that, for every \(h\in\Hcal\) and every enumeration
\(x_{1:\infty}\) of \(\supp(h)\), there is a
\(B_{h,x_{1:\infty}}\in\mathbb N_0\) satisfying
$$
    \ell_h(\widehat x_t)=0
    \qquad\text{for every }t>B_{h,x_{1:\infty}}.
$$
\end{definition}

\begin{definition}[Zero-rate generation in the limit
\cite{panigrahi2026hallucinationrates}]
\label{def:zero-rate-generation}
A class \(\Hcal\) is \emph{zero-rate generatable in the limit} if there
exists a generator \(G\) such that, for every \(h\in\Hcal\) and every
enumeration \(x_{1:\infty}\) of \(\supp(h)\),
$$
    \limsup_{T\to\infty}
    \frac{M_T(G,h;x_{1:T})}{T}=0.
$$
\end{definition}

We also recall the two criteria defined earlier. A class is
\emph{uniformly mistake-bounded generatable} if there exist a generator
\(G\) and \(B\in\mathbb N_0\) such that
$$
    M_\infty(G,h;x_{1:\infty})\leq B
$$
for every \(h\in\Hcal\) and every infinite play realizable by \(h\).
It is \emph{realizably no-regret generatable} if there exists a generator
\(G\) satisfying
$$
    \sup_{\substack{x_{1:T}\\h\in\Hcal(x_{1:T})}}
    M_T(G,h;x_{1:T})=o(T).
$$
In the latter display, the supremum ranges over all pairs consisting of a
length-\(T\) demonstration sequence and a language realizing that sequence.

\subsection{Proofs for Section~\ref{subsec:comparison-realizable-criteria}}

For the theorem and its proof, abbreviate the six properties as follows:
$$
\begin{array}{ll}
\mathsf U: & \text{uniform generation},\\
\mathsf M: & \text{uniform mistake-bounded generation},\\
\mathsf N: & \text{non-uniform generation},\\
\mathsf R: & \text{realizable no-regret generation},\\
\mathsf L: & \text{generation in the limit},\\
\mathsf Z: & \text{zero-rate generation in the limit}.
\end{array}
$$
An implication between two properties means that every nonempty UUS class
satisfying the first also satisfies the second.

\begin{theorem*}[Formal statement of
Theorem~\ref{thm:comparison-realizable-criteria}]
The following implications hold:
$$
    \mathsf U\Longrightarrow\mathsf M
    \Longrightarrow\mathsf R\Longrightarrow\mathsf Z,
    \qquad
    \mathsf U\Longrightarrow\mathsf N
    \Longrightarrow\mathsf L\Longrightarrow\mathsf Z,
    \qquad
    \mathsf M\Longrightarrow\mathsf L.
$$
These implications are complete: for any distinct properties
\(P,Q\in\{\mathsf U,\mathsf M,\mathsf N,\mathsf R,\mathsf L,\mathsf Z\}\),
if there is no directed path from \(P\) to \(Q\) using the displayed
arrows, there exists a nonempty UUS class satisfying \(P\) but not \(Q\).
In particular, each displayed implication is strict, and any two properties
with no directed path in either direction are incomparable.
\end{theorem*}

\begin{proof}
We prove the implications, their strictness, and the incomparability
claims in turn.

\medskip
\noindent\emph{Implications.}
If \(G\) witnesses \(\mathsf U\) with cutoff \(B\), then it makes at most
\(B\) mistakes on every realizable play, proving \(\mathsf U\Rightarrow
\mathsf M\). The same cutoff works for each target separately, proving
\(\mathsf U\Rightarrow\mathsf N\).

Suppose \(G\) witnesses \(\mathsf M\) with bound \(B\). Fix \(T\), a
demonstration sequence \(x_{1:T}\), and \(h\in\Hcal(x_{1:T})\).
Since \(\supp(h)\) is infinite, this prefix extends to an infinite play
realizable by the same \(h\). Therefore
$$
    M_T(G,h;x_{1:T})\leq B.
$$
The bound is uniform in the prefix and target, so it proves
\(\mathsf M\Rightarrow\mathsf R\). On each enumeration of a target,
the same generator makes finitely many mistakes, hence has a finite
last-mistake cutoff. This proves \(\mathsf M\Rightarrow\mathsf L\).

Every enumeration of a target is an infinite realizable play for that target.
Thus a target-dependent cutoff also applies to each enumeration, proving
\(\mathsf N\Rightarrow\mathsf L\). A uniform \(o(T)\) mistake bound
applies to every fixed target enumeration, proving
\(\mathsf R\Rightarrow\mathsf Z\). Finally, finitely many mistakes have
asymptotic frequency zero, proving \(\mathsf L\Rightarrow\mathsf Z\).

\medskip
\noindent\emph{Strictness of the implications.}
The classes \(\Hcal_1\) and \(\Hcal_\infty\) from
Theorem~\ref{thm:compactness-gap} satisfy
$$
    \MFUnif(\Hcal_1)=1,
    \qquad
    \MF_T(\Hcal_\infty)=1\quad(T\geq1),
    \qquad
    \MFUnif(\Hcal_\infty)=\infty.
$$
By Theorems~\ref{thm:mf-infinite-characterization}
and~\ref{thm:horizon-anytime-main}, \(\Hcal_1\) satisfies \(\mathsf M\),
whereas \(\Hcal_\infty\) satisfies \(\mathsf R\) but not \(\mathsf M\).
Both classes consist of cofinite languages and are subclasses of the
cofinite class in Proposition~\ref{prop:cofinite-linear-profile}.
The non-uniform generator from that proposition therefore witnesses
\(\mathsf N\) for each of these subclasses.

The class \(\Hcal_1\) does not satisfy \(\mathsf U\). To prove this,
fix any generator \(G\) and any proposed cutoff \(B\in\mathbb N_0\).
Choose \(B\) distinct demonstrations and put \(y=G(x_{1:B})\), using
the empty history if \(B=0\). Legality gives
\(y\notin\{x_1,\ldots,x_B\}\). The language with support
\(\Xcal\setminus\{y\}\) belongs to \(\Hcal_1\), realizes this prefix,
and makes the output on round \(B+1\) invalid. Extend the prefix to an
infinite realizable play for this language. Thus no pair \((G,B)\)
witnesses \(\mathsf U\).

By Proposition~\ref{prop:cofinite-linear-profile},
\(\Hcal_{\mathrm{cof}}\) satisfies \(\mathsf N\) and
$$
    \MF_T(\Hcal_{\mathrm{cof}})=T\qquad(T\geq1).
$$
Theorem~\ref{thm:horizon-anytime-main} rules out \(\mathsf R\).
The implications proved above give the remaining positive properties
used below.

The class \(\Hcal_1\) satisfies \(\mathsf M\) and \(\mathsf N\)
but not \(\mathsf U\), so both implications from \(\mathsf U\)
are strict. The class \(\Hcal_\infty\) satisfies \(\mathsf R\)
and \(\mathsf L\) but not \(\mathsf M\), so
\(\mathsf M\Rightarrow\mathsf R\) and
\(\mathsf M\Rightarrow\mathsf L\) are strict.

Lemma~\ref{lem:row-mistake-nonuniform} below constructs a nonempty UUS
class \(\Hcal_{\mathrm{row}}\) satisfying \(\mathsf M\) but not
\(\mathsf N\). Since \(\mathsf M\Rightarrow\mathsf L\), this
proves that \(\mathsf N\Rightarrow\mathsf L\) is strict.
The class \(\Hcal_{\mathrm{cof}}\) satisfies \(\mathsf N\), hence
\(\mathsf Z\), but not \(\mathsf R\), proving that
\(\mathsf R\Rightarrow\mathsf Z\) is strict.
Finally, Lemma~\ref{lem:transversal-regret-limit} below constructs a
nonempty UUS class \(\Hcal_{\mathrm{tr}}\) satisfying \(\mathsf R\)
but not \(\mathsf L\). It satisfies \(\mathsf Z\) by
\(\mathsf R\Rightarrow\mathsf Z\), so
\(\mathsf L\Rightarrow\mathsf Z\) is strict as well.

\medskip
\noindent\emph{Incomparable criteria.}
Exactly three unordered pairs have no directed path in either direction.
For \(\mathsf M\) and \(\mathsf N\), the class
\(\Hcal_{\mathrm{row}}\) satisfies \(\mathsf M\) but not
\(\mathsf N\), whereas \(\Hcal_\infty\) satisfies
\(\mathsf N\) but not \(\mathsf M\).
For \(\mathsf N\) and \(\mathsf R\), the class
\(\Hcal_{\mathrm{cof}}\) satisfies \(\mathsf N\) but not
\(\mathsf R\), whereas \(\Hcal_{\mathrm{row}}\) satisfies
\(\mathsf R\) but not \(\mathsf N\).
For \(\mathsf R\) and \(\mathsf L\), the class
\(\Hcal_{\mathrm{tr}}\) satisfies \(\mathsf R\) but not
\(\mathsf L\), whereas \(\Hcal_{\mathrm{cof}}\) satisfies
\(\mathsf L\) but not \(\mathsf R\).

Combining all these arguments, the proofs are complete. 
\end{proof}

\begin{lemma}[A uniform mistake bound without non-uniform generation]
\label{lem:row-mistake-nonuniform}
There exists a nonempty UUS class \(\Hcal_{\mathrm{row}}\) that
admits a uniform bound of two mistakes but is not non-uniformly
generatable.
\end{lemma}

\begin{proof}
Let
$$
    \Xcal=\bigsqcup_{i\geq1}R_i,
    \qquad R_i=\{r_{i,1},r_{i,2},\ldots\}.
$$
For each nonempty \(S\subseteq\mathbb N\) and each function
\(f:S\to\mathbb N\), define
$$
    \supp(h_{S,f})=
    \bigcup_{i\in S}\bigl(R_i\setminus\{r_{i,f(i)}\}\bigr),
    \qquad
    \Hcal_{\mathrm{row}}=
    \{h_{S,f}:\varnothing\neq S\subseteq\mathbb N,
                     \ f:S\to\mathbb N\}.
$$
This class is nonempty and UUS, since each language contains an infinite
row with one point removed.

To obtain a uniform mistake bound, output any point on the first round.
If the first demonstration lies in \(R_i\), subsequently output points
of \(R_i\) avoiding all previous outputs and demonstrations. Such points
always exist. For any target realizing the demonstrations, \(i\in S\),
so only one point of this row is invalid. There is at most one mistake
after the first round, and hence at most two mistakes in total.

To disprove \(\mathsf N\), fix a generator \(G\). We construct a
\emph{single} target for which no cutoff works uniformly over its
realizable plays. For every \(i\geq1\), let
$$
    P_i=(r_{i,1},\ldots,r_{i,i}),\qquad y_i=G(P_i).
$$
If \(y_i\in R_i\), write \(y_i=r_{i,q_i}\); legality implies \(q_i>i\).
Form a simple undirected graph on \(\mathbb N\) by adding the edge
\(\{i,j\}\) whenever \(y_i\in R_j\) and \(j\neq i\).
Each edge can be oriented from one of the indices that generated it, so
that every vertex has outdegree at most one. Every finite subgraph
therefore has at most as many edges as vertices. Consequently, each
connected component has at most one cycle: two cycles in one component,
together with a finite connecting path, would give a finite subgraph
with more edges than vertices. Each component is thus a tree or a
unicyclic graph and admits a proper three-coloring. Choose an infinite
color class \(S\). If \(i\in S\) and \(y_i\in R_j\) with \(j\neq i\),
then \(j\notin S\).

Define \(h^\star=h_{S,f}\) by
$$
    f(i)=
    \begin{cases}
        q_i,&y_i\in R_i,\\
        i+1,&y_i\notin R_i,
    \end{cases}
    \qquad i\in S.
$$
For every \(i\in S\), the prefix \(P_i\) is positive for this same
\(h^\star\), since \(f(i)>i\). Its next output \(y_i\) is negative:
either it is the excluded point of \(R_i\), or it belongs to a row
outside \(S\). Given any \(B\in\mathbb N_0\), choose \(i\in S\) with
\(i>B\) and extend \(P_i\) to an infinite play realizable by
\(h^\star\). The output on round \(i+1>B\) is invalid. Thus for this
fixed target there is no valid cutoff, and \(G\) cannot witness
\(\mathsf N\). Since \(G\) was arbitrary, \(\Hcal_{\mathrm{row}}\)
satisfies \(\mathsf M\) but not \(\mathsf N\).
\end{proof}

\begin{lemma}[Realizable no-regret generation without generation in the limit]
\label{lem:transversal-regret-limit}
There exists a nonempty UUS class \(\Hcal_{\mathrm{tr}}\) that
is realizably no-regret generatable but is not generatable in the limit.
Moreover, for every horizon \(T\geq1\), it admits a generator
making at most one mistake on every realizable play of length \(T\).
\end{lemma}

\begin{proof}
Partition a countably infinite space into finite blocks
$$
    \Xcal=\bigsqcup_{n\geq1}B_n,\qquad |B_n|=n+1,
$$
and define
$$
    \Hcal_{\mathrm{tr}}=
    \{h_R:\supp(h_R)=\Xcal\setminus R,\quad
            |R\cap B_n|\leq1\text{ for every }n\}.
$$
The class is nonempty and UUS, since every language contains at least
one point from each block. Unlike \(\Hcal_\infty\), this class allows \(R\) to be infinite,
which lets the construction below reserve an invalid point in each new
block visited by the generator while keeping a single target in the class.

For a fixed horizon \(T\), choose a block
with at least \(2T\) points and output distinct points of that block,
avoiding all previous demonstrations. Before round \(t\leq T\), at
most \(2(t-1)\) points are excluded, so this strategy is legal.
Every target excludes at most one point of the block; hence the strategy
makes at most one mistake. Theorems~\ref{thm:mf-finite-characterization}
and~\ref{thm:horizon-anytime-main} give \(\mathsf R\).

Now we show it is not generatable in the limit. Fix an arbitrary generator \(G\) and an enumeration
\(u_1,u_2,\ldots\) of \(\Xcal\). Starting with \(R_0=\varnothing\),
construct the play recursively. On round \(t\), let
\(\widehat x_t=G(x_{1:t-1})\). If its block has no point in
\(R_{t-1}\), set \(R_t=R_{t-1}\cup\{\widehat x_t\}\); otherwise set
\(R_t=R_{t-1}\). Reveal the first point in the fixed enumeration that
does not belong to
$$
    R_t\cup\{x_1,\ldots,x_{t-1}\}.
$$
This point exists because the displayed set is finite. Put
\(R_\infty=\bigcup_{t\geq0}R_t\). At most one point is reserved in
each block, so \(h^\star=h_{R_\infty}\) belongs to
\(\Hcal_{\mathrm{tr}}\). A demonstration is outside the reserve when
revealed and cannot be reserved later: any later addition is a legal
generator output and hence is not a previous demonstration. Thus all
demonstrations belong to \(\supp(h^\star)\).

They also enumerate the entire support. Indeed, fix
\(u_k\notin R_\infty\). Until it is demonstrated, it is always
eligible for selection. The priority rule can select a point ahead of
it at most \(k-1\) times, since demonstrations are distinct. Therefore
\(u_k\) is eventually demonstrated.

Suppose, for a contradiction, that \(G\) makes only finitely many
mistakes on this enumeration. The first output in each new block is
reserved and is a mistake under \(h^\star\); therefore only finitely
many blocks are visited. After the last mistake, every output must be
a nonreserved point in their finite union. All such points are
eventually demonstrated, after which none is a legal output. This is
a contradiction. Thus \(G\) makes infinitely many mistakes on an
enumeration of the single target \(h^\star\). Since \(G\) was
arbitrary, \(\mathsf L\) fails for \(\Hcal_{\mathrm{tr}}\).
\end{proof}

\subsection{Proofs for Section~\ref{subsec:noisy-versus-agnostic}}

\begin{proposition*}[Restatement of Proposition~\ref{prop:agnostic-contamination-transfer}]
Suppose a deterministic generator \(G\) and a function
\(r:\mathbb N\to[0,\infty)\) satisfy
\[
  \sup_{\substack{x_{1:T}\\h\in\Hcal}}
  \Reg_T(G,h;x_{1:T})
  \leq r(T)
  \qquad
  \text{for every }T\geq1.
\]
Then, for every \(h\in\Hcal\) and every demonstration sequence,
\[
  M_T(G,h;x_{1:T})
  \leq
  r(T)+\sum_{t=1}^T\ell_h(x_t).
\]
Consequently, if \(r(T)=o(T)\) and the demonstrations contain \(o(T)\)
invalid examples relative to a target \(h^\star\), then the generator
produces \(o(T)\) invalid examples relative to the same target.
\end{proposition*}

\begin{proof}
For every comparator \(h\), the definition of regret gives the exact
identity
\[
  M_T(G,h;x_{1:T})
  =
  \Reg_T(G,h;x_{1:T})
  +
  \sum_{t=1}^T\ell_h(x_t).
\]
The assumed regret bound proves the displayed inequality.  If both terms on
its right-hand side are \(o(T)\), then their sum is also \(o(T)\), which
proves the final assertion.
\end{proof}

\subsection{Proofs for Section~\ref{subsec:finite-countable}}

\begin{theorem*}[Restatement of Theorem~\ref{thm:finite-class-bounds}]
If \(1\leq|\Hcal|=N<\infty\), then
\[
  \MFUnif(\Hcal)\leq\lfloor\log N\rfloor
\]
and
\[
  \SMF_T(\Hcal)\leq\sqrt{8T\log N}
  \qquad
  \text{for every }T\geq1.
\]
\end{theorem*}
\begin{proof}
If \(N=1\), both claims are immediate, so suppose \(N>1\).

Identify each hypothesis \(h\in\Hcal\) with its support
\(\supp(h)\).  Theorem~5.1 of
\cite{kleinberg2026mistakebounded} provides an anytime generator whose
total number of realizable mistakes is at most
\(\lfloor\log N\rfloor\).  The uniform total-mistake characterization,
Theorem~\ref{thm:mf-infinite-characterization}, therefore gives
\[
    \MFUnif(\Hcal)
    \leq
    \lfloor\log N\rfloor.
\]

For the agnostic bound, Lemma~7.1 of
\cite{kleinberg2026mistakebounded} shows that, for every
\(\gamma\in(0,3/4]\), there exists a generator \(G_\gamma\) such that,
for every demonstration sequence \(x_{1:T}\) and every \(h\in\Hcal\),
\[
    M_T(G_\gamma,h;x_{1:T})
    \leq
    (1+2\gamma)
    \sum_{t=1}^T\ell_h(x_t)
    +
    \frac{\log N}{\gamma}.
\]
Subtracting the comparator's loss on the demonstrations gives
\[
    \Reg_T(G_\gamma,h;x_{1:T})
    \leq
    2\gamma T
    +
    \frac{\log N}{\gamma}.
\]

If \(\log N\leq 9T/8\), take
\[
    \gamma
    :=
    \sqrt{\frac{\log N}{2T}}
    \leq
    \frac34.
\]
Then
\[
    \Reg_T(G_\gamma,h;x_{1:T})
    \leq
    \sqrt{8T\log N}.
\]
If \(\log N>9T/8\), the trivial bound
\[
    \SMF_T(\Hcal)\leq T
\]
is already smaller than \(\sqrt{8T\log N}\).  The finite-horizon agnostic
characterization, Theorem~\ref{thm:smf-characterization}, therefore yields
\[
    \SMF_T(\Hcal)
    \leq
    \sqrt{8T\log N}
\]
for every \(T\geq1\).
\end{proof}

\begin{proposition*}[Restatement of Proposition~\ref{prop:cofinite-linear-profile}]
There exists a class \(\Hcal_{\mathrm{cof}}\) which is countable, UUS, and non-uniformly
generatable. Nevertheless,
\[
  \MF_T(\Hcal_{\mathrm{cof}})
  =
  \SMF_T(\Hcal_{\mathrm{cof}})
  =T
  \qquad
  \text{for every }T\geq1.
\]
Consequently, \(\Hcal_{\mathrm{cof}}\) is neither realizably nor
agnostically no-regret generatable.
\end{proposition*}

\begin{proof}

Fix an
enumeration
\[
    \Xcal=\{z_1,z_2,\ldots\}.
\]
For every finite \(E\subseteq\Xcal\), define \(h_E\) by
\[
    \supp(h_E):=\Xcal\setminus E,
\]
and let
\[
    \Hcal_{\mathrm{cof}}
    :=
    \{h_E:E\subseteq\Xcal,\ |E|<\infty\}.
\]

The set of finite subsets of a countable set is countable, so
\(\Hcal_{\mathrm{cof}}\) is countable. Every cofinite subset of the infinite
space \(\Xcal\) is infinite, and hence the class is UUS.

We next give an explicit non-uniform generator. At round \(t\), after seeing
the demonstrations \(x_{1:t-1}\), let
\[
  k_t
  :=
  \min\{k\geq t:z_k\notin\{x_1,\ldots,x_{t-1}\}\},
\]
and output \(z_{k_t}\). This is a legal output, and the minimum exists because
only finitely many points have been demonstrated. Fix a target
\(h^\star:=h_E\), and put
\[
  B_E
  :=
  \max\bigl(\{k:z_k\in E\}\cup\{0\}\bigr).
\]
For every \(t>B_E\), we have \(k_t\geq t>B_E\), so
\(z_{k_t}\notin E\). Thus no mistake occurs after round \(B_E\), proving
non-uniform generatability.

Now we prove this class is not no-regret generatable.
Fix now a horizon \(T\) and an arbitrary horizon-\(T\) generator. Construct
the demonstrations recursively. After the generator outputs \(\widehat x_t\),
choose
\[
  x_t
  \notin
  \{x_1,\ldots,x_{t-1},\widehat x_1,\ldots,\widehat x_t\}.
\]
Such a point exists because \(\Xcal\) is infinite. At the end of the play,
let
\[
  E:=\{\widehat x_1,\ldots,\widehat x_T\}.
\]
Set \(h^\star:=h_E\).  The construction ensures that every demonstration lies in
\(\supp(h^\star)\), while every generated example lies outside it. The play is
therefore realizable by \(h^\star\), and the generator makes \(T\) mistakes.
Since no length-\(T\) play can contain more than \(T\) mistakes, the
finite-horizon characterization yields
\[
  \MF_T(\Hcal_{\mathrm{cof}})=T.
\]
Finally,
\[
  \MF_T(\Hcal_{\mathrm{cof}})
  \leq
  \SMF_T(\Hcal_{\mathrm{cof}})
  \leq T,
\]
where the first inequality is Proposition~\ref{prop:mf-le-smf}. The two linear profiles rule out
realizable and agnostic no-regret generation.
\end{proof}

\section{Proofs of Theorems~\ref{thm:finite-ldim-realizable}
and~\ref{thm:finite-ldim-agnostic}}
\label{app:finite-ldim-upper}

\subsection{Weighted sequential leaf capacity}

A class of finite Littlestone dimension cannot maintain many competing
possibilities indefinitely: after sufficiently many binary splits, some of
those possibilities must merge or disappear. We
turn this limitation into a quantity that measures the remaining complexity
of the class.  Later we will construct a generator that chooses its outputs so that this quantity
cannot grow too quickly.  This controls both the number of realizable
mistakes and the agnostic regret.

For a class \(\Vcal\subseteq\{0,1\}^{\Xcal}\), a bounded weight function
\(w:\Vcal\to[0,\infty)\), a point \(z\in\Xcal\), and
\(b\in\{0,1\}\), write
\[
  \Vcal^b(z):=\{h\in\Vcal:h(z)=b\}.
\]
Define the weighted sequential leaf capacity recursively by
\[
\begin{aligned}
  \LeafCap_0(\Vcal,w)
  &:=\sup_{h\in\Vcal}w(h),\\
  \LeafCap_n(\Vcal,w)
  &:=\sup_{z\in\Xcal}
  \left(
    \LeafCap_{n-1}(\Vcal^0(z),w)
    +
    \LeafCap_{n-1}(\Vcal^1(z),w)
  \right).
\end{aligned}
\]
The empty class has capacity zero.

The recursion can be represented by an adaptive binary tree.  At each
internal node, choose a point \(z\in\Xcal\) and split the current hypothesis
class into those satisfying \(h(z)=0\) and those satisfying \(h(z)=1\).
The point chosen at the next node may depend on the labels encountered
earlier, so different branches may use different points.

After \(n\) such splits, every leaf contains the hypotheses consistent with
the labels along its path.  The value of a leaf is the largest weight of a
hypothesis reaching it.  Then \(\LeafCap_n(\Vcal,w)\) is the largest possible
sum of the leaf values over all adaptive trees of depth \(n\).  In
particular, when \(w\equiv1\), it is simply the largest possible number of
nonempty leaves.

The capacity is positively homogeneous: for every \(c\geq0\),
\[
  \LeafCap_n(\Vcal,cw)
  =
  c\,\LeafCap_n(\Vcal,w),
\]
where \((cw)(h):=c\,w(h)\).  This follows directly by induction on \(n\):
the identity holds for the supremum at depth zero, and the recursive sum
preserves multiplication by \(c\) at every later depth.

It is also subadditive over disjoint
unions: if \(\Acal\cap\Bcal=\varnothing\), then
\[
  \LeafCap_n(\Acal\cup\Bcal,w)
  \leq
  \LeafCap_n(\Acal,w)+\LeafCap_n(\Bcal,w).
\]
Indeed, the claim for \(n=0\) follows from
\(\sup_{\Acal\cup\Bcal}w\leq\sup_{\Acal}w+\sup_{\Bcal}w\).  For
\(n\geq1\), we just need to apply the induction hypothesis to the two label restrictions
at each possible root query and then take the supremum over that query.

The following lemma is the sequential analogue of the classical
Sauer--Shelah lemma; see \cite{bhaskar2021thicket}.  We include its
proof because its recursive structure will also be used below.

\begin{lemma}[Sequential Sauer--Shelah bound]
\label{lem:app-dynamic-sauer}
If \(\Ldim(\Vcal)\leq d\), then
\[
  \LeafCap_n(\Vcal,\one)
  \leq
  \sum_{i=0}^{\min\{d,n\}}\binom ni.
\]
\end{lemma}

\begin{proof}
Let
\[
  S(n,d):=\sum_{i=0}^{\min\{d,n\}}\binom ni.
\]
We prove the assertion by induction on \(n+d\).  If \(\Vcal\) is empty,
there is nothing to prove.  If \(n=0\), then
\(\LeafCap_0(\Vcal,\one)=1=S(0,d)\).  If \(d=0\), no point can induce two
nonempty label restrictions; otherwise that point would itself form a
depth-one shattered tree.  Consequently each query has at most one
nonempty child, and induction on \(n\) gives
\(\LeafCap_n(\Vcal,\one)\leq1=S(n,0)\).

Now suppose \(n,d\geq1\), and fix a possible root query \(z\).  Splitting \(\Vcal\) according to the label assigned to \(z\) gives the two
subclasses
\[
  \Vcal^0(z)=\{h\in\Vcal:h(z)=0\},
  \qquad
  \Vcal^1(z)=\{h\in\Vcal:h(z)=1\}.
\]
Each subclass has Littlestone dimension at most \(d\).  Moreover, they
cannot both have dimension \(d\).  Otherwise, each subclass would shatter
a tree of depth \(d\); placing \(z\) at the root and attaching these two
trees to its zero and one branches would give a depth-\((d+1)\) tree
shattered by \(\Vcal\). After interchanging the labels if
necessary, we may therefore write
\[
  \Ldim(\Vcal^0(z))\leq d-1,
  \qquad
  \Ldim(\Vcal^1(z))\leq d.
\]
The induction hypothesis applies to both depth-\((n-1)\) capacities, so
Pascal's identity yields
\[
\begin{aligned}
  &\LeafCap_{n-1}(\Vcal^0(z),\one)
  +\LeafCap_{n-1}(\Vcal^1(z),\one)\\
  &\qquad\leq
  S(n-1,d-1)+S(n-1,d)
  =S(n,d).
\end{aligned}
\]
This bound holds for every \(z\).  Taking the supremum in the definition of
\(\LeafCap_n\) completes the induction.
\end{proof}

\subsection{Proof of the realizable upper bound}
The next lemma captures the main one-round guarantee.  One generation round
corresponds to two levels of the capacity tree: one for the generated point
and one for the demonstration.  The lemma shows that the generator can
choose its output so that, regardless of the adversary's response, the
remaining capacity does not increase after these two levels are removed.

\begin{lemma}
\label{lem:app-realizable-capacity}
Let \(n\geq2\), let \(D\subseteq\Xcal\) be nonempty, and suppose that
\(w\) takes only finitely many values. Then the maximum below is attained. 
Choose
\[
  \widehat z
  \in
  \operatorname*{arg\,max}_{z\in D}
  \LeafCap_{n-1}(\Vcal^1(z),w).
\]
For any \(x\in D\), let
\[
  \Vcal':=\Vcal^1(x)
\]
and define, for every \(h\in\Vcal'\),
\[
  w'(h):=w(h)2^{\ell_h(\widehat z)}.
\]
In other words, the update doubles the weight of every surviving hypothesis
that labels \(\widehat z\) negative and leaves all other surviving weights
unchanged.
Then
\[
  \LeafCap_{n-2}(\Vcal',w')
  \leq
  \LeafCap_n(\Vcal,w).
\]
\end{lemma}

\begin{proof}
For \(a,b\in\{0,1\}\), define
\[
  \Vcal_{ab}
  :=
  \{h\in\Vcal:h(\widehat z)=a,\ h(x)=b\},
  \qquad
  C_{ab}:=\LeafCap_{n-2}(\Vcal_{ab},w).
\]
The surviving class \(\Vcal'=\Vcal^1(x)\) consists of two disjoint parts:
\(\Vcal_{01}\), whose hypotheses label \(\widehat z\) negative, and
\(\Vcal_{11}\), whose hypotheses label \(\widehat z\) positive.  The
weights of hypotheses in \(\Vcal_{01}\) double because they regard
\(\widehat z\) as a mistake, while the weights in \(\Vcal_{11}\) remain
unchanged.  Subadditivity and homogeneity of the capacity therefore give
\[
  \LeafCap_{n-2}(\Vcal',w')
  \leq
  2C_{01}+C_{11}.
\]
It remains to show that the original capacity is at least this quantity.

Consider the branch consisting of hypotheses that label \(\widehat z\)
negative, namely \(\Vcal^0(\widehat z)\), and split this subclass according
to their labels on \(x\).  The two resulting subclasses are
\(\Vcal_{00}\) and \(\Vcal_{01}\).  Therefore, the capacity recursion gives
\[
  \LeafCap_{n-1}(\Vcal^0(\widehat z),w)
  \geq
  C_{00}+C_{01}
  \geq
  C_{01}.
\]
Next, maximality of \(\widehat z\) over \(D\), together with
\(x\in D\), implies
\[
  \LeafCap_{n-1}(\Vcal^1(\widehat z),w)
  \geq
  \LeafCap_{n-1}(\Vcal^1(x),w).
\]
Now split the subclass \(\Vcal^1(x)\) according to the hypotheses' labels
on \(\widehat z\).  The two resulting subclasses are
\(\Vcal_{01}\) and \(\Vcal_{11}\), so the capacity recursion gives
\[
  \LeafCap_{n-1}(\Vcal^1(x),w)
  \geq
  C_{01}+C_{11}.
\]
Finally, use \(\widehat z\) as the root query in the depth-\(n\)
capacity.  Adding the preceding bounds yields
\[
\begin{aligned}
  \LeafCap_n(\Vcal,w)
  &\geq
  \LeafCap_{n-1}(\Vcal^0(\widehat z),w)
  +\LeafCap_{n-1}(\Vcal^1(\widehat z),w)\\
  &\geq 2C_{01}+C_{11},
\end{aligned}
\]
which proves the claim.
\end{proof}

We now combine the preceding two lemmas.  At horizon \(T\), begin with a
capacity tree of depth \(2T\).  Each generation round uses two levels of
this tree, and Lemma~\ref{lem:app-realizable-capacity} ensures that the
remaining capacity does not increase.  After \(T\) rounds, a compatible
hypothesis on which the generator has made \(m\) mistakes has weight
\(2^m\).  Its weight is therefore bounded by the initial capacity, which
the sequential Sauer--Shelah bound controls through the Littlestone
dimension.  Taking logarithms then gives the desired mistake bound.

\begin{theorem*}[Restatement of Theorem~\ref{thm:finite-ldim-realizable}]
Let \(\Hcal\) be nonempty and UUS, and suppose
\[
  0<d:=\Ldim(\Hcal)<\infty.
\]
Then for every \(T\),
\[
  \MF_T(\Hcal)
  \leq
  \left\lfloor
    \log\left(
      \sum_{i=0}^{\min\{d,2T\}}\binom{2T}{i}
    \right)
  \right\rfloor.
\]
Consequently, for fixed \(d\),
\[
  \MF_T(\Hcal)
  =O\left(d\log\left(1+\frac{T}{d}\right)\right)
  =o(T).
\]
Moreover, there exists one deterministic anytime generator \(G\) such that,
simultaneously for every \(T\geq1\),
\[
  \sup_{\substack{x_{1:T}\\h\in\Hcal(x_{1:T})}}
  M_T(G,h;x_{1:T})
  =
  O\left(d\log\left(1+\frac{T}{d}\right)\right).
\]
\end{theorem*}

\begin{proof}
Fix a horizon \(T\). We construct a horizon-\(T\) generator \(G_T\) use Lemma~\ref{lem:app-realizable-capacity}. After \(t\) rounds, the version space is
$$
  \Vcal_t
  :=
  \{h\in\Hcal:h(x_s)=1\text{ for every }s\leq t\}.
$$
Initially, \(\Vcal_0=\Hcal\), and every hypothesis has weight one. In each round, we discard hypotheses inconsistent with the new demonstration and double the weight of each remaining hypothesis for which the generated point is a mistake, leaving all other weights unchanged. Consequently, for every \(h\in\Vcal_t\),
$$
  w_t(h)
  =
  2^{M_t(G_T,h;x_{1:t})}.
$$

At the start of round \(t+1\), define
\[
  D_t:=\Xcal\setminus\{x_1,\ldots,x_t\},
  \qquad
  n_t:=2(T-t).
\]
Choose
\[
  \widehat x_{t+1}
  \in
  \operatorname*{arg\,max}_{z\in D_t}
  \LeafCap_{n_t-1}(\Vcal_t^1(z),w_t).
\]

Now consider any realizable response
\(x_{t+1}\in D_t\).  The update is
\[
  \Vcal_{t+1}=\Vcal_t^1(x_{t+1}),
  \qquad
  w_{t+1}(h)
  =w_t(h)2^{\ell_h(\widehat x_{t+1})}.
\]
Since \(n_t\geq2\) for \(t<T\),
Lemma~\ref{lem:app-realizable-capacity} applies with
\(\widehat z=\widehat x_{t+1}\) and \(x=x_{t+1}\), giving
\[
  \LeafCap_{n_t-2}(\Vcal_{t+1},w_{t+1})
  \leq
  \LeafCap_{n_t}(\Vcal_t,w_t).
\]
Because \(n_t-2=2(T-(t+1))\), these inequalities telescope over the
\(T\) rounds.  At the terminal time the remaining capacity has depth zero,
so
\[
\begin{aligned}
  \sup_{h\in\Vcal_T}2^{M_T(G_T,h;x_{1:T})}
  &=\LeafCap_0(\Vcal_T,w_T)\\
  &\leq\LeafCap_{2T}(\Hcal,\one).
\end{aligned}
\]
Applying Lemma~\ref{lem:app-dynamic-sauer}
shows that every \(h\in\Hcal(x_{1:T})=\Vcal_T\) satisfies
\[
  M_T(G_T,h;x_{1:T})
  \leq
  \log\left(
    \sum_{i=0}^{\min\{d,2T\}}\binom{2T}{i}
  \right).
\]
This proves the desired upper bound. It remains to simplify the binomial expression.  If \(d\leq2T\), the
standard estimate gives
\[
  \sum_{i=0}^{d}\binom{2T}{i}
  \leq
  \left(\frac{2eT}{d}\right)^d,
\]
whose base-two logarithm is
\(O(d\log(1+T/d))\).  If \(d>2T\), the binomial sum is \(2^{2T}\); in
this regime \(T/d<1\), and the elementary lower bound
\(\log(1+T/d)\geq T/d\) shows that \(2T=O(d\log(1+T/d))\).  Therefore
\[
  \MF_T(\Hcal)
  =O\left(d\log\left(1+\frac{T}{d}\right)\right).
\]
For fixed \(d\), this is \(O(d\log T)=o(T)\).

Finally, choose a sufficiently large universal constant \(C\) and set
\[
  b(T):=C d\log\left(1+\frac{T}{d}\right).
\]
The preceding estimate makes \(b\) an upper bound on \(\MF_T(\Hcal)\).  It is
also nondecreasing and unbounded.  We therefore can use the quantitative conclusion
of Theorem~\ref{thm:horizon-anytime-main}: that result converts this
finite-horizon envelope into one deterministic anytime generator with loss
at most \(4b(T)\) simultaneously for all \(T\).  This is the asserted
anytime rate.
\end{proof}

\subsection{Proof of the agnostic upper bound}

The agnostic setting has the same two-level structure: one level corresponds
to the generated point and the other to the demonstration.  The difference
is that the demonstration need not be positive for a comparator, so no
hypothesis can be removed.  Instead, we increase the weights of comparators
that outperform the generator and decrease the weights of those that do
worse.  The next lemma shows that this update can again be controlled by
two levels of the capacity tree.

\begin{lemma}
\label{lem:app-agnostic-capacity}
Let \(n\geq2\), let \(D\subseteq\Xcal\) be nonempty, and let
\(\eta\in[0,1]\).  Suppose that \(w\) takes only finitely many values.
Then the maximum below is attained.  Choose
\[
  \widehat z
  \in
  \operatorname*{arg\,max}_{z\in D}
  \LeafCap_{n-1}(\Vcal^1(z),w).
\]
For every \(x\in D\), define
\[
  w'(h)
  :=
  w(h)\bigl(1+\eta(h(x)-h(\widehat z))\bigr).
\]
Thus the weight of \(h\) increases when \(h\) outperforms the generator,
decreases when it performs worse, and otherwise remains unchanged.

Then
\[
  \LeafCap_{n-2}(\Vcal,w')
  \leq
  \LeafCap_n(\Vcal,w).
\]
\end{lemma}

\begin{proof}

For \(a,b\in\{0,1\}\), let
\[
  \Vcal_{ab}
  :=
  \{h\in\Vcal:h(\widehat z)=a,\ h(x)=b\},
  \qquad
  C_{ab}:=\LeafCap_{n-2}(\Vcal_{ab},w).
\]
The weight update leaves the weights unchanged on
\(\Vcal_{00}\) and \(\Vcal_{11}\), multiplies them by \(1+\eta\) on
\(\Vcal_{01}\), and multiplies them by \(1-\eta\) on
\(\Vcal_{10}\).  Since these four subclasses are disjoint, subadditivity
and homogeneity give
\[
  \LeafCap_{n-2}(\Vcal,w')
  \leq
  C_{00}+(1+\eta)C_{01}+(1-\eta)C_{10}+C_{11}.
\]
To compare this expression with the old capacity, we derive two lower
bounds on \(\LeafCap_n(\Vcal,w)\).  The first counts every subclass once,
while the second gives an additional copy of \(C_{01}\).  Their convex
combination will reproduce the coefficients in the updated capacity.

For the first bound, we split \(\Vcal\) according to the hypotheses' labels on
\(\widehat z\), and then split each of the two resulting subclasses
according to their labels on \(x\).  The four resulting subclasses are
\(\Vcal_{00},\Vcal_{01},\Vcal_{10}\), and \(\Vcal_{11}\).  Applying the
capacity recursion twice gives
\[
  \LeafCap_n(\Vcal,w)
  \geq
  C_{00}+C_{01}+C_{10}+C_{11}.
\]

For the second bound, again split \(\Vcal\) according to the label on
\(\widehat z\).  Within \(\Vcal^0(\widehat z)\), splitting according to
the label on \(x\) gives
\[
  \LeafCap_{n-1}(\Vcal^0(\widehat z),w)
  \geq
  C_{00}+C_{01}.
\]
For the other subclass, the maximizing choice of \(\widehat z\) gives
\[
  \LeafCap_{n-1}(\Vcal^1(\widehat z),w)
  \geq
  \LeafCap_{n-1}(\Vcal^1(x),w).
\]
Splitting \(\Vcal^1(x)\) according to the label on \(\widehat z\) gives
\[
  \LeafCap_{n-1}(\Vcal^1(x),w)
  \geq
  C_{01}+C_{11}.
\]
Adding these bounds yields
\[
  \LeafCap_n(\Vcal,w)
  \geq
  C_{00}+2C_{01}+C_{11}.
\]
Finally,
\[
\begin{aligned}
  &C_{00}+(1+\eta)C_{01}+(1-\eta)C_{10}+C_{11}\\
  &\quad=(1-\eta)
  (C_{00}+C_{01}+C_{10}+C_{11})
  +\eta(C_{00}+2C_{01}+C_{11}).
\end{aligned}
\]
Since \(\eta\in[0,1]\), this convex combination is at most
\(\LeafCap_n(\Vcal,w)\), proving the lemma.
\end{proof}

We now apply the same capacity framework to agnostic regret.  No comparator
is removed after a demonstration; instead, its weight records how well it
has performed relative to the generator.  The preceding lemma ensures that
the total weighted complexity remains controlled, while a comparator with
large regret would acquire a large weight.  Combining these observations
with the sequential Sauer--Shelah bound yields the following result.

\begin{theorem*}[Restatement of Theorem~\ref{thm:finite-ldim-agnostic}]
Let \(\Hcal\) be nonempty and UUS, and suppose
\[
  0<d:=\Ldim(\Hcal)<\infty.
\]
For every \(T\) and every \(\eta\in(0,1/2]\), there is a deterministic
generator such that
\[
  \SMF_T(\Hcal)
  \leq
  \frac{1}{\eta}
  \log\left(
    \sum_{i=0}^{\min\{d,2T\}}\binom{2T}{i}
  \right)
  +\eta T.
\]
Consequently, for fixed \(d\),
\[
  \SMF_T(\Hcal)
  =O\left(\sqrt{dT\log\left(1+\frac{T}{d}\right)}
  \right)
  =o(T).
\]
Moreover, there exists one deterministic anytime generator \(G\) such that,
simultaneously for every \(T\geq1\),
\[
  \sup_{\substack{x_{1:T}\\h\in\Hcal}}
  \Reg_T(G,h;x_{1:T})
  = O\left(\sqrt{dT\log\left(1+\frac{T}{d}\right)}
  \right).
\]
\end{theorem*}

\begin{proof}
  Fix \(T\) and \(\eta\in(0,1/2]\), and put
  \[
    \lambda:=\frac{\eta}{\log e}.
  \]
  Since \(\log\) is base two, \(\log e>1\), and hence
  \(\lambda\in(0,1/2]\).  For each \(h\in\Hcal\), define
\[
  w_t(h)
  :=
  \prod_{s=1}^{t}
  \bigl(1+\lambda(h(x_s)-h(\widehat x_s))\bigr).
\]
The signed increment \(h(x_s)-h(\widehat x_s)\) is exactly the one-round
regret
\[
  \ell_h(\widehat x_s)-\ell_h(x_s)
  =h(x_s)-h(\widehat x_s).
\]
Thus a comparator with positive cumulative advantage tends to acquire a
large weight.  All factors lie in
  \(\{1-\lambda,1,1+\lambda\}\), so the weights are positive and, at each fixed
time \(t\), take only finitely many values.

As in the realizable proof, put
\[
  D_t:=\Xcal\setminus\{x_1,\ldots,x_t\},
  \qquad
  n_t:=2(T-t),
\]
and choose
\[
  \widehat x_{t+1}
  \in
  \operatorname*{arg\,max}_{z\in D_t}
  \LeafCap_{n_t-1}(\Hcal^1(z),w_t).
\]
The same finite-range argument used above shows that the maximizer exists.
The UUS assumption ensures that \(D_t\) is nonempty.

The point at which this argument differs from the realizable one is the
update: the adversary's response does not remove any comparator.  Instead,
for every \(h\in\Hcal\),
\[
  w_{t+1}(h)
  =w_t(h)
  \bigl(1+\lambda(h(x_{t+1})-h(\widehat x_{t+1}))\bigr).
  \]
  Lemma~\ref{lem:app-agnostic-capacity}, used with parameter \(\lambda\),
  therefore gives, for every legal
response \(x_{t+1}\in D_t\),
\[
  \LeafCap_{n_t-2}(\Hcal,w_{t+1})
  \leq
  \LeafCap_{n_t}(\Hcal,w_t).
\]
Since \(n_t-2=2(T-(t+1))\), the potential telescopes exactly as in the
realizable argument:
\[
\begin{aligned}
  \sup_{h\in\Hcal}w_T(h)
  &=\LeafCap_0(\Hcal,w_T)\\
  &\leq\LeafCap_{2T}(\Hcal,\one).
\end{aligned}
\]

  We now convert this multiplicative estimate into regret.  For
  \(u\in[-1/2,1/2]\), the base-two logarithm satisfies
  \[
    \log(1+u)\geq(\log e)(u-u^2).
  \]
  Applying this inequality to every factor in
\(w_T(h)\) gives, for each comparator \(h\),
\[
\begin{aligned}
  \log w_T(h)
  &\geq
  (\log e)\lambda\sum_{t=1}^{T}
  \bigl(h(x_t)-h(\widehat x_t)\bigr)
  -(\log e)\lambda^2\sum_{t=1}^{T}
  \bigl(h(x_t)-h(\widehat x_t)\bigr)^2\\
  &\geq
  \eta\Reg_T(G_T,h;x_{1:T})-\eta^2T.
\end{aligned}
\]
  Here we used \((\log e)\lambda=\eta\) and
  \((\log e)\lambda^2=\eta^2/(\log e)\leq\eta^2\).
  Combining the lower bound on \(\log w_T(h)\) with the capacity upper bound
and dividing by \(\eta\) yields
\[
  \Reg_T(G_T,h;x_{1:T})
  \leq
  \frac{\log\LeafCap_{2T}(\Hcal,\one)}{\eta}
  +\eta T.
  \]
Because \(\Hcal\) is nonempty,
\(\LeafCap_{2T}(\Hcal,\one)\geq1\).  Moreover,
Lemma~\ref{lem:app-dynamic-sauer} gives
\[
  \LeafCap_{2T}(\Hcal,\one)
  \leq
  \sum_{i=0}^{\min\{d,2T\}}\binom{2T}{i}.
\]
Substituting this estimate into the preceding regret bound yields, together with Theorem~\ref{thm:smf-characterization}, gives
\[
  \SMF_T(\Hcal)
  \leq
  \frac{1}{\eta}
  \log\left(
    \sum_{i=0}^{\min\{d,2T\}}\binom{2T}{i}
  \right)
  +\eta T.
\]

The binomial estimate from the realizable proof gives
\[
  \log\left(
    \sum_{i=0}^{\min\{d,2T\}}\binom{2T}{i}
  \right)
  =
  O\left(
    d\log\left(1+\frac{T}{d}\right)
  \right).
\]
Choose
\[
  \eta
  :=
  \min\left\{
    \frac12,
    \sqrt{\frac{d\log(1+T/d)}{T}}
  \right\}.
\]
Substituting this choice into the preceding bound gives
\[
  \SMF_T(\Hcal)
  =
  O\left(
    \sqrt{dT\log\left(1+\frac{T}{d}\right)}
    +
    d\log\left(1+\frac{T}{d}\right)
  \right).
\]
For fixed \(d\), the second term is eventually dominated by the first, so
\[
  \SMF_T(\Hcal)
  =
  O\left(\sqrt{dT\log\left(1+\frac{T}{d}\right)}
  \right)
  =
  o(T).
\]

Finally, after increasing the implicit constant if necessary, the
right-hand side defines a nondecreasing and unbounded finite-horizon
envelope.  The quantitative conclusion of
Theorem~\ref{thm:agnostic-horizon-anytime-main} converts this envelope into
one anytime generator, proving the last assertion simultaneously for all
horizons.
\end{proof}

\section{Proof of Theorem~\ref{thm:sharp-ldim-generation}}
\label{app:finite-ldim-sharpness}
This appendix proves the sharpness result Theorem~\ref{thm:sharp-ldim-generation} in three steps.  We first
construct a basic class of Littlestone dimension one that can force
logarithmically many mistakes.  We then combine several independent copies of this construction so that
both the number of forced mistakes and the Littlestone dimension add. Finally, we place constructions of all relevant
sizes into one countable class, allowing the adversary to select the scale
appropriate for the given horizon.  This gives the desired lower bound, and
the matching upper bound follows from Appendix~\ref{app:finite-ldim-upper}.

\begin{lemma}
\label{lem:app-single-component-forcing}
For every integer \(m\geq1\), there is a finite UUS class
\(\mathcal C_m\) such that
\[
  \Ldim(\mathcal C_m)=1,
  \qquad
  \MF_{2^m-1}(\mathcal C_m)\geq m.
\]
\end{lemma}

\begin{proof}
Put \(K:=2^m\).  Let
\(X:=\{x_1,\ldots,x_K\}\), and let
\(A_1,\ldots,A_K\) be pairwise disjoint countably infinite sets, all
disjoint from \(X\).  On the example space
\(X\cup\bigcup_{i=1}^K A_i\), define
\[
  \supp(h_i)
  :=
  (X\setminus\{x_i\})\cup A_i,
  \qquad
  \mathcal C_m:=\{h_i:i\in[K]\}.
\]
We call a point $x_i\in X$ to be a \emph{core point}. Every hypothesis has infinite support, so \(\mathcal C_m\) is UUS.

We first calculate its Littlestone dimension.  A core point \(x_i\) has
zero restriction \(\{h_i\}\), while a point in \(A_i\) has one
restriction \(\{h_i\}\).  These are the only nonconstant types of split,
so every nonconstant split has a singleton child.  Consequently no query
can have two children of Littlestone dimension at least one, and no
depth-two tree is shattered.  On the other hand, two distinct hypotheses
differ on a core point, so a depth-one tree is shattered.  Hence
\(\Ldim(\mathcal C_m)=1\).

It remains to prove the generation lower bound.  Fix an arbitrary
deterministic horizon-\((K-1)\) generator.  The adversary maintains a set \(S\subseteq[K]\) of hypotheses that remain
possible targets.  Initially \(S=[K]\).  Whenever the adversary reveals a
core point \(x_i\), the hypothesis \(h_i\), which labels \(x_i\) negative,
is removed from \(S\).  Hence \(i\in S\) precisely when \(x_i\) has not yet
been revealed, and every hypothesis indexed by \(S\) labels all
demonstrations so far positive.
Assign each remaining candidate \(h_i\) the weight
\[
w_i:=2^{c_i},
\]where \(c_i\) is the number of mistakes made so far relative to \(h_i\).
Thus the weight of a candidate doubles whenever the generator makes another
mistake for that candidate.  We construct an adversary strategy that will maintain
\[
W:=\sum_{i\in S}w_i\geq K.
\]Initially, all \(K\) candidates remain and have weight one, so \(W=K\).
As candidates are eliminated, this invariant forces their weight to
concentrate on the surviving candidates.
The adversary stops as soon as some surviving candidate has weight at least
\(K\).  Until then, every surviving weight is a power of two strictly
smaller than \(K\).

The adversary always responds with a fresh core point, thereby removing one
candidate from \(S\).  We choose this response so that the invariant
\(W\geq K\) is preserved.  Every legal generator output is either a core
point or a point in one of the private tails, leading to the following four
cases.

\textbf{Case 1.}
Suppose the generator outputs a core point \(x_i\), and there exists
\(j\in S\setminus\{i\}\) with \(w_j\leq w_i\).  Reveal \(x_j\).  This
removes candidate \(j\), while candidate \(i\) remains and its weight
doubles because \(h_i(x_i)=0\).  Hence
\[
  W'
  =
  W-w_j+w_i
  \geq
  W
  \geq
  K.
\]

\textbf{Case 2.}
Suppose the generator outputs \(x_i\), but \(w_i\) is the unique smallest
weight in \(S\).  Reveal \(x_i\), thereby removing candidate \(i\).

Every other weight is a power of two strictly larger than \(w_i\), and is
therefore a multiple of \(2w_i\).  Consequently, \(W/w_i\) is odd.  On the
other hand, \(w_i<K\), so \(K/w_i\) is even.  Thus \(W\neq K\).  Since
\(W\geq K\) and both quantities are multiples of \(w_i\),
\[
  W\geq K+w_i.
\]
It follows that
\[
  W'
  =
  W-w_i
  \geq
  K.
\]

\textbf{Case 3.}
Suppose the generator outputs a point in \(A_i\) with \(i\notin S\).
This output is invalid for every surviving candidate.  Choose an index
\(j\in S\) of minimum weight and reveal \(x_j\).  Candidate \(j\) is
removed, while the weights of all remaining candidates double.  Therefore
\[
  W'
  =
  2(W-w_j).
\]
Before the stopping condition is reached, \(S\) contains at least two
indices; otherwise its only remaining weight would equal \(W\geq K\).
Thus \(w_j\leq W/2\), and hence
\[
  W'\geq W\geq K.
\]

\textbf{Case 4.}
Suppose the generator outputs a point in \(A_i\) with \(i\in S\).  Let
\[
  U
  :=
  \sum_{j\in S\setminus\{i\}}w_j.
\]
We claim that \(w_i\leq U\).  Otherwise,
\[
  W=w_i+U<2w_i\leq K,
\]
where the last inequality follows because \(w_i\) is a power of two
strictly smaller than \(K\).  This contradicts \(W\geq K\).

Reveal \(x_i\), removing the only surviving candidate for which the
generated tail point is valid.  Every remaining candidate regards that
output as a mistake, so all remaining weights double.  Therefore
\[
  W'
  =
  2U
  \geq
  w_i+U
  =
  W
  \geq
  K.
\]

These four cases cover every possible generator output and show that the
invariant is preserved after every round.

It remains to show that the stopping condition is reached within \(K-1\)
rounds.  Each round removes one candidate.  If the process had not stopped
after \(K-1\) rounds, only one candidate would remain.  Its weight would
equal \(W\geq K\), contradicting the assumption that the stopping condition
had not been reached.  Hence some surviving index \(i^*\) reaches
\[
  w_{i^*}
  =
  2^{c_{i^*}}
  \geq
  K
  =
  2^m
\]
within \(K-1\) rounds, and therefore \(c_{i^*}\geq m\).
By construction,
every response is positive for \(h_{i^*}\).  If the process stops early,
use fresh points of \(A_{i^*}\) as demonstrations to pad the play to
horizon \(K-1\); the mistakes already forced are unaffected.
The generator was arbitrary.  The finite-horizon characterization in
Theorem~\ref{thm:mf-finite-characterization} therefore gives
\(\MF_{K-1}(\mathcal C_m)\geq m\), which is the desired conclusion.
\end{proof}

We now combine independent copies of the construction from
Lemma~\ref{lem:app-single-component-forcing}.  This makes both the
Littlestone dimension and the number of forced mistakes add across
components.

\begin{lemma}
\label{lem:app-product-construction}
For every \(q,m\geq1\), there is a finite UUS class \(\Bcal_{q,m}\) such
that
\[
  \Ldim(\Bcal_{q,m})=q,
  \qquad
  \MF_{2q(2^m-1)}(\Bcal_{q,m})\geq qm.
\]

\end{lemma}

\begin{proof}
Put \(K:=2^m\). Take pairwise disjoint core components

$$
X_\ell:=\{x_{\ell,1},\ldots,x_{\ell,K}\},
\qquad
\ell\in[q],
$$

and, for every vector
\(\boldsymbol i=(i_1,\ldots,i_q)\in[K]^q\), a countably infinite private tail
\(A_{\boldsymbol i}\), with all these sets pairwise disjoint. Define

$$
\supp(h_{\boldsymbol i})
:=
\bigcup_{\ell=1}^{q}
\bigl(X_\ell\setminus\{x_{\ell,i_\ell}\}\bigr)
\cup A_{\boldsymbol i},
\qquad
\Bcal_{q,m}
:=
\{h_{\boldsymbol i}:\boldsymbol i\in[K]^q\}.
$$

The class is finite, and every member has an infinite private tail, so it is UUS.

We first prove that

$$
\Ldim(\Bcal_{q,m})=q.
$$

For the lower bound, consider the depth-\(q\) tree whose query at level
\(\ell\) is \(x_{\ell,1}\). Given any root-to-leaf label sequence, choose

$$
i_\ell
:=
\begin{cases}
1, & \text{if the label at level \(\ell\) is zero},\\
2, & \text{if the label at level \(\ell\) is one}.
\end{cases}
$$

The resulting hypothesis \(h_{\boldsymbol i}\) realizes the chosen path.
Thus the tree is shattered, and

$$
\Ldim(\Bcal_{q,m})\geq q.
$$

For the upper bound, consider the following realizable online prediction
rule. On a core point from \(X_\ell\), predict one until the unique omitted
point \(x_{\ell,i_\ell}\) is observed with label zero. This can cause at most
one mistake and then identifies \(i_\ell\). Hence core points cause at most
one mistake per coordinate.

On a private-tail point, predict zero unless the complete target vector is
already known. A negative label is predicted correctly, whereas a positive
label identifies the target hypothesis immediately, because each private
tail belongs to exactly one hypothesis. If this positive label causes a
mistake after \(r\) coordinates have already been identified, then
\(r\leq q-1\), and the total number of mistakes is at most

$$
r+1\leq q.
$$

Once the target is known, every subsequent label is predicted correctly.
Thus the rule makes at most \(q\) mistakes on every realizable sequence. If
a depth-\((q+1)\) tree were shattered, choosing at every node the label
opposite to the rule's prediction would give a realizable path on which the
rule makes \(q+1\) mistakes, a contradiction. Therefore

$$
\Ldim(\Bcal_{q,m})\leq q,
$$

which proves the equality.

We now prove the generation lower bound by running the
single-component strategy from
Lemma~\ref{lem:app-single-component-forcing}
independently in each coordinate.

Fix an arbitrary deterministic horizon-\(2q(K-1)\) generator.
For each coordinate \(\ell\), maintain a nonempty candidate set
\(S_\ell\subseteq[K]\) and weights
\[
    w_{\ell,i},
    \qquad i\in S_\ell.
\]
Initially,
\[
    S_\ell=[K],
    \qquad
    w_{\ell,i}=1.
\]
As long as coordinate \(\ell\) is unresolved, maintain
\[
    W_\ell
    :=
    \sum_{i\in S_\ell}w_{\ell,i}
    \geq K,
\]
with every \(w_{\ell,i}\) a power of two strictly smaller than \(K\).

The weights have the same interpretation as in
Lemma~\ref{lem:app-single-component-forcing}: doubling
\(w_{\ell,i}\) records a generated point that is invalid for every
maintained target whose \(\ell\)-th coordinate equals \(i\).

A coordinate \(\ell\) becomes resolved as soon as some surviving
weight reaches \(K\). At that moment, choose an index
\(i_\ell^*\in S_\ell\) satisfying
\[
    w_{\ell,i_\ell^*}\geq K,
\]
replace \(S_\ell\) permanently by the singleton
\[
    S_\ell:=\{i_\ell^*\},
\]
and never update this coordinate again. Thus, once a coordinate
resolves, its eventual value is frozen.

At every time, let
\[
    \mathcal P
    :=
    \prod_{\ell=1}^q S_\ell
\]
be the pool of target vectors that the adversary continues to
maintain. This need not contain every hypothesis consistent with the
demonstrations, because freezing a coordinate may discard additional
consistent hypotheses. We maintain the invariant that every
demonstration revealed so far is positive for every
\(h_{\boldsymbol i}\) with \(\boldsymbol i\in\mathcal P\).
Freezing only shrinks \(\mathcal P\), so it preserves this invariant.

While at least one coordinate remains unresolved, respond to the
generator as follows.

\medskip
\noindent
\textbf{Case 1.}
Suppose the generator outputs a core point \(x_{\ell,i}\) from an
unresolved coordinate \(\ell\). Apply the corresponding core-point
response from the proof of
Lemma~\ref{lem:app-single-component-forcing}
to coordinate \(\ell\).

That response reveals a core point \(x_{\ell,j}\), removes \(j\) from
\(S_\ell\), doubles precisely the surviving weights corresponding to
coordinate values for which \(x_{\ell,i}\) is invalid, and preserves
\[
    W_\ell\geq K.
\]
If a surviving weight reaches \(K\), immediately freeze coordinate
\(\ell\) at one such index.

\medskip
\noindent
\textbf{Case 2.}
Suppose the generator outputs a point from
\(A_{\boldsymbol i}\), where
\[
    \boldsymbol i\in\mathcal P.
\]
Choose any unresolved coordinate \(\ell\), reveal
\(x_{\ell,i_\ell}\), and remove \(i_\ell\) from \(S_\ell\).

The generated tail point is valid only for \(h_{\boldsymbol i}\), and
this hypothesis is eliminated by the response. Consequently, the
generated point is invalid for every target vector that remains in
\(\mathcal P\). Charge this mistake to coordinate \(\ell\) by doubling
all remaining weights in that coordinate.

The calculation from Case~4 in the proof of
Lemma~\ref{lem:app-single-component-forcing}
shows that the updated weights still satisfy
\[
    W_\ell\geq K.
\]
Indeed, writing
\[
    U
    :=
    \sum_{j\in S_\ell\setminus\{i_\ell\}}w_{\ell,j},
\]
that calculation gives
\[
    w_{\ell,i_\ell}\leq U
    \qquad\text{and hence}\qquad
    2U\geq W_\ell\geq K.
\]
If a surviving weight reaches \(K\), freeze coordinate \(\ell\).

\medskip
\noindent
\textbf{Case 3.}
Suppose the generated point is invalid for every
\(h_{\boldsymbol i}\) with \(\boldsymbol i\in\mathcal P\).
Choose any unresolved coordinate \(\ell\) and apply Case~3 from the
proof of Lemma~\ref{lem:app-single-component-forcing}:
choose an index \(j\in S_\ell\) of minimum weight, reveal
\(x_{\ell,j}\), remove \(j\), and double all remaining weights in that
coordinate.

An unresolved coordinate has at least two candidates; otherwise its
single weight would equal \(W_\ell\geq K\), and it would already be
resolved. Thus
\[
    w_{\ell,j}\leq\frac{W_\ell}{2},
\]
and the updated total satisfies
\[
    2(W_\ell-w_{\ell,j})
    \geq
    W_\ell
    \geq
    K.
\]
Freeze coordinate \(\ell\) if a surviving weight reaches \(K\).

\medskip
\noindent
\textbf{Case 4.}
The remaining possibility is that the generator outputs a core point
\(x_{\ell,i}\) from a resolved coordinate \(\ell\), with
\[
    i\neq i_\ell^*.
\]
This point is positive for every target remaining in \(\mathcal P\).
Reveal \(x_{\ell,i}\) itself, after which it becomes unavailable for
future output.

Notice that the omitted point \(x_{\ell,i_\ell^*}\) is invalid for
every maintained target and therefore belongs to Case~3, not
Case~4.

\medskip

These cases are exhaustive. A core point from an unresolved coordinate
belongs to Case~1. A private-tail point belongs to Case~2 if its vector
is in \(\mathcal P\); otherwise it is invalid for every maintained
target and belongs to Case~3. For a resolved coordinate, its omitted
core point belongs to Case~3, while every other available core point
belongs to Case~4.

Moreover, every response is positive for every target remaining in
\(\mathcal P\). Thus the target-consistency invariant is preserved.

The same argument remains valid when the construction is embedded in a
larger disjoint example space and every hypothesis is extended by zero
outside its own component. Any generated point outside the component is
invalid for every maintained target and therefore falls under Case~3.

Call a round in Cases~1--3 a progress round. Every progress round
removes one candidate from an unresolved coordinate. A coordinate can
undergo at most \(K-1\) such removals before it resolves: after
\(K-1\) removals, only one candidate would remain, and its weight would
equal
\[
    W_\ell\geq K.
\]
Hence there are at most
\[
    q(K-1)
\]
progress rounds.

Case~4 can occur at most \(K-1\) times for each resolved coordinate,
because there are only \(K-1\) core points
\(x_{\ell,i}\) with \(i\neq i_\ell^*\), and each such point is
immediately revealed and becomes unavailable. Thus there are at most
another
\[
    q(K-1)
\]
Case~4 rounds. Consequently, every coordinate resolves within
\[
    2q(K-1)
\]
rounds.

Let
\[
    \boldsymbol i^*
    :=
    (i_1^*,\ldots,i_q^*)
\]
be the final frozen vector. For each coordinate \(\ell\), the weight
\(w_{\ell,i_\ell^*}\) started at one and reached at least
\[
    K=2^m.
\]
It therefore doubled at least \(m\) times.

Every doubling of this final surviving weight occurred on a round whose
generated point is invalid for \(h_{\boldsymbol i^*}\). In Case~1, the
weight \(w_{\ell,i_\ell^*}\) doubles only when the generated core point
is invalid for coordinate value \(i_\ell^*\). In Cases~2 and~3, the
generated point is invalid for every target remaining in
\(\mathcal P\), including \(h_{\boldsymbol i^*}\).

Only one coordinate is updated on each progress round. Therefore, the
\(m\) doublings in each of the \(q\) coordinates occur on \(qm\)
distinct rounds. It follows that the generator makes at least
\[
    qm
\]
mistakes relative to \(h_{\boldsymbol i^*}\).

By the target-consistency invariant, every demonstration is positive
for \(h_{\boldsymbol i^*}\). If all coordinates resolve before round
\(2q(K-1)\), complete the trajectory using fresh demonstrations from
\(A_{\boldsymbol i^*}\). The mistakes already forced are unaffected.
Therefore,
\[
    \MF_{2q(K-1)}(\Bcal_{q,m})
    \geq
    qm.
\]
Since \(K=2^m\), this becomes
\[
    \MF_{2q(2^m-1)}(\Bcal_{q,m})
    \geq
    qm,
\]
as claimed.
\end{proof}

Next, we combine independent copies of the construction from
Lemma~\ref{lem:app-product-construction} to obtain the desired lower bound.  This makes both the
Littlestone dimension and the number of forced mistakes add across
components.
\begin{theorem*}[Restatement of Theorem~\ref{thm:sharp-ldim-generation}]
For every integer \(d\geq2\), there exists a countable UUS class
\(\Hcal_d^{\mathrm{sharp}}\) satisfying
\[
  \Ldim(\Hcal_d^{\mathrm{sharp}})=d
\]
such that, for every \(T\geq3\),
\[
  \MF_T(\Hcal_d^{\mathrm{sharp}})
  =
  \Theta\left(
    d\log\left(1+\frac{T}{d}\right)
  \right),
\]
where the implicit constants are universal.
\end{theorem*}

\begin{proof}
Fix \(d\geq2\).  For every \(1\leq q\leq d-1\) and \(m\geq1\), take a
disjoint copy of \(\Bcal_{q,m}\) from
Lemma~\ref{lem:app-product-construction} and add a new marker \(s_{q,m}\).
Extend every \(h\in\Bcal_{q,m}\) to the common disjoint example space by
\[
  \supp(\widetilde h)
  :=
  \{s_{q,m}\}\cup\supp(h),
\]
and let
\[
  \Hcal_d^{\mathrm{sharp}}
  :=
  \{\widetilde h:
      h\in\Bcal_{q,m},\ 1\leq q\leq d-1,\ m\geq1\}.
\]
There are countably many pairs \((q,m)\), and every component contains
finitely many hypotheses, so \(\Hcal_d^{\mathrm{sharp}}\) is countable.  Each hypothesis
retains its countably infinite private tail, so the class is UUS.

We next verify \(\Ldim(\Hcal_d^{\mathrm{sharp}})=d\).  A single marked component has the
same Littlestone dimension \(q\) as \(\Bcal_{q,m}\): its marker is constant
one on that component, and every point outside its component is constant
zero, so the extension introduces no new nonconstant split within the
component.

For the lower bound, select a component with \(q=d-1\) and use its marker
as the root query.  The positive restriction contains that component and
has Littlestone dimension \(d-1\).  The negative restriction contains a
different component with \(q=d-1\), because all hypotheses in the second
component label the selected marker zero.  It too has dimension at least
\(d-1\).  Placing the marker above shattered trees of depth \(d-1\) in
these two restrictions gives a shattered tree of depth \(d\).  Hence
\(\Ldim(\Hcal_d^{\mathrm{sharp}})\geq d\).

For the upper bound, observe that every point is either a unique marker or
belongs to the universe of a unique component.  Its positive restriction
is therefore contained in a single marked component, of dimension at most
\(d-1\).  If a point rooted a shattered tree of depth \(d+1\), both of its
label restrictions would have Littlestone dimension at least \(d\), which
is impossible for the positive restriction.  Thus
\(\Ldim(\Hcal_d^{\mathrm{sharp}})\leq d\).

We now prove the generation lower bound.  Fix \(T\geq3\), and choose any
parameters satisfying
\[
  1\leq q\leq d-1,
  \qquad
  m\geq1,
  \qquad
  1+2q(2^m-1)\leq T.
\]
Against an arbitrary horizon-\(T\) generator, reveal \(s_{q,m}\) on the
first round.  This response restricts the eventual target to the selected
component.  During the next \(2q(2^m-1)\) rounds, use the product adversary
from Lemma~\ref{lem:app-product-construction}.   The first-round marker is disjoint from the component
universe and does not interfere with the construction.  The product
adversary forces \(qm\) mistakes, and fresh points in the final target's
private tail pad the trajectory to length \(T\).  Hence
\[
  \MF_T(\Hcal_d^{\mathrm{sharp}})
  \geq
  \max_{\substack{1\leq q\leq d-1,\ m\geq1\\
                   1+2q(2^m-1)\leq T}}
  qm.
\]

It remains to select \(q\) and \(m\).  First suppose \(3\leq T\leq2d-1\).
Set
\[
  m=1,
  \qquad
  q=\left\lfloor\frac{T-1}{2}\right\rfloor.
\]
Then \(1\leq q\leq d-1\), the horizon constraint holds, and
\(q\geq T/4\).  Since the base-two logarithm satisfies
  \(\log(1+x)\leq(\log e)x\) for \(x\geq0\),
\[
  d\log\left(1+\frac{T}{d}\right)
  \leq(\log e)T.
\]
Thus \(qm=q\) is at least a universal constant multiple of the desired
quantity.

Now suppose \(T\geq2d\).  Set
\[
  q=d-1,
  \qquad
  m=
  \left\lfloor
    \log\left(1+\frac{T-1}{2(d-1)}\right)
  \right\rfloor.
\]
The definition of \(m\) gives
\(2^m\leq1+(T-1)/(2(d-1))\), so the horizon constraint holds.  Put
\(y:=T/d\geq2\).  Since \(d\geq2\), we have \(q=d-1\geq d/2\), and
\[
  1+\frac{T-1}{2(d-1)}
  \geq
  1+\frac y2
  \geq
  \sqrt{1+y}.
\]
The logarithm inside the floor is at least one.  For every \(a\geq1\),
\(\lfloor a\rfloor\geq a/2\); therefore
\[
  m
  \geq
  \frac12
  \log\left(1+\frac{T-1}{2(d-1)}\right)
  \geq
  \frac14\log(1+y).
\]
It follows that \(qm\) is at least a universal constant multiple of
\(d\log(1+T/d)\).  Combining the two regimes proves
\[
  \MF_T(\Hcal_d^{\mathrm{sharp}})
  =
  \Omega\left(d\log\left(1+\frac{T}{d}\right)\right).
\]

For the matching upper bound, we use
Theorem~\ref{thm:finite-ldim-realizable}, proved in
Appendix~\ref{app:finite-ldim-upper}, together with the already established
identity \(\Ldim(\Hcal_d^{\mathrm{sharp}})=d\).  It gives
\[
  \MF_T(\Hcal_d^{\mathrm{sharp}})
  =
  O\left(d\log\left(1+\frac{T}{d}\right)\right).
\]
The lower and upper estimates complete the proof.
\end{proof}

\section{Finite VC Dimension Does Not Imply No-Regret Generation}
\label{app:finite-vc}

We show that the finite-Littlestone assumption in
Theorems~\ref{thm:finite-ldim-realizable}
and~\ref{thm:finite-ldim-agnostic} cannot be replaced by finite VC
dimension. The example is a countable class of infinite paths in a
binary tree. Although no pair of examples is shattered, the adversary
can extend the demonstrated path so that every generated example is
invalid.

\begin{proposition}
\label{prop:finite-vc-linear-regret}
There exists a countable UUS class \(\Hcal\) such that
$$
    \operatorname{VCdim}(\Hcal)=1
$$
and
$$
    \MF_T(\Hcal)=\SMF_T(\Hcal)=T
    \qquad
    \text{for every }T\geq1.
$$
Consequently, finite VC dimension does not imply either realizable
or agnostic no-regret generation.
\end{proposition}

\begin{proof}
Let the example space consist of all nonempty finite binary strings:
$$
    \Xcal:=\bigcup_{n\geq1}\{0,1\}^n.
$$
For an infinite binary sequence \(b=(b_1,b_2,\ldots)\), define
\(h_b\in\{0,1\}^{\Xcal}\) by
$$
    \supp(h_b):=\{b_{1:n}:n\geq1\}.
$$
Thus, an example is valid under \(h_b\) precisely when it is a
nonempty prefix of \(b\). Consider the class
$$
    \Hcal
    :=
    \left\{
        h_b:
        b\in\{0,1\}^{\N}
        \text{ is eventually zero}
    \right\}.
$$
Every eventually-zero sequence is determined by a finite binary
prefix followed by zeros, so \(\Hcal\) is countable. Each support
contains one string of every positive length and is therefore
infinite. Hence \(\Hcal\) is UUS.

\medskip
\noindent
\emph{VC dimension.}
The singleton consisting of the one-bit string \(0\) is shattered:
the sequence \(000\cdots\) gives label one, whereas \(1000\cdots\)
gives label zero.

To see that no pair is shattered, take distinct strings
\(u,v\in\Xcal\). If neither is a prefix of the other, no infinite
sequence has both as prefixes, so the labeling
\(h(u)=h(v)=1\) is impossible. Otherwise, suppose without loss of
generality that \(u\) is a proper prefix of \(v\). Every hypothesis
satisfies
$$
    h(v)=1 \quad\Longrightarrow\quad h(u)=1,
$$
so the labeling \(h(u)=0\), \(h(v)=1\) is impossible.
Thus \(\operatorname{VCdim}(\Hcal)=1\).

\medskip
\noindent
\emph{Forcing a mistake on every round.}
Fix a horizon \(T\geq1\) and an arbitrary deterministic horizon-\(T\)
generator \(G_T\). We construct a legal demonstration sequence
\(x_{1:T}\) and a single hypothesis \(h^\star\in\Hcal\) satisfying
$$
    h^\star(x_t)=1,
    \qquad
    h^\star(\widehat x_t)=0
    \qquad
    \text{for every }t\leq T,
$$
where \(\widehat x_t=G_T(x_{1:t-1})\).

We construct a binary path one bit at a time. Let \(p_0\) be the
empty string. At round \(t\), extend \(p_{t-1}\) by one bit to obtain
a string \(p_t\) of length \(t\), and reveal
$$
    x_t:=p_t.
$$
In particular, before round \(t\), every nonempty prefix of
\(p_{t-1}\) has already been demonstrated.

Call two finite strings \emph{incompatible} if neither is a prefix
of the other. We choose \(p_t\) so that \(\widehat x_t\) is
incompatible with \(p_t\). Because the generator's output is fresh,
it cannot be a nonempty prefix of \(p_{t-1}\). There are therefore
two cases:
\begin{enumerate}
    \item
    If \(\widehat x_t\) is incompatible with \(p_{t-1}\), set
    \(p_t=p_{t-1}0\). The output remains incompatible with this
    extension.

    \item
    Otherwise, \(\widehat x_t\) strictly extends \(p_{t-1}\).
    Let \(a\in\{0,1\}\) be its next bit, so that
    \(\widehat x_t=p_{t-1}a u\) for some possibly empty string \(u\).
    Set
    $$
        p_t=p_{t-1}(1-a).
    $$
    The strings \(p_t\) and \(\widehat x_t\) now disagree at
    position \(t\), and hence are incompatible.
\end{enumerate}
Here juxtaposition denotes concatenation. At the first round,
every possible output strictly extends the empty string, so the
second case applies.

The demonstrations are pairwise distinct because \(|p_t|=t\).
Moreover, incompatibility persists under extension. Consequently,
after round \(t\), every output
\(\widehat x_1,\ldots,\widehat x_t\) is incompatible with \(p_t\).
Every \(p_t\) has an eventually-zero infinite extension, so the
demonstrations remain jointly realizable throughout the construction.

After \(T\) rounds, let \(b\) be the infinite sequence obtained by
appending zeros to \(p_T\), and set \(h^\star:=h_b\).
Then \(h^\star\in\Hcal\). Each demonstration \(x_t=p_t\) is a
prefix of \(b\), whereas every generated output is incompatible
with \(p_T\) and therefore cannot be a prefix of \(b\). Thus
$$
    M_T(G_T,h^\star;x_{1:T})
    =
    \sum_{t=1}^T\ell_{h^\star}(\widehat x_t)
    =
    T.
$$
Since all demonstrations are valid under \(h^\star\), we also have
$$
    \Reg_T(G_T,h^\star;x_{1:T})
    =
    M_T(G_T,h^\star;x_{1:T})
    -
    \sum_{t=1}^T\ell_{h^\star}(x_t)
    =
    T.
$$

Although \(h^\star\) is specified at the end of the construction,
it is one hypothesis witnessing every round of the trajectory.
For fixed \(G_T\) and \(T\), the entire construction can be
simulated in advance, yielding a fixed target and demonstration
sequence. The finite sequence can also be extended to a complete
enumeration of \(\supp(h^\star)\) by revealing the remaining
prefixes of \(b\).

Because \(G_T\) was arbitrary, the minimax characterizations in
Theorems~\ref{thm:mf-finite-characterization}
and~\ref{thm:smf-characterization} imply
$$
    \MF_T(\Hcal)\geq T,
    \qquad
    \SMF_T(\Hcal)\geq T.
$$
The reverse inequalities follow because each round contributes
at most one mistake and at most one unit of regret. This proves
the claimed equalities for every \(T\geq1\).
\end{proof}

\end{document}